\documentclass[twocolumn,switch,times]{article}
\usepackage{preprint}
\usepackage{hyperref}
\usepackage[authoryear,round]{natbib}

\hypersetup{
    colorlinks=true,
    linkcolor=blue,
    urlcolor=blue,
    citecolor=blue,
    anchorcolor=black
}
\usepackage[utf8]{inputenc}	% allow utf-8 input
\usepackage[T1]{fontenc}
\usepackage[table,dvipsnames]{xcolor}
\usepackage{lineno}					% Line numbers
\usepackage{tikz} 					% ORCiD insertion
\usepackage{fancyhdr}
\usepackage{geometry}
\usepackage{lipsum}
\usepackage{scalerel}
\usepackage{sidecap}
\usepackage{times}
\usepackage{doi}
\usepackage{amssymb}
\usepackage{amsmath}
\usepackage{bbm}
\usepackage{bm}
\usepackage{array,booktabs}
\usepackage{upgreek}
\usepackage{algorithm}
\usepackage{algpseudocodex}
\usepackage{appendix}
\usepackage{authblk}
\usepackage{placeins}

\usepackage{newfloat}
\DeclareFloatingEnvironment[name={Supplementary Figure}]{suppfigure}
\usepackage{sidecap}
\sidecaptionvpos{figure}{c}
\usepackage{etoolbox}
\usepackage{microtype}
\usepackage{tablefootnote}
\usepackage{siunitx}

\usepackage{titlesec}
\titlespacing\section{0pt}{12pt plus 3pt minus 3pt}{1pt plus 1pt minus 1pt}
\titlespacing\subsection{0pt}{10pt plus 3pt minus 3pt}{1pt plus 1pt minus 1pt}
\titlespacing\subsubsection{0pt}{8pt plus 3pt minus 3pt}{1pt plus 1pt minus 1pt}

\definecolor{lime}{HTML}{A6CE39}
\definecolor{lightgray}{gray}{0.9} % For table row coloring

\newcommand{\mathbfcal}[1]{\bm{\mathcal{#1}}}
\newcommand{\real}{\mathbbm{R}}
\newcommand{\integer}{\mathbbm{Z}}
\newcolumntype{P}[1]{>{\centering\arraybackslash}p{#1}}
\newcolumntype{R}[1]{>{\raggedleft\arraybackslash}p{#1}}
\newcolumntype{L}[1]{>{\raggedright\arraybackslash}p{#1}}
\newcommand{\Continue}{\textbf{continue}}
\newcommand{\appendixref}[1]{%
  \hyperref[#1]{Appendix~\ref*{#1}}\unskip
}
\newcommand*{\belowrulesepcolor}[1]{% 
  \noalign{% 
    \kern-\belowrulesep 
    \begingroup 
      \color{#1}% 
      \hrule height\belowrulesep 
    \endgroup 
  }%
} 
\newcommand*{\aboverulesepcolor}[1]{% 
  \noalign{% 
    \begingroup 
      \color{#1}% 
      \hrule height\aboverulesep 
    \endgroup 
    \kern-\aboverulesep 
  }%
}

\title{Benchmarking individual tree segmentation using multispectral airborne laser scanning data: the FGI-EMIT dataset}

\usepackage{authblk}

\author[1,\small\ensuremath{\ast}]{Lassi Ruoppa}
\author[1]{Tarmo Hietala}
\author[1]{Verneri Seppänen}
\author[1]{Josef Taher}
\author[1]{Teemu Hakala}
\author[1]{Xiaowei Yu}
\author[1,2]{Antero Kukko}
\author[1]{\\Harri Kaartinen}
\author[1,2]{Juha Hyyppä}

\affil[1]{Department of Remote Sensing and Photogrammetry, Finnish Geospatial Research Institute FGI, The National Land Survey of Finland, Vuorimiehentie 5, FI-02150, Espoo, Finland}
\affil[2]{Department of Built Environment, School of Engineering, Aalto University, P.O. Box 11000, FI-00076, Aalto, Finland}

\begin{document}

\twocolumn[\begin{@twocolumnfalse}

\maketitle

\begin{abstract}
Individual tree segmentation (ITS) from LiDAR point clouds is fundamental for applications such as forest inventory, carbon monitoring and biodiversity assessment. Traditionally, ITS has been achieved with unsupervised geometry-based algorithms, while more recent advances have shifted toward supervised deep learning (DL). In the past, progress in method development was hindered by the lack of large-scale benchmark datasets, and the availability of novel data formats, particularly multispectral (MS) LiDAR, remains limited to this day, despite evidence that MS reflectance can improve the accuracy of ITS. This study introduces FGI-EMIT, the first large-scale multispectral airborne laser scanning benchmark dataset for ITS. Captured at wavelengths 532, 905, and 1,550 nm, the dataset consists of 1,561 manually annotated trees, with a particular focus on small understory trees. Using FGI-EMIT, we comprehensively benchmarked four conventional unsupervised algorithms and four supervised DL approaches. Hyperparameters of unsupervised methods were optimized using a Bayesian approach, while DL models were trained from scratch. Among the unsupervised methods, Treeiso achieved the highest test set F1-score of 52.7\%. The DL approaches performed significantly better overall, with  the best model, ForestFormer3D, attaining an F1-score of 73.3\%. The most significant difference was observed in understory trees, where ForestFormer3D exceeded Treeiso by 25.9 percentage points. An ablation study demonstrated that current DL-based approaches generally fail to leverage MS reflectance information when it is provided as additional input features, although single channel reflectance can improve accuracy marginally, especially for understory trees. A performance analysis across point densities further showed that DL methods consistently remain superior to unsupervised algorithms, even at densities as low as 10 points/m$^2$. To support future benchmarking efforts and method development, we make the FGI-EMIT dataset publicly available (\textcolor{Red}{\emph{link to data will be added upon acceptance of the manuscript}}).
\end{abstract}


\vspace{0.5cm}

\end{@twocolumnfalse}]

\renewcommand{\thefootnote}{\small\ensuremath{\ast}}
\footnotetext[1]{Corresponding author: Lassi Ruoppa (lassi.ruoppa@nls.fi)}
\renewcommand{\thefootnote}{\arabic{footnote}}
% Reset footnote counter
\setcounter{footnote}{0}

\section{Introduction} \label{section:introduction}

Individual tree segmentation (ITS) from point clouds acquired with light detection and ranging (LiDAR) scanners is one of the most widely studied segmentation tasks within forestry, since tree level information is essential for a wide range of applications in both ecological and economical contexts. Examples include studying leaf phenology and carbon dynamics \citep{dai2018new}, updating forest inventories, estimating growth and identifying trees with high biodiversity value \citep{cao2023benchmarking}.

Traditionally, ITS has been achieved with unsupervised algorithms that rely on point cloud geometry and complex heuristic rules. Early methods were primarily designed for sparse airborne laser scanning (ALS) data, where understory trees are often not visible, and thus operated on 2D canopy projections \citep{hyyppä2001segmentation,popescu2004seeing,koch2006detection}. Later developments shifted to processing the data in 3D formats, such as voxels \citep{wang2008lidar,mongus2015efficient}, and the more recent 3D approaches generally outperform 2D methods \citep{cao2023benchmarking,zhang2024individual}, although accurately detecting small understory trees remains a persistent challenge. Following the rapid advances of deep learning (DL) in computer vision and remote sensing, several DL-based ITS models have been proposed in recent years \citep[see e.g.][]{wielgosz2023point2tree,xiang2024automated,wielgosz2024segmentanytree,xi2025new,xiang2025forestformer3d}. These models typically achieve state-of-the-art performance across several forest and data types and often surpass modern unsupervised algorithms \citep{xiang2024automated}. However, their reliance on large amounts of manually annotated training data limits their generalizability, particularly to previously unseen forest types.

In the past, the lack of publicly available, large-scale benchmark datasets was considered one of the primary factors limiting the development of forest point cloud segmentation models \citep{lines2022ai,hamedianfar2022deep}. The introduction of datasets such as LAUTx \citep{tockner2022automatic}, Wytham Woods \citep{calders2022laser}, FOR-Instance \citep{puliti2023forinstance}, and its successor FOR-InstanceV2 \citep{xiang2025forestformer3d}, has helped address this limitation in recent years. Nevertheless, the demand for additional publicly available data with manually generated annotations remains high, particularly since transformer-based models, which currently constitute the state of the art in several computer vision tasks, are substantially more data-hungry than conventional convolutional neural networks (CNNs) \citep{wang2022towards}. Another notable gap lies in the limited availability of novel point cloud data formats, such as multispectral (MS) LiDAR, despite evidence that multi-channel reflectance information can improve the accuracy of both instance \citep{dai2018new,huo2020individual} and semantic segmentation \citep{ruoppa2025unsupervised,takhtkeshha20253d} in forest data. Moreover, existing datasets generally focus exclusively on forested environments and exclude man-made structures such as buildings or vehicles, which limits the applicability of models trained on such data in urban forests where previously unseen object classes are often misclassified as trees.

Given the large number of existing ITS methods, systematic benchmarking is extremely important. Accurate and extensive comparisons enable end-users to select the most suitable approach for their specific application and ensure optimal performance in downstream tasks. As a consequence, several studies have benchmarked ITS algorithms over the years. However, effectively all of these studies suffer from several limitations, primarily stemming from the historical lack of large-scale, manually labeled ITS datasets. Moreover, due to the rapid development of DL-based ITS models in recent years, most previous benchmarks no longer reflect the current state of the art.

While some early studies even evaluated performance primarily based on visual inspection \citep{gupta2010comparative}, the most notable limitation of most prior ITS benchmarks is the reliance on field inventory data as ground truth. In these cases, the correctness of predicted segments is assessed only by location and height \citep{edson2011airborne,kaartinen2012international,eysn2015benchmark,wang2016international,ma2022performance,liu2023study,saeed2024performance,zhang2024individual}, ignoring factors such as shape. Some studies also considered crown size \citep{vauhkonen2011comparative,wallace2014evaluating,nemmaoui2024benchmarking}, yet segments considered correct could still contain multiple trees. More recent works have matched predictions to crown polygons derived from aerial imagery \citep{aubrykientz2019comparative,fraser2025quantifying}. However, as noted by \citet{steier2024is} and \citet{allen2025manual}, 2D annotations result in severely overestimating accuracy in multi-layered forests. Even studies using 3D annotations, such as \citet{cao2023benchmarking}, matched predictions based on 2D intersection over union (IoU) of crown polygons, capturing crown shape but not overall segment quality. Robust evaluation with 3D IoU, which is standard in modern 3D instance segmentation, remains rare and typically limited in scope.

Parameter tuning presents another significant issue. Despite the strong dependence of ITS performance on hyperparameters, most previous benchmarks do not address parameter optimization \citep[see e.g.][]{edson2011airborne,kaartinen2012international,eysn2015benchmark,wang2016international}. Others have relied on manually selecting the hyperparameters based on trial and error \citep{ma2022performance,saeed2024performance,zhang2024individual,cherlet2024benchmarking}, with only a few works employing robust hyperparameter optimization strategies, such as grid search \citep{aubrykientz2019comparative,cao2023benchmarking,nemmaoui2024benchmarking}. As a result, reported performance may often not reflect the true potential of the algorithms, reducing the reliability of comparisons.

Finally, the limited availability of open-source implementations has restricted several studies to mostly ITS methods available in the lidR library \citep{roussel2020lidr}, limiting benchmarking efforts to a narrow subset of algorithms \citep[see e.g.][]{cao2023benchmarking,nemmaoui2024benchmarking,saeed2024performance}. Similarly, many works have primarily focused on 2D local-maximum-based approaches \citep{kaartinen2012international,eysn2015benchmark,fraser2025quantifying}. While some recent works include DL-based methods, these are usually restricted to one \citep{zhang2024individual} or two \citep{cherlet2024benchmarking} models, which are not necessarily trained from scratch.

These limitations highlight a significant research gap: the lack of comprehensive ITS benchmarks conducted on large-scale datasets with 3D annotations, robust evaluation metrics, and systematic hyperparameter optimization. In this work, we present \textbf{FGI-EMIT} (\textbf{F}innish \textbf{G}eospatial Research \textbf{I}nstitute’s \textbf{E}spoonlahti \textbf{M}ultispectral \textbf{I}ndividual \textbf{T}rees), a multispectral, high-density ALS benchmark dataset for individual tree segmentation. Captured in the Espoonlahti district of Espoo, Finland, the dataset covers boreal forests composed of both natural and planted trees spanning more than 20 species, alongside a variety of man-made structures. Using FGI-EMIT, we perform an extensive performance comparison of individual tree segmentation methods, including both unsupervised algorithms and state-of-the-art deep learning approaches. Our main contributions are as follows:
\begin{enumerate}
    \item We introduce FGI-EMIT, the first large-scale multispectral LiDAR dataset for individual tree segmentation. The dataset consists of 1,561 manually annotated trees, comparable in size to the original FOR-Instance dataset \citep{puliti2023forinstance}, with particular emphasis on small understory trees that are notoriously challenging to segment. FGI-EMIT is also the first ITS dataset to include built environment, extending applicability to urban forests. \textcolor{Red}{\emph{The dataset will be made available upon acceptance of the manuscript. Samples of the data are available upon request.}}
    \item We present a comprehensive performance comparison of existing individual tree segmentation methods, evaluating four widely used unsupervised algorithms and four recent deep learning models. To ensure a fair comparison, we determine optimal hyperparameter values for each unsupervised algorithm using the FGI-EMIT training set and Bayesian optimization. Similarly, the supervised deep learning models are trained from scratch on the dataset. The benchmark results demonstrate that 3D deep learning approaches not only outperform unsupervised algorithms by a significant margin in terms of accuracy, but also maintain this advantage across a wide range of point densities.
    \item We conduct the first ever ablation study on the effects of utilizing multi-channel reflectance information as input features for DL-based ITS, demonstrating that while single-channel reflectance can improve segmentation accuracy in some cases, particularly for small understory trees, existing architectures fail to effectively exploit reflectance features, with effects ranging from modestly positive to clearly detrimental.
\end{enumerate}

\section{Related work}

\subsection{Point cloud individual tree segmentation datasets}

Following the recent rise of machine and deep learning in remote sensing applications, several LiDAR-based forest point cloud datasets with instance-level annotations have been released. One of the earliest efforts was NeonTreeEvaluation \citep{weinstein2021benchmark}, whose instance annotations were derived from 2D unmanned aerial vehicle (UAV) image bounding boxes, resulting in limited applicability due to low 3D annotation quality. The first public datasets with true 3D instance annotations were LAUTx \citep{tockner2022automatic} and the dataset of \citet{weiser2022terrestrial}, both captured in mixed temperate forests. The former consists of personal laser scanning (PLS) point clouds, while the latter combines data from multiple platforms, including ALS, UAV laser scanning (ULS), and terrestrial laser scanning (TLS).

A pioneering work in the context of forest segmentation benchmark datasets, FOR-Instance \citep{puliti2023forinstance} was the first machine-learning-ready forest point cloud dataset providing both instance and semantic annotations. The dataset was later extended into FOR-InstanceV2 \citep{xiang2025forestformer3d} by combining it with other existing datasets and introducing additional annotated data. FOR-InstanceV2 is currently the largest forest point cloud segmentation dataset, spanning multiple continents and scanner types, including ULS, TLS and mobile laser scanning (MLS). EvoMS \citep{ruoppa2025unsupervised} remains the only publicly available dataset with multispectral reflectance information and instance-level annotations. However, since the dataset was primarily intended for semantic segmentation, it is relatively limited in size, which limits its suitability for training instance segmentation models.

\autoref{table:dataset_comparison} provides an overview of publicly available point cloud datasets for individual tree segmentation. The comparison includes key dataset attributes, specifically dataset size in terms of annotated trees, availability of semantic labels, number of available reflectance channels, sensor modalities, and forest types represented.

Since manual 3D point cloud annotation is extremely time-consuming, many existing datasets have utilized some degree of automation during the labeling process. For example, the Wytham Woods \citep{calders2022laser} and TreeLearn \citep{henrich2024treelearn} datasets were first segmented into individual trees using automated algorithms, after which errors were corrected manually. By contrast, annotations in WildForest3D \citep{kalinicheva2022multilayer} were derived directly from expert field measurements. In fact, based on the associated publications, only four of the datasets presented in \autoref{table:dataset_comparison} contain annotations that were produced entirely manually in 3D. Specifically, LAUTx, FOR-Instance, EvoMS, and ForestSemantic, the last of which is only partially public. All others were either labeled in 2D or included algorithmic steps in the annotation pipeline. While most datasets incorporated a manual correction phase, the use of automated methods may still introduce systematic biases, potentially affecting hyperparameter tuning, DL model training, and performance evaluation. Synthetic data presents a promising research direction for reducing the time required for manual annotation. For example, \citet{she2025scaling} recently proposed CAMP3D, a pipeline for generating instance-labeled forest data using a combination of Unreal Engine and HELIOS++ \citep{winiwarter2022virtual}.

Although NeonTreeEvaluation is currently the largest dataset containing LiDAR point clouds with tree instance annotations, its ground truth labels were derived from 2D bounding boxes created for aerial imagery \citep{weinstein2021benchmark}. This approach leads to systematic errors: adjacent trees in dense areas are merged into single instances, and understory trees are incorrectly assigned to nearby overstory trees. As a result, the dataset may have limited suitability for training accurate point cloud ITS models, and the LiDAR data is more appropriate as auxiliary information for UAV-imagery-based 2D segmentation models.

Outside of LiDAR point cloud datasets designed specifically for segmentation, several other forestry-related datasets have been released in recent years. While not directly applicable for training ITS models, such datasets can serve as valuable auxiliary resources for future work, for example in the context of self-supervised pretraining. Perhaps the most closely related to ITS are LiDAR datasets containing raw point clouds together with manually collected tree heights and locations \citep[see e.g.][]{eysn2015benchmark,liang2018international,monnet2023airborne,dubrovin2024open}. While not applicable for training instance segmentation models, the tree location information can be used for DL-based tree detection. Another important category is UAV imagery annotated with bounding boxes or pixel-wise labels \citep[see e.g.][]{velasquezcamacho2023implementing,velasquezcamacho2023urban,veitchmichaelis2024oamtcd}. In some cases, such imagery can also be transformed into photogrammetric point clouds \citep{cloutier2023quebec,cloutier2024influence}. Finally, several public datasets focus on tree species classification, consisting of point clouds depicting individual trees and the associated species labels. Perhaps most notably, the FOR-species20K \citep{puliti2025benchmarking} includes 20,000 trees across 33 species and three scanner types. Similarly, \citet{taher2025multispectral} introduced a multispectral LiDAR dataset comprising 6,000 trees from nine species. Synthetic datasets have also been proposed, for example TreeNet3D \citep{tang2024treenet3d}, which provides species-labeled tree models. For a more extensive review of publicly available forest datasets, please refer to e.g. \citet{ouaknine2025openforest}.

\begin{table*}[!t]
    \centering
    \caption{Comparison of existing LiDAR individual tree segmentation datasets with instance annotations.} \bigskip
    \rowcolors{1}{}{lightgray}
    \tiny{
    \begin{tabular}{L{1.7cm}P{0.9cm}P{0.8cm}P{1.0cm}P{0.8cm}L{2.0cm}L{2.6cm}L{2.1cm}L{2.1cm}}
        \toprule
        \textbf{Name} & \textbf{Number of trees} & \textbf{Semantic labels} & \textbf{Number of reflectance channels} & \textbf{Scanner type} & \textbf{Forest type} & \textbf{Other information} & \textbf{Dataset} & \textbf{Associated paper(s)} \\ \midrule \midrule \belowrulesepcolor{lightgray}
        NeonTreeEvaluation & 30,975 && 1 & ALS & Boreal, temperate, and subtropical forest, including coniferous-dominated, deciduous-dominated, and mixed stands & Includes RGB and hyperspectral imagery. Annotations generated by draping 2D bounding boxes over point clouds. Consequently, quality of annotations varies. & \citet{weinstein2020weecology} & \citet{weinstein2021benchmark} \\
        LAUTx & 516 && 1 & PLS & Mixed temperate forest & Includes species annotations for a subset of trees. & \citet{tockner2022lautx} & \citet{tockner2022automatic} \\
        German individual tree point clouds and measurements & 1,491 && 1 & ALS, ULS, TLS & Mixed temperate forest & Mixed annotation method (manual or automated, with an optional manual error correction step). Majority of trees captured with multiple different scanners. Contains both leaf-on and leaf-off data. & \citet{weiser2022terrestrial} & \citet{weiser2022individual} \\
        WildForest3D & 1,568 && 1 & ULS & Mixed temperate forest & Instance annotations derived from expert field measurements, not manual point cloud segmentation. Only a subsection of each plot is annotated. & \citet{kalinicheva2022wildforest3d} & \citet{kalinicheva2022multilayer} \\
        Wytham Woods & 835 && 0 & TLS & Deciduous-dominated temperate forest & Individual trees extracted using an automated algorithm with manual inspection and error correction. & \citet{calders2022terrestrial} & \citet{calders2022laser} \\
        Wytham Woods \& Sepilok forest & 556 && 0 & ALS & Deciduous-dominated temperate and tropical forest & Individual tree annotations propagated from TLS data using nearest-neighbor interpolation. TLS annotations were generated using an automated algorithm with manual inspection and error correction. & \citet{cao2022international} & \citet{cao2023benchmarking} \\
        FOR-Instance & 1,130 & $\checkmark$ & 1 & ULS & Coniferous-dominated boreal and temperate forest, deciduous-dominated temperate forest, subtropical forest & Dataset spans multiple countries across two continents. Two different sensors used in data collection. & \citet{puliti2023forinstancedata} & \citet{puliti2023forinstance} \\
        TreeLearn & 200 (6,665) && 0 & MLS & Deciduous-dominated temperate forest & Individual trees extracted using an automated algorithm with manual inspection and error correction for one of the test plots. Number of trees with no manual correction step provided (in brackets). & \citet{henrich2023treelearndataset} & \citet{henrich2024treelearn} \\
        Tropical forest from French Guiana & 281 & $\checkmark$ & 1 & ULS & Tropical forest & Instance and semantic annotations propagated from TLS data using nearest-neighbor interpolation. TLS annotations were generated using an automated algorithm with manual inspection and error correction. & \citet{bai2023uva} & \citet{bai2023semantic} \\
        ForestSemantic & 673 & $\checkmark$ & 1 & TLS & Coniferous-dominated and mixed boreal forest & Dataset only partially available. & \citet{mspace2024forestsemantic} & \citet{liang2024forestsemantic} \\
        NIBIO\_MLS & 258 & $\checkmark$ & 1 & MLS & Coniferous-dominated and deciduous-dominated boreal forest & Later added to FOR-InstanceV2. & \citet{puliti2024nibio} & \citet{wielgosz2023point2tree,wielgosz2024segmentanytree} \\
        EvoMS & 356 & $\checkmark$ & 3 & ALS & Coniferous-dominated and deciduous-dominated boreal forest & The point clouds have been normalized by subtracting a digital terrain model from the $z$-coordinates and ground points have been removed. & \citet{ruoppa2025evoms} & \citet{ruoppa2025unsupervised} \\
        FOR-InstanceV2 & 11,035 & $\checkmark$ & 0 & ULS, MLS, TLS & Coniferous-dominated boreal and temperate forest, deciduous-dominated boreal and temperate forest, subtropical forest, tropical forest & Dataset spans multiple countries across three continents. Combination of the original FOR-Instance dataset, the dataset of \citet{bai2023semantic}, NIBIO\_MLS, and new data. Some of the semantic annotations have been generated using automated algorithms. & \citet{xiang2025forinstancev2} & \citet{xiang2025forestformer3d} \\ \aboverulesepcolor{lightgray} \midrule
        %\belowrulesepcolor{lightgray}
        \textbf{FGI-EMIT} & \textbf{1,561} && \textbf{3} & \textbf{ALS} & \textbf{Coniferous-dominated, deciduous-dominated and mixed boreal forest, planted trees} & Dataset includes built environment. && \\
        %\aboverulesepcolor{lightgray}
        \bottomrule
    \end{tabular}
    }
    \label{table:dataset_comparison}
\end{table*}

\subsection{Individual tree segmentation}

Individual tree segmentation is one of the most extensively studied point cloud segmentation tasks in forestry. Early approaches generally relied on unsupervised heuristic algorithms, which segmented trees based primarily on geometric properties of the point cloud, sometimes augmented with additional information such as reflectance or RGB color. Following the rapid advances of deep learning across computer vision and remote sensing, more recent works have focused on DL-based ITS approaches.

\subsubsection{Conventional segmentation algorithms}

Conventional point cloud individual tree segmentation algorithms can be divided into two general categories: 2D methods, which operate on a projection of the input, and 3D methods, which process the original point cloud directly. While the vast majority of conventional ITS methods are unsupervised, some utilize simple forms of supervised machine learning. Given the importance of ITS for several downstream applications, the topic has been extensively studied, and a large number of algorithms have been proposed over the years. As such, this section does not aim to provide an exhaustive review, rather, we present a concise overview of the primary methodological directions that have been popular in the literature. Please refer to e.g. \citet{zhen2016trends} for a more detailed review of trends in ITS research and to \citet{cao2023benchmarking} for a comparison of the most highly cited unsupervised ITS algorithms.

In the past, ITS was primarily based on data from passive sensors, such as aerial imagery. Consequently, many 2D approaches for LiDAR point clouds are built on similar principles, including template matching, region growing, and watershed segmentation \citep{zhen2016trends}. Effectively all such methods first transform the input point cloud into a canopy height model (CHM). A common two-step procedure is usually applied \citep[see e.g.][]{hyyppä2001segmentation,popescu2004seeing,koch2006detection}: ground points are first detected using algorithms such as the cloth simulation filter (CSF) \citep{zhang2016easy} or progressive morphological filter \citep{zhang2003progressive}, and the $z$-coordinates are subsequently normalized by subtracting the ground surface. The normalized points are then rasterized into a grid, with each cell assigned the maximum $z$-value of its points. Empty cells are typically populated using, for example, bilinear interpolation, and the CHM is often smoothed to reduce noise with Gaussian \citet{koch2006detection,yu2011predicting}, low-pass \citep{dalponte2016tree}, or mean \citep{silva2016imputation} filters.

Tree detection from the CHM is usually performed by identifying the local maxima, considered as treetops, using a standard maximum filter \citep{hyyppä1999detecting,hyyppä2001segmentation,koch2006detection,silva2016imputation}. Several adaptations of the filter have been proposed for ITS, including variable-size windows \citep{pitkänen2004adaptive,kaartinen2012international}, circular windows \citep{dalponte2016tree} and combinations of the two \citep{popescu2004seeing}. The detected maxima usually serve as seed points for standard image segmentation algorithms, such as region growing \citep{hyyppä1999detecting,hyyppä2001segmentation,dalponte2016tree}, marker-controlled watershed \citep{koch2006detection,yu2011predicting}, or centroidal Voronoi tessellation \citep{silva2016imputation}. Beyond local-maxima approaches, template matching has also been successfully applied to CHM-based tree segmentation \citep{pirotti2010assessing,huo2020individual}.

The range of approaches is considerably broader for 3D ITS methods. One of the most widely adopted 3D strategies is clustering. In an early example, \citet{lee2010adaptive} generated an initial oversegmentation using region growing from automatically detected seed points and then merged the resulting segments into trees using agglomerative hierarchical clustering. \citet{ferraz20123d} proposed a mean-shift-clustering-based segmentation method with three kernel sizes tailored to different forest strata. \citet{ferraz2016lidar} later refined the method into AMS3D, which employs an adaptive kernel size informed by tree size allometry. Clustering-based strategies remain popular in later work. For example, \citet{zhang2024individual} introduced an approach combining hierarchical filtering and clustering.

Several 3D ITS algorithms take advantage of the fact that point clouds can easily be represented as graphs. \citet{strimbu2015graph} constructed a weighted hierarchical graph based on multiple cohesion criteria and partitioned it into connected components to extract individual trees. By contrast, \citet{wang2020unsupervised} transformed the point cloud into a superpoint graph, detected wood points using geometric properties and graph optimization, and finally segmented tree instances by partitioning the graph according to path distances from the identified trunks. More recently, \citet{xi20223d} proposed Treeiso, which applies a two-stage graph clustering procedure and then merges the resulting clusters into tree instances based on several heuristic rules.

Another line of research has focused on hybrid strategies that divide the point cloud into multiple layers, process them in 2D, and then merge the results into 3D tree segments. \citet{wang2008lidar} split the point cloud into horizontal layers and extracted crown contours from each level with a hierarchical-morphology-based algorithm, merging the results into full trees by utilizing graph connectivity. \citet{duncanson2014efficient} instead began with a standard watershed segmentation and divided the resulting segments into multiple layers based on vertical profiles. These layers were then used to generate new CHMs for different canopy strata, which were resegmented with watershed. \citet{ayrey2017layer} introduced layer stacking, which clustered the point cloud at multiple horizontal layers using $k$-means, subsequently constructed an overlap map of the clusters, and extracted trees using a combination of local maxima of the overlap map and clustering. Similarly, building on the methodology of \citet{hyyppä2020undercanopy}, \citet{hakula2023individual} employed DBSCAN at multiple layers together with line fitting for tree detection, followed by a $k$-nearest neighbors classifier to finalize the segmentation.

Many 3D ITS algorithms cannot be explicitly categorized, as they combine multiple strategies and complex heuristic rules. For instance, \citet{li2012new} proposed a sequential growing algorithm that begins from the highest point and expands according to a set of relative spacing rules. Similarly, \citet{hamraz2016robust} introduced a sequential approach in which the global maximum of the point cloud is first identified, and the corresponding tree is then delineated using crown boundaries derived from vertical profiles. \citet{mongus2015efficient} applied marker-based watershed segmentation on a voxelized 3D point cloud, with markers obtained from trunk detection and 2D CHM segmentation. Finally, \citet{burt2019extracting} proposed Treeseg, which employs a combination of several generic point cloud processing techniques, including Euclidean clustering, principal component analysis and shape fitting.

Several works have also explored more novel data sources and sensor fusion. Examples, include multispectral imagery combined with LiDAR \citep{popescu2004seeing}, hyperspectral imagery integrated with LiDAR \citep{qin2022individual}, and multispectral LiDAR, which has been found to enhance the performance of both mean-shift-clustering-based \citep{dai2018new} and template-matching-based \citep{huo2020individual} ITS methods.

\subsubsection{Deep learning-based segmentation approaches}

Following the success of artificial intelligence in computer vision and remote sensing, research on ITS has increasingly shifted toward deep-learning-based models. Compared to conventional unsupervised algorithms, DL-based methods generally achieve higher segmentation accuracy \citep[see e.g.][]{xiang2024automated,xi2025new} and require less extensive hyperparameter tuning. Their main limitation, however, is that effectively all existing approaches are fully supervised, demanding large amounts of manually annotated training data, which is extremely time-consuming to generate. DL-based individual tree segmentation methods can be broadly divided into three categories: rasterization/depth-image-based 2D approaches, hybrid approaches that combine deep learning with heuristic algorithms, and fully 3D-based methods.

Similarly to unsupervised algorithms, 2D DL approaches transform the input point cloud into a two-dimensional representation, perform segmentation, and then propagate the result back to the original 3D data. Several works have explored transforming LiDAR point clouds into depth images in the $xy$-plane and subsequently segmenting them using DL-based image segmentation models. In an early example, \citet{windrim2019forest} performed bounding-box-based instance segmentation on LiDAR-derived raster images using Faster R-CNN \citep{ren2017faster} and subsequently applied a VoxNet-inspired 3D convolutional neural network (CNN) for semantic segmentation of the detected trees. In later work, segmentation performance was further improved with the inclusion of LiDAR reflectance as an input feature \citep{windrim2020detection}.

YOLO \citep{redmon2016you} has been a particularly popular choice for 2D ITS from point clouds. \citet{chang2022twostage} used RandLA-Net \citep{hu2020randlanet} to separate TLS point clouds into tree and non-tree points, followed by YOLOv3 for bounding-box ITS on depth images and a clustering-based refinement step for the resulting 3D segments. Similarly, \citet{sun2022individual} and \citet{jarahizdaeh2025advancing} used YOLOv4 and YOLOv7 for bounding-box-based ITS, applied to depth images derived from ALS and ULS data, respectively. To address the scarcity of annotated training data often encountered in 3D deep learning, \citet{sun2022individual} employed three distinct generative adversarial networks (GANs) to synthesize additional training samples. Somewhat similarly, \citet{straker2023instance} performed ITS from ALS-derived rasters using YOLOv5. Interestingly, they found that downsampling input point clouds to a constant density prior to rasterization slightly improved segmentation accuracy.

Beyond LiDAR point clouds, a substantial amount of research has investigated 2D DL-based ITS from aerial imagery, employing models such as Mask R-CNN \citep{braga2020tree,chadwick2020automatic}, RetinaNet \citep{weinstein2020cross}, and SAM2 \citep{chen2025zero}. The use of multi- and hyperspectral data has also been explored. For example, \citet{dersch2022novel} applied a DETR-based \citep{carion2020end} model for ITS from multispectral UAV imagery, while \citet{long2024scale} proposed a scale pyramid graph network for simultaneous ITS and tree species classification from hyperspectral imagery. Several studies have further integrated LiDAR data into their pipelines: \citet{weinstein2020cross} used image annotations derived from LiDAR point clouds segmented with an unsupervised ITS algorithm for pretraining, and \citet{zhu2025leveraging} leveraged LiDAR points classified as trees by an unsupervised algorithm as auxiliary information for SAM2 prompts.

Hybrid ITS methods integrate deep learning into their pipelines but ultimately rely on heuristic algorithms with user-defined hyperparameters to complete the segmentation. In most cases, deep learning is employed for semantic segmentation of the input point cloud, after which unsupervised algorithms use the classification as a starting point for instance segmentation. A common strategy is to first identify wood or trunk points and then use them to guide tree-level segmentation.

Several works have used PointNet++ \citep{qi2017pointnet++} and modified variants as a part of a hybrid ITS framework. FSCT \citep{krisanski2021forest} employed a PointNet++ \citep{qi2017pointnet++}-based network to classify points into terrain, foliage, trunk, and coarse woody debris. Trunk points were then clustered into tree skeletons, followed by cylinder fitting and sorting. Finally, foliage points were assigned to the nearest trunk to form complete tree instances. Point2Tree \citep{wielgosz2023point2tree} follows a similar paradigm, using PointNet++ to classify trunk points and then applying the graph-based segmentation method of \citet{wilkes2023tls2trees} to generate tree instances. Similarly, \citet{zhu2024synergizing} used a modified PointNet++ to identify wood points in rubber tree plantations, which were then used as root points for graph-based instance segmentation.

Other works have used deep learning for trunk detection from the lower forest strata. \citet{wang2019individual} employed Faster R-CNN to locate tree trunks from 2D projections of rubber plantation point clouds, and used them as seed points for a region-growing-based ITS algorithm. More recently, \citet{ding2025dualbranch} proposed a dual-branch transformer framework that detected trunk instances directly and subsequently employed them to guide a hierarchical $k$-nearest-neighbors classifier for tree-level segmentation.

Another category of hybrid ITS methods employs deep learning for semantic segmentation of the input into tree and non-tree points, after which unsupervised algorithms are applied to the set of tree points. \citet{chen2021individual} used PointNet \citep{qi2017pointnet} to classify points into trees, buildings, and other objects, and subsequently rasterized the tree points and segmented them based on local maxima and gradient information. Similarly, \citet{xia2023study} employed a modified RandLA-Net to semantically segment photogrammetric RGB point clouds, followed by mean shift clustering to delineate tree instances. In the only study to date that has utilized multispectral data for DL-based ITS, \citet{yang2024improved} employed PTv1 \citep{zhao2021point} to extract tree points and then segmented individual trees using a method similar to the watershed algorithm.

The final category of deep-learning-based ITS approaches are 3D methods, which operate directly on the original point clouds and employ a fully DL-based end-to-end framework. While computationally the most demanding, they generally yield the highest segmentation accuracy. A significant amount of research has focused on adapting generic 3D instance segmentation models based on bottom-up instance grouping for forest data. \citet{xiang2023towards} introduced a modified PointGroup \citep{jiang2020pointgroup} model for ITS, adding a feature embedding branch trained with contrastive loss to better separate closely grouped instances, which the original architecture struggled with. Tree instances were then constructed by clustering predicted center offsets concatenated with the feature embeddings. The same framework later served as the basis for both ForAINet \citep{xiang2024automated} and SegmentAnyTree \citep{wielgosz2024segmentanytree}. The former extended the model for panoptic segmentation, while the latter emphasized sensor-agnostic performance through a novel augmentation strategy that randomly downsamples training data.

Like \citet{xiang2023towards}, several other ITS methods also grouped trees into instances based on predicted center offsets. TreeLearn \citep{henrich2024treelearn} predicted center offsets in the $xy$-plane and identified trunk locations using both offset and verticality features, while TreeisoNet \citep{xi2025new} combined four consecutive neural networks, for semantic segmentation, stem base detection, and both 2D and 3D offset prediction. \citet{zhang2023towards} adapted the HAIS \citep{chen2021hierarchical} framework for ITS, employing a PTv1 feature extractor and a loss function tailored to forest data. RsegNet \citep{wang2025rsegnet} followed a similar clustering-based workflow but introduced a novel CosineU-Net feature extractor. \citet{li2025twostage} proposed a two-stage model that integrated treetop locations from depth maps predicted by Mask R-CNN \citep{he2020mask} with center offsets from a 3D U-Net. SPA-Net \citep{zhu2025spanet} replaced the offset branch typical of bottom-up instance segmentation with a sparse geometric proposal module and generated instance predictions based on connected components of graph representations.

Most recently, \citet{xiang2025forestformer3d} introduced ForestFormer3D, a transformer-based model for panoptic segmentation of forest point clouds. Building on OneFormer3D \citep{kolodiazhnyi2024oneformer3d}, the model incorporates several modifications designed to improve ITS accuracy, including ISA-guided query point selection and one-to-many instance matching during training. In contrast to most prior 3D methods, which construct tree instances by clustering predicted center offsets, ForestFormer3D directly predicts instance masks, rendering the model independent of user-defined clustering hyperparameters. 

Recent works have also explored novel architectural innovations in 3D DL-based ITS. \citet{xiu2025individual} proposed 3DPS-Net, a SAM-inspired \citep{kirillov2023segment} architecture that generates tree instance masks from prompt points. These prompts can be either automatically sampled at random or manually provided by the user, with the latter yielding a higher segmentation accuracy. On the other hand, \citet{destouches2025weakly} introduced a rating module that enables weak supervision to refine the initial outputs of an ITS model. While their experiments utilized SegmentAnyTree, the module can, in principle, be integrated into any DL-based ITS framework.

Outside of forestry, a large number of works have addressed the comparatively simpler task of segmenting urban roadside trees from point clouds using deep learning \citep[see e.g.][]{wang2020hierarchical,luo2021individual,jiang2023instance}. However, since roadside trees are typically well separated and arranged in geometrically consistent patterns, these approaches are not directly applicable to the more complex, multi-layered structure of forest point clouds.

\subsection{Multispectral LiDAR}

Multispectral LiDAR, part of the next generation of laser scanning systems, simultaneously acquires measurements at multiple distinct wavelengths. The additional spectral information provided by MS data has been shown to be beneficial in various classification and segmentation tasks \citep{kaasalainen2007toward,kaasalainen2019multispectral}. Its most common applications are found in forestry and ecology \citep{takhtkesha2024multispectral}, where it has been utilized in a wide range of tasks, including tree species classification \citep[see e.g.][]{yu2017single,budei2018identifying,lindeberg2021classification,taher2025multispectral}, forest environment classification \citep{hopkinson2016multisensor}, stem volume estimation \citep{axelsson2023use}, individual tree segmentation \citep{dai2018new,huo2020individual,yang2024improved} and leaf--wood separation \citep[see e.g.][]{li2013separating,howe2015capabilities,li2018utilization}. However, to the best of our knowledge, no previous study has investigated the use of multiple distinct reflectance channels as input features in a fully DL-based ITS framework.

Beyond forestry and ecology, multispectral data has been employed in a variety of remote sensing tasks, such as road mapping \citep{karila2017feasibility}, change detection \citep{matikainen2017object,matikainen2019toward}, and land cover classification. The latter is by far the most widely studied non-forestry application, with methods ranging from unsupervised algorithms \citep[see e.g.][]{wichmann2015evaluating,bakula2016testing} to machine learning \citep{wang2014airborne,teo2017analysis} and deep learning approaches \citep[see e.g.][]{pan2020landcover,li2022agfpnet,zhang2022introducing}, including weakly supervised \citep{chen2024feature} and unsupervised methods \citep{oinonen2024unsupervised,takhtkeshha2024automatic}.

Notably, multispectral features have consistently improved model accuracy over monospectral and geometry-only inputs in several segmentation and classification tasks. Examples include tree species classification in boreal forests \citep{yu2017single,kukkonen2019multispectral,hakula2023individual,taher2025multispectral}, individual tree segmentation \citep{dai2018new,huo2020individual}, land cover classification \citep{wang2014airborne,matikainen2017object,teo2017analysis}, and semantic segmentation of forest data \citep{ruoppa2025unsupervised,takhtkeshha20253d}.

\section{Materials}

This section describes the acquisition, preprocessing, and annotation of \textbf{FGI-EMIT}, our multispectral individual tree segmentation benchmark dataset. Additionally, we detail the data partitioning strategy and the evaluation metrics used to objectively compare the performance of different ITS methods. For further details on how FGI-EMIT should be used for benchmarking purposes, please refer to \appendixref{appendix:utilizing_for_benchmarking}.

\subsection{Study area}

\begin{figure*}[!t]
\centering
\includegraphics[width=\textwidth]{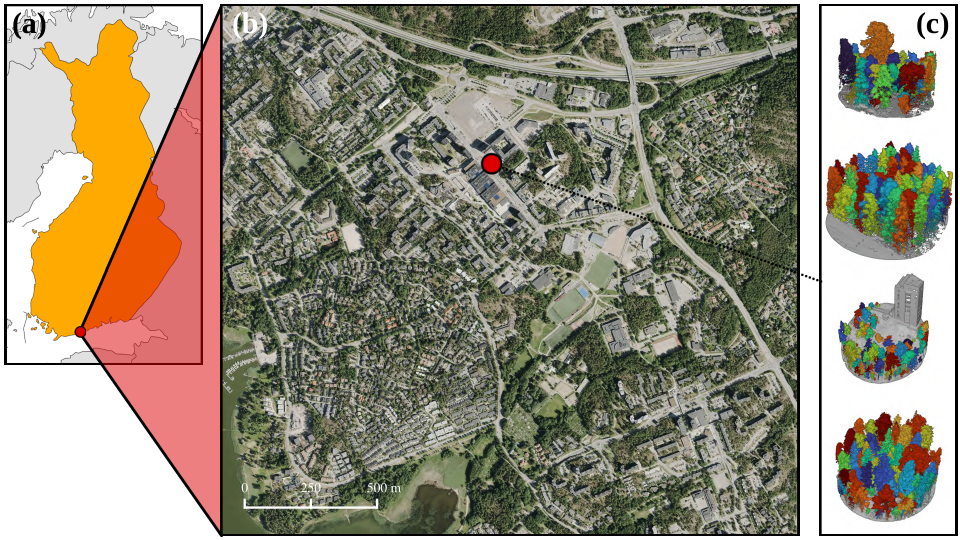}
\caption{Overview of the study area in the Espoonlahti district of Espoo, Finland. (a) Map of Finland with the location of Espoonlahti highlighted. (b) Orthophoto of the Espoonlahti district from summer 2024 (30 cm pixel size). Image obtained from the \citet{espoo2024data}. (c) Examples of test forest plots with manually generated instance annotations.}
\label{figure:study_area_overview}
\end{figure*}

The study area (centered approximately at 60.1462$^\circ$N, 24.6587$^\circ$E) is located in the Espoonlahti district of Espoo, Finland. The area is characterized by diverse levels of built environment, including both high-rise and low-rise residential areas, as well as recreational and non-recreational boreal forests of varying densities. These forests include both natural and planted trees, across more than 20 distinct species \citep{taher2025multispectral}. The forested areas predominantly consist of coniferous-dominated dry and rocky forests and mixed forests with both coniferous and deciduous species. However, small deciduous-dominated regions are also present, particularly along the coastline. This high variability in both tree species and forest types makes the study area ideal for developing an instance segmentation dataset with strong generalization potential across diverse environments. An overview of the study area is shown in \autoref{figure:study_area_overview}.

\subsection{Data acquisition}

FGI-EMIT is based on the same raw dataset as \citet{oinonen2024unsupervised} and \citet{taher2025multispectral}, which was captured using the Finnish Geospatial Research Institute’s (FGI’s) in-house developed laser scanning system HeliALS-TW. The system consists of three separate RIEGL LiDAR scanners (RIEGL Laser Measurement Systems GmbH, Austria), the VUX-1HA, miniVUX-3UAV, and VQ-840-G, referred to as scanners 1, 2, and 3, respectively. The technical specifications of the scanners are summarized in \autoref{table:scanner_specifications}. HeliALS-TW is equipped with a NovAtel ISA-100C inertial measurement unit (IMU), a NovAtel PwrPak7 global navigation satellite system (GNSS) receiver, and a NovAtel GNSS-850 antenna for positioning.

The HeliALS-TW system was mounted to a helicopter and used to survey the study area by flying two perpendicular flight lines at an altitude of approximately 100 meters above ground level and a speed of 14 m/s. Data acquisition was carried out on July 20 and 28, 2023 in leaf-on conditions.

The trajectories of the two flights were computed in Waypoint Inertial Explorer \citep[version 8.90,][]{novatel2022waypoint} using a virtual GNSS base station from the Trimnet service (Geotrim Oy, Finland), located approximately at the center of the study area. The raw LiDAR data from each scanner were georeferenced in RiPROCESS \citep[version 1.9.0,][]{riegl2021riprocess} using the GNSS and IMU measurements. The resulting georeferenced monospectral point clouds from scanners 1, 2, and 3 were then merged into a multispectral point cloud using a Python script with KD-tree-based $k$-nearest neighbors interpolation ($k=1$). If no neighbors from one or more scanners were found within a 0.25 m radius, the respective reflectance fields were left empty. The same procedure was applied to other scanner-specific spectral features (see \appendixref{appendix:additional_point_cloud_attributes}). The final high-density ($>1,000$ points/m$^2$) multispectral point cloud contains reflectance from three distinct wavelengths and the combined geometry from all three scanners.

\begin{table*}[!t]
    \centering
    \caption{Technical specifications of the LiDAR scanners in the multispectral HeliALS-TW system. The beam divergence and diameter of scanner 2 are expressed as two values due to the elliptical shape of its beam. The maximum scanning angle of scanner 3 is expressed as \emph{angle in flight direction} $\times$ \emph{the angle perpendicular to flight direction}. ($\ast$) The scan pattern of scanner 2 is circular. ($\dagger$) The receiver aperture of scanner 3 is 6 mrad. The table has been reproduced based on \citet{oinonen2024unsupervised} and \citet{taher2025multispectral}, where the same system was used.} \bigskip
    \begin{tabular}{lrrr}
    \toprule
    \textbf{Scanner} & \textbf{1} & \textbf{2} & \textbf{3} \\ \midrule \midrule
    \textbf{Model} & VUX-1HA & miniVUX-1DL & VQ-840-G \\
    \textbf{Wavelength (nm)} & 1,550 & 905 & 532 \\
    \textbf{Approximate point density (points/m$^2$)} & 630 & 200 & 420 \\
    \textbf{Maximum number of returns} & 9 & 5 & 5 \\
    \textbf{Maximum scanning angle ($^\circ$)} & 360 & 46$^\ast$ & 28$\times$40 \\
    \textbf{Laser beam divergence (mrad)} & 0.5 & 0.5$\times$1.6 & 1$^\dagger$ \\
    \textbf{Laser beam diameter at ground level (cm)} & 5 & 5$\times$16 & 10 \\
    \textbf{Range accuracy (mm)} & 5 & 15 & 20 \\
    \textbf{Pulse repetition rate (kHz)} & 1,017 & 100 & 200 \\
    \textbf{Scan rate (Hz)} & 143 & 72 & 100 \\
    \bottomrule
    \end{tabular}
    \label{table:scanner_specifications}
\end{table*}

\subsection{Data preprocessing} \label{section:data_preprocessing}

A total of 31 cylindrical plots of varying sizes ($\text{diameter}\in[40\text{ m},60\text{ m}]$) were designated as potential test sites from the raw point cloud. A cylindrical shape was selected because it preserves the vertical structure of trees by avoiding cuts along the $z$-axis, in addition to ensuring computational efficiency of geometric queries \citep{xiang2023towards}. Moreover, the boundaries of cylindrical plots more closely resemble the natural shape of the tree crowns in comparison to, for example, rectangular plots.

The plots were selected by an expert based on aerial imagery, inspection of the point clouds, and on-site visits. The selection aimed to capture a wide variety of forest and tree species types, including varying levels of built environment within the plots. Each plot was assigned a unique integer identifier (ID) between 1001 and 1031. Ultimately, due to time and resource constraints, only 19 of the 31 plots were included in the FGI-EMIT dataset. However, the remaining plots could potentially be added to the dataset if it is expanded in the future.

\begin{figure*}[!t]
\centering
\includegraphics[width=\textwidth]{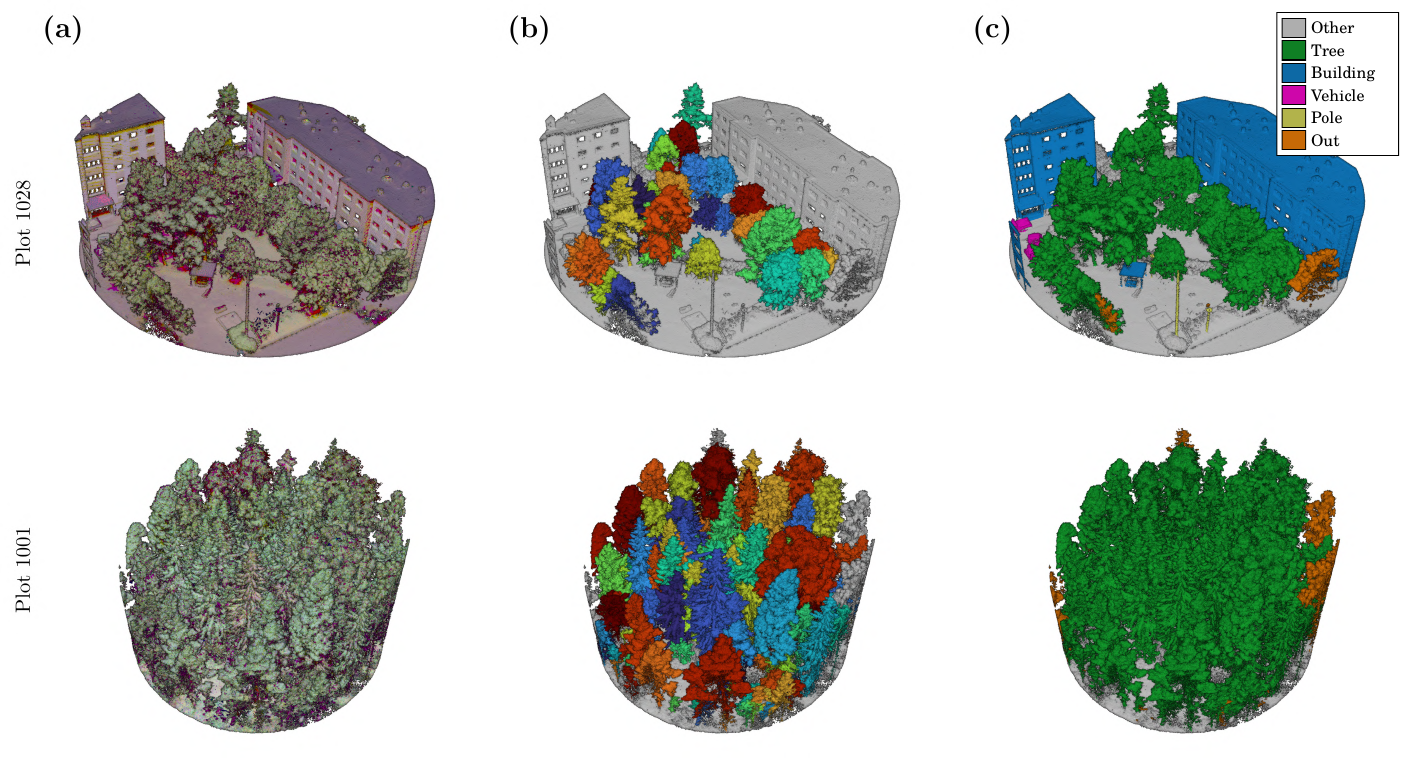}
\caption{Examples of original data and manually generated annotations for two forest plots (IDs 1001 and 1028). (a) Original point cloud with pseudo-colors generated from scaled reflectance values of scanners 1, 2, and 3 assigned to the red, green, and blue channels, respectively. (b) Instance annotations of individual trees, where each tree instance is shown in a distinct color and non-tree points are shown in gray. (c) Semantic annotations of the data, with each class assigned a distinct color.}
\label{figure:plot_annotation_examples}
\end{figure*}

The 19 cylindrical plots were extracted from the raw point clouds, and noise points were subsequently removed using a statistical outlier filter. The filter removed points whose distance to their neighbors differed significantly from the average distance across the point cloud. The filter parameters, i.e. the number of neighbors and the threshold ratio for the standard deviation, were set to 20 and 3, respectively. Examples of preprocessed multispectral point clouds of the cylindrical plots are shown in \autoref{figure:plot_annotation_examples} (a).

\subsubsection{Instance annotations}

For each of the 19 plots, individual trees were manually annotated at the point level using the point cloud processing tool CloudCompare \citep{girardeu2024cloudcompare} and a workflow similar to that described by e.g. \citet{puliti2023forinstance} and \citet{ruoppa2025unsupervised}. All sections of the point cloud that could be definitively identified as a tree and were at least 3 meters in height were segmented into individual instances. Since all plots contained partial trees along their perimeter, only those where $\gtrapprox50\%$ of the tree appeared to be within the plot boundaries were segmented. Examples of the instance annotations are shown in \autoref{figure:plot_annotation_examples} (b).

The manual instance annotation followed a two step process:
\begin{enumerate}
    \item Extract each tree instance from the original point cloud as accurately as possible.
    \item Inspect each extracted tree segment, identify possible errors, and correct them, for example, by merging missing points or removing points not belonging to the tree instance.
\end{enumerate}

In total, manually creating the tree instance annotations required approximately 560 hours of work from a team of two annotators. Adjacent trees with intertwined crowns were separated as accurately as was practically possible. Nevertheless, while the annotations are of high quality, a small number of erroneously segmented points is to be expected due to the inherently complex nature of 3D forest point cloud data. Such minor inaccuracies are common in manually generated forest data annotations, as noted by e.g. \citet{kaijaluoto2022semantic}, \citet{puliti2023forinstance} and \citet{ruoppa2025unsupervised}.

\subsubsection{Semantic annotations} \label{section:semantic_annotations}

In addition to instance annotations, FGI-EMIT includes rudimentary semantic labels for common categories present in the data. Specifically, in addition to trees, man-made objects, including buildings, vehicles and pole-like structures, have each been assigned their own semantic class. Furthermore, partial trees along the plot perimeters that were not given instance labels have been assigned to a separate class. Finally, any points not belonging to the previously described categories, primarily ground and understory vegetation, have been grouped into their own class. The definitions of each semantic category in the dataset are listed in  \autoref{table:semantic_categories} and visual examples are shown in \autoref{figure:plot_annotation_examples} (c).

\begin{table}[!b]
    \centering
    \caption{Descriptions of the semantic categories available in the data.} \bigskip
    \begin{tabular}{lcp{5.6cm}}
        \toprule
        \textbf{Name} & \textbf{ID} & \textbf{Description} \\ \midrule \midrule
        Other & 0 & Points not belonging to any other category. A majority of points in this class are either ground or understory vegetation. \\
        Tree & 1 & Points that are part of any tree instance. \\
        Building & 2 & Buildings and similar smaller man-made structures, such as sheds. This also includes structures attached to buildings, e.g. fences. \\
        Vehicle & 3 & Motorized vehicles, such as cars, trucks and buses. \\
        Pole & 4 & Vertical pole-like structures, such as lamp-posts, traffic signs and utility poles. \\
        Out & 5 & Points belonging to trees that reside mostly outside the plot boundaries and were therefore not assigned to any tree instance. \\
        \bottomrule
    \end{tabular}
    \label{table:semantic_categories}
\end{table}

The primary purpose of the semantic annotations is to facilitate the filtering of objects in categories that could hinder the performance of ITS methods. This issue is most relevant for conventional unsupervised approaches, which often rely on geometry-based heuristics for segmentation and are not designed to distinguish between trees and man-made objects with similar vertical structures, such as utility poles, or in some cases even buildings. In contrast, DL-based methods are generally not affected by this problem, as they can be trained to disregard objects in non-tree classes.

\subsubsection{Height and location computations} \label{section:height_and_location}

An $xy$-location and height were computed for each tree instance based on the point cloud data. Tree location was determined by computing the centroid of a convex hull fitted around an $xy$-plane projection of the crown points. The uppermost 3 meters of each individual tree segment were considered part of the crown.

Tree height was determined as the difference between the highest point in the tree segment and the lowest point of the ground beneath it. The highest point of each tree, $z_{\max,\text{tree}}$, was defined as the maximum $z$-coordinate in the corresponding point cloud. To account for outliers, if the difference in $z$-coordinate between the highest and second-highest points in a tree segment exceeded 0.25 m, the second-highest point was used as $z_{\max,\text{tree}}$ instead. Ground height was estimated by first extracting a circular region with a radius of 0.5 m centered at the tree location from the set of non-tree points in the plot. Subsequently, the ground height $z_{\min,\text{ground}}$ was set to the minimum $z$-coordinate within this region. In the rare case that the circular region contained no points due to occlusions, $z_{\min,\text{ground}}$ was set to the minimum $z$-coordinate of the tree segment, $z_{\min,\text{tree}}$. Finally, tree height was computed as follows:
\begin{equation}
    h_{\text{tree}}=z_{\max,\text{tree}}-\min\left\{z_{\min,\text{ground}},z_{\min,\text{tree}}\right\}.
\end{equation}

\subsubsection{Tree crown categories} \label{section:crown_categories}

\begin{figure*}[!t]
\centering
\includegraphics[width=\textwidth]{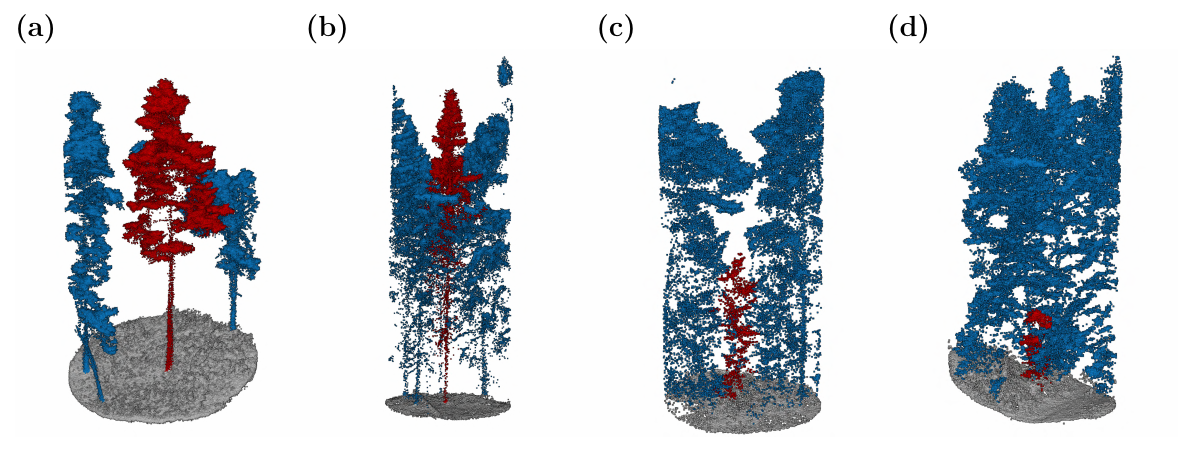}
\caption{Visual examples of trees from each crown category. In (a)--(d), the tree belonging to the corresponding crown category is highlighted in red, while adjacent trees are shown in blue and non-tree points in gray. (a) Example of a tree from crown category A (plot 1013, tree number 12). (b) Example of a tree from crown category B (plot 1001, tree number 12). (c) Example of a tree from crown category C (plot 1022, tree number 168). (d) Example of a tree from crown category D (plot 1018, tree number 34).}
\label{figure:crown_category_examples}
\end{figure*}

To facilitate the analysis of how relative crown positioning affects the accuracy of individual tree segmentation, all tree instances were classified into four distinct crown categories. Following \citet{yu2017single} and \citet{hakula2023individual}, the categories were defined based on height differences and distances relative to neighboring trees. Two trees were considered neighbors if the distance between their locations in the $xy$-plane was less than 3 m. The crown category specifications and corresponding tree counts are listed in \autoref{table:crown_categories}. A visual example of a tree from each category is shown in \autoref{figure:crown_category_examples}. Each tree instance was assigned a category by an automated algorithm that utilized the tree positions and heights computed as described in \autoref{section:height_and_location}. As a result, some trees along the plot perimeters may have been erroneously assigned to a more dominant category, since information about partial neighboring trees located mostly outside the plot boundaries was unavailable.

\begin{table*}[!t]
    \centering
    \caption{Specifications of the four tree crown categories. Two trees were considered neighbors if the distance between them was less than 3 m.}
    \bigskip
    \begin{tabular}{ccp{5cm}p{7.4cm}}
        \toprule
        \textbf{Category} & \textbf{Number of trees} & \textbf{Description} & \textbf{Definition} \\
        \midrule \midrule
        A & 611 (39\%) & Isolated or dominant trees & Tree has no neighboring trees or is $\geq2$ m higher than all neighbors. \\
        B & 308 (20\%) & Group of similar trees & Tree has at least one neighboring tree and is $<2$ m higher than all neighbors. \\
        C & 451 (29\%) & Tree alongside a dominant tree & Tree has at least one neighboring tree which is $\geq2$ m higher and at a distance of $\geq1.5$ m. \\
        D & 191 (12\%) & Tree under a dominant tree & Tree has at least one neighboring tree which is $\geq2$ m higher and at a distance of $<1.5$ m.\\ \bottomrule
    \end{tabular}
    \label{table:crown_categories}
\end{table*}

\subsection{Data usage} \label{section:data_usage}

\subsubsection{Training, validation and test data} \label{section:training_and_test}

The 19 plots were divided into training and test sets, to be used as follows:
\begin{itemize}
    \item \textbf{Training set:} intended for method development. This data can be used, for example, to train ML- and DL-based individual tree segmentation models and/or to optimize hyperparameters of unsupervised segmentation algorithms.
    \item \textbf{Test set:} intended \textbf{only} for evaluating the accuracy of individual tree segmentation methods. To prevent data leakage, the test set must not be used at any stage of model training or hyperparameter optimization.
\end{itemize}

Following common conventions in machine and deep learning, approximately 70\% of the data was assigned to the training set, with the remaining 30\% forming the test set. Specifically, 6 of the 19 plots were randomly assigned to the test set. Due to the substantial variation in plot complexity, we employed stratified random sampling to ensure that both the training and test sets contained all plot types present in the dataset. In practice, the plots were first divided into three groups based on forest density (see \appendixref{appendix:forest_types}) and two plots were then randomly sampled from each group. Plots with IDs 1002, 1004, 1008, 1012, 1018 and 1028 were assigned to the test set. These plots contain a total of 463 individual tree segments, while the remaining 1098 segments comprise the training set. \autoref{table:plot_statistics} presents descriptive statistics summarizing the characteristics of each plot in the dataset. In addition to plot-level statistics, an average value of each metric is provided for both the training and test sets.

\begin{table}[!b]
    \centering
    \caption{Plot level descriptive statistics of each plot in the FGI-EMIT dataset. $n$ trees denotes the number of individual tree instances.}
    \bigskip
    \begin{tabular}{lrrr}
        \toprule
        \textbf{Plot ID} & \textbf{n trees} & \textbf{Area (ha)} & \textbf{Density (n trees/ha)} \\ \midrule \midrule
        \textbf{Training set} &&& \\ \midrule
        1001 & 133 & 0.126 & 1058.380 \\
        1003 & 54 & 0.126 & 429.718 \\
        1005 & 87 & 0.196 & 443.087 \\
        1009 & 103 & 0.196 & 524.575 \\
        1010 & 155 & 0.196 & 789.409 \\
        1013 & 28 & 0.196 & 142.603 \\
        1019 & 8 & 0.283 & 28.294 \\
        1020 & 62 & 0.283 & 219.280 \\
        1022 & 213 & 0.126 & 1695.000 \\
        1023 & 48 & 0.283 & 169.765 \\
        1024 & 62 & 0.196 & 315.763 \\
        1027 & 96 & 0.196 & 488.924 \\
        1031 & 49 & 0.196 & 249.555 \\ 
        \textbf{Average} & \textbf{84.5} & \textbf{0.200} & \textbf{504.181} \\ \midrule \midrule
        \textbf{Test set} &&& \\ \midrule
        1002 & 93 & 0.126 & 740.070 \\
        1004 & 58 & 0.126 & 461.549 \\
        1008 & 34 & 0.196 & 173.161 \\
        1012 & 31 & 0.196 & 157.882 \\
        1018 & 216 & 0.196 & 1100.079 \\
        1028 & 31 & 0.283 & 109.640 \\
        \textbf{Average} & \textbf{77.2} & \textbf{0.187} & \textbf{457.064} \\
        \bottomrule
    \end{tabular}
    \label{table:plot_statistics}
\end{table}

Although FGI-EMIT does not include a designated validation data, a portion of the training data can be set aside for purposes such as model selection, hyperparameter tuning and overfitting prevention. The percentage of training data allocated to validation can be freely determined by the user to suit the particular use case. While creating a separate validation set is not strictly necessary, it is nevertheless strongly recommended, particularly for ML and DL-based approaches.

\subsubsection{Accuracy metrics} \label{section:accuracy_metrics}

To assess the accuracy of individual tree segmentation methods, one-to-one correspondences between the generated tree segments and ground truth instances must first be established. Following common conventions in DL-based point cloud ITS \citep[see e.g.][]{xiang2024automated,wielgosz2024segmentanytree,xi2025new}, predicted instances are matched to the ground truth based on the intersection over union metric. Let $\mathbfcal{P}^{\text{gt}}$ and $\mathbfcal{P}^{\text{pred}}$ denote the sets of ground truth and predicted instances, respectively. The IoU between the $i$th ground truth instance $\mathcal{P}^\text{gt}_i\in\mathbfcal{P}^{\text{gt}}$ and the $j$th predicted segment $\mathcal{P}^\text{pred}_j\in\mathbfcal{P}^{\text{pred}}$ is defined as:
\begin{equation}
    \text{IoU}(\mathcal{P}^\text{gt}_i,\mathcal{P}^\text{pred}_j)=\frac{N^\text{intersection}_{i,j}}{N_i^\text{gt}+N_j^{\text{pred}}-N_{i,j}^{\text{intersection}}},
\end{equation}
where $N_i^\text{gt}$ and $N_j^{\text{pred}}$ denote the number of points in $\mathcal{P}^\text{gt}_i$ and $\mathcal{P}^\text{pred}_j$, respectively, while $N^\text{intersection}_{i,j}$ denotes the number of points in their intersection, that is, the number of points shared between the two point clouds. Following \citet{xiang2023review}, we define an operator that compares a ground truth instance $\mathcal{P}^{\text{gt}}_i$ with all predicted instances and returns the maximum IoU:
\begin{equation}
    \text{maxIoU}(\mathcal{P}^{\text{gt}}_i)=\max_{\mathcal{P}^{\text{pred}}_ j\in\mathbfcal{P}^{\text{pred}}}\left\{\text{IoU}(\mathcal{P}^{\text{gt}}_i,\mathcal{P}^{\text{pred}}_ j)\right\},
\end{equation}

Similarly to \citet{xiang2024automated,henrich2024treelearn,xiang2025forestformer3d}, each ground truth instance $\mathcal{P}^{\text{gt}}_i\in\mathbfcal{P}^{\text{gt}}$ is matched to the predicted instance $\mathcal{P}^{\text{pred}}_j$ that corresponds to $\text{maxIoU}(\mathcal{P}^{\text{gt}}_i)$, provided that the IoU exceeds the threshold $\text{IoU}_{\text{thresh}}=50\%$. Matches with IoU values below the threshold are discarded. Notably, since the IoU-based matches are unique when the threshold is $\geq50\%$ \citep{kirillov2019panoptic}, the matching process is trivial. We note that at lower IoU thresholds, a single predicted instance may be matched to multiple ground truth instances. In such cases, uniqueness must be enforced by retaining only the match with the highest IoU.

All predicted segments for which a matching ground truth instance is identified are considered true positives (TPs), while predicted segments without a corresponding ground truth match are classified as false positives (FPs). Finally, any ground truth instances that remain unmatched are considered false negatives (FNs). To assess the quality of individual tree segmentation, we employ standard accuracy metrics, precision, recall and F1-score, which are defined as follows:
\begin{align}
    &\text{Precision}=\frac{\text{TP}}{\text{TP}+\text{FP}}\\
    &\text{Recall}=\frac{\text{TP}}{\text{TP}+\text{FN}}\\
    &\text{F1-score}=2\cdot\frac{\text{Precision}\cdot\text{Recall}}{\text{Precision}+\text{Recall}}
\end{align}
To evaluate the segmentation performance by crown category, we compute the category-level recall, defined as:
\begin{equation}
    \text{Recall}_{\text{X}}=\frac{\text{TP}_{\text{X}}}{\text{TP}_{\text{X}}+\text{FN}_{\text{X}}},
\end{equation}
where TP$_{\text{X}}$ and FN$_{\text{X}}$ denote the number of true positive and false negative instances, respectively, within a given crown category $\text{X}\in\{\text{A},\text{B},\text{C},\text{D}\}$. Notably, metrics that require the number of false positives cannot be computed at the category level, since unmatched predicted instances have no associated crown category. Finally, following \citep{xiang2024automated,xiang2025forestformer3d}, we compute coverage (Cov) to assess the overall level of alignment between ground truth instances and predicted segments. Coverage is defined as the mean of $\text{maxIoU}(\cdot)$ across the entire set of ground truth instances. Formally:
\begin{equation}
    \text{Cov}=\frac{1}{|\mathbfcal{P}^{\text{gt}}|}\sum_{\mathcal{P}^{\text{gt}}_i \in\mathbfcal{P}^{\text{gt}}}\text{maxIoU}(\mathcal{P}^{\text{gt}}_i),
\end{equation}
where $|\cdot|$ denotes the cardinality of a set. It should be noted that coverage also includes $\text{maxIoU}(\cdot)$ values below the matching threshold $\text{IoU}_{\text{thresh}}$.

In addition to the metrics described above, average precision (AP) is another commonly used measure for evaluating the accuracy of 3D instance segmentation models. AP is defined as the area under the precision-recall curve, which is obtained by computing precision and recall across all distinct confidence thresholds. Following \citet{schult2023mask3d,kolodiazhnyi2024oneformer3d}, we adopt the ScanNet-style AP \citep{dai2017scannet}, which differs slightly in how ground truth and predicted segments are matched compared to the approach used for computing the other metrics. Specifically, each ground truth instance is matched to the highest-confidence prediction with an IoU above the user-defined threshold $\text{IoU}_{\text{thresh}}$. The matching procedure is summarized in \autoref{algorithm:match}. 

Average precision can be computed at different IoU thresholds to assess performance under varying levels of strictness. We denote these variants as AP$_\text{X}$, where X indicates the IoU threshold. For example, AP$_{50}$ refers to the metric computed with $\text{IoU}_{\text{thresh}}=50\%$. Since computing AP requires each predicted instance to be associated with a confidence score, a value rarely provided by conventional unsupervised ITS algorithms, the metric is generally only applicable to deep-learning-based approaches.

\begin{algorithm}[!b]
\caption{Procedure used for matching predicted segments $\mathbfcal{P}^{\text{pred}}$ to ground truth  instances $\mathbfcal{P}^\text{gt}$ when computing average precision. The procedure $\Call{GetPossibleMatches}{\mathcal{P}_i^{\text{gt}},\mathbfcal{P}^{\text{pred}}}$ returns all predicted segments in $\mathbfcal{P}^\text{pred}$ that overlap with the ground truth instance $\mathcal{P}_i^{\text{gt}}$. The resulting set of possible matches, $\mathbfcal{P}_i^{\text{pos-match}}$, is sorted in ascending order of confidence using $\Call{SortByConfidence}{\mathbfcal{P}_i^{\text{pos-match}}}$. The function $\Call{IoU}{\mathcal{P}^\text{gt}_j,\mathcal{P}^\text{pred}_i}$ computes the intersection over union between $\mathcal{P}^\text{gt}_j$ and $\mathcal{P}^\text{pred}_i$. A match is established if the IoU exceeds the user defined threshold IoU$_{\text{thresh}}$, which is set to IoU$_{\text{thresh}}=50\%$ by default.} \label{algorithm:match}
\begin{algorithmic}[1]
    \Procedure{Match}{$\mathbfcal{P}^\text{gt},\mathbfcal{P}^{\text{pred}},\text{IoU}_{\text{thresh}}$}
        \State Matches $\gets\{\}$\Comment{Dictionary for storing matches}
        \ForAll{$\mathcal{P}_i^{\text{gt}}\in\mathbfcal{P}^{\text{gt}}$}
            \State $\mathbfcal{P}_i^{\text{pos-match}}\gets$ \Call{GetPossibleMatches}{$\mathcal{P}_i^{\text{gt}},\mathbfcal{P}^{\text{pred}}$}
            \State $\mathbfcal{P}_i^{\text{pos-match}}\gets$ \Call{SortByConfidence}{$\mathbfcal{P}_i^{\text{pos-match}}$}
            \ForAll{$\mathcal{P}_j^{\text{pred}}\in\mathbfcal{P}_i^{\text{pos-match}}$}
                \If{$\mathcal{P}_j^{\text{pred}}\in\text{Matches}^{\text{keys}}$}
                    \State\Continue
                \EndIf
                \State IoU$_{i,j}\gets$\Call{IoU}{$\mathcal{P}_i^{\text{gt}},\mathcal{P}_j^{\text{pred}}$}
                \If{$\text{IoU}_{i,j}\geq\text{IoU}_{\text{thresh}}$}
                    \State$\text{Matches}\{\mathcal{P}_j^{\text{pred}}\}\gets(\mathcal{P}_i^{\text{gt}},\text{IoU}_{i,j})$
                    \State \textbf{break}
                \EndIf
            \EndFor
        \EndFor
        \State\Return Matches
    \EndProcedure
\end{algorithmic}
\end{algorithm}

\section{Methods}

\subsection{Unsupervised ITS algorithms}

Our performance comparison includes four conventional, unsupervised individual tree segmentation algorithms: watershed \citep{yu2011predicting}, 3D adaptive mean shift \citep{ferraz2016lidar}, layer stacking \citep{ayrey2017layer} and Treeiso \citep{xi20223d}. While selecting the algorithms, the primary goal was to include a diverse range of methodological approaches, spanning both established and more recent methods. In addition, each algorithm's prevalence in relevant literature and availability of source code were key factors for inclusion. 

\subsubsection{Watershed} \label{section:watershed_description}

Watershed-delineation-based segmentation is one of the pioneering algorithms for individual tree segmentation from LiDAR point clouds. Despite its age and relative simplicity, the method remains widely used to this day due to its computational efficiency and relatively strong performance, particularly in less complex forest environments composed mainly of large, well-separated trees. While several variations of the algorithm with minor differences have been proposed \citep[see e.g.][]{koch2006detection,zhang2014hybrid,eysn2015benchmark}, we adopt the implementation described by \citet{yu2011predicting}. A Python implementation of the algorithm was used in all experiments.

As a first step in the segmentation process, the input point cloud is normalized by subtracting the estimated ground elevation from the $z$-coordinates (see \autoref{section:setup_unsupervised_its}). Next, a rectangular grid with a user-defined cell size is overlaid on the point cloud. A canopy height model (CHM) is constructed by assigning each cell the maximum $z$-value of points within it and empty cells are populated using bilinear interpolation. Cells with a height value below 2 meters are classified as background. Finally, a Gaussian filter is applied to the CHM to smooth local irregularities and reduce noise.

Following the creation of the CHM, local maxima are identified by applying a maximum filter and subsequently selecting cells whose values remain unchanged. The detected local maxima are interpreted as treetops and are used as seed points in a marker-controlled watershed transformation. The resulting segments represent delineated individual tree crowns. To form the final individual tree segments, all 3D points under the area of each 2D crown segment are retrieved from the original point cloud.

\subsubsection{3D adaptive mean shift}

3D adaptive mean shift (AMS3D)
\citep{ferraz2016lidar} is an individual tree segmentation algorithm based on mean shift clustering \citep{comaniciu2002meanshift}. AMS3D models the $xyz$-coordinates of a point cloud as a multimodal distribution, where each mode, defined as a local maximum of both density and height, corresponds to a treetop. Mean shift clustering is then applied to detect these tree locations and segment the corresponding points. For our experiments, we utilized an R implementation of the algorithm\footnote{\url{https://rdrr.io/github/niknap/MeanShiftR/}}.

Mean shift is a non-parametric algorithm for detecting local maxima, i.e. modes, of a density function. Given $n\in\integer$ multidimensional samples $\mathbf{x}_i\in\real^d$, the algorithm computes the weighted mean as follows:
\begin{equation}
    m_{h,g}(\mathbf{x})=\frac{\sum_{i=1}^n\mathbf{x}_i g\left(\left|\left|\frac{\mathbf{x}-\mathbf{x}_i}{h}\right|\right|^2\right)}{\sum_{i=1}^ng\left(\left|\left|\frac{\mathbf{x}-\mathbf{x}_i}{h}\right|\right|^2\right)},
\end{equation}
where $||\cdot||$ is the Euclidean norm, $g(\cdot)$ is some kernel function, and $h$ is the kernel bandwidth, a smoothing parameter that determines the contribution of each sample. Starting from an initial estimate $\mathbf{x}^0$, the algorithm iteratively shifts the kernel toward higher-density regions by updating:
\begin{equation}
    \mathbf{x}^{t+1}=m_{h,g}(\mathbf{x}^t),
\end{equation}
until convergence. The final clusters are formed by grouping together all data points that converged to the same mode.

In the context of ITS, each sample $\mathbf{x}_i$ corresponds to a point in the 3D point cloud $\mathcal{P}\in\real^{n\times 3}$, represented as a vector $(x_i,y_i,z_i)$. AMS3D decomposes the kernel function $g(\cdot)$ into two separate kernels for the horizontal and vertical domains:
\begin{equation}
    g\left(\left|\left|\frac{\mathbf{x}-\mathbf{x}_i}{h}\right|\right|^2\right)=g^s\left(\left|\left|\frac{\mathbf{x}^s-\mathbf{x}^s_i}{h^s}\right|\right|^2\right)g^z\left(\left|\left|\frac{\mathbf{x}^z-\mathbf{x}^z_i}{h^z}\right|\right|^2\right),
\end{equation}
where $g^s(\cdot)$, $\mathbf{x}^s$ and $h^s$ are the kernel function, sample coordinates, and bandwidth in the horizontal domain, while $g^z(\cdot)$, $\mathbf{x}^z$, and $h^z$ represent the equivalent quantities in the vertical domain.

To address variations in point density and vegetation structure across forest strata, AMS3D employs an adaptive bandwidth model. The objective is to scale the diameter and height of the 3D kernel to approximately correspond to the mean crown diameter and depth at a given height. Formally, the bandwidth of point $\mathbf{x}_i$ is defined by a linear model:
\begin{align}
    h^s(\mathbf{x}_i)&=s^sz_i\\
    h^z(\mathbf{x}_i)&=s^zz_i,
\end{align}
where $s^s$ and $s^z$ are user-defined slope hyperparameters. In practice, these parameters control the diameter and height of the cylindrical kernel used in mean shift clustering.

\subsubsection{Layer stacking}

Layer stacking \citep{ayrey2017layer} is a hybrid ITS algorithm that divides the input point cloud into multiple layers along the vertical axis and processes them as separate 2D rasters. The segmented layers are subsequently stacked to reconstruct 3D tree segments. A Python port of the official layer stacking R implementation\footnote{\url{https://github.com/eayrey/Layer-Stacking}} was used in all experiments.

The algorithm begins by generating a canopy height model from the input point cloud and detecting local maxima, similarly to the watershed ITS algorithm (see \autoref{section:watershed_description}). The point cloud is then divided into layers at one meter height intervals, with the points in each layer projected onto the $xy$-plane. Each layer is clustered using $k$-means and the CHM-derived local maxima as seed points. Polygonal buffers are placed around the identified clusters, and an overlap map is constructed by stacking the buffered polygons across all layers. Local maxima of the overlap map are detected using a maximum filter and tree cores are constructed by applying a buffer of user-defined width around each maximum.

Following tree core construction, layer stacking performs an additional $k$-means clustering step for each layer. The clustering is performed three times, with each iteration using local maxima from the overlap map obtained at progressively finer resolutions as seed points. The resulting clusters are again buffered into polygons, which are then assigned to all tree cores they overlap with. The algorithm filters potential errors by removing abnormally large polygons and those overlapping with multiple tree cores. The final tree segments are constructed by extracting all points within the validated polygons assigned to each tree core across all layers.

\subsubsection{Treeiso}

A 3D point cloud can be represented as a graph $G=(V,E)$, where the nodes $V$ correspond to individual points and the edges $E$ connect neighboring points. By utilizing this graph representation, Treeiso \citep{xi20223d} models individual tree segmentation as a graph clustering problem, where specific edges are cut to form tree segments. We employed the official Python implementation\footnote{\url{https://github.com/truebelief/artemis_treeiso}} of Treeiso in our experiments.

Treeiso employs $\ell_0$ cut-pursuit \citep{landrieu2017cutpursuit} for graph clustering. The algorithm partitions a graph by cutting edges such that the total variation is minimized. Formally, this corresponds to the following minimization problem:
\begin{equation}
    \min_g\left\{\sum_{i\in V} w^V_i||\mathbf{x}_i-g_i||^2+\lambda\sum_{(i,j)\in E}w^E_{ij}\mathbbm{1}(g_i-g_j\neq 0)\right\},
\end{equation}
where $w^V_i$ is the weight of node $i$, $w^E_{ij}$ is the weight of edge $(i,j)$, $\lambda$ is the regularization strength, $\mathbf{x}_i$ is the value of the node (i.e. its $xyz$-coordinates), $g_i$ is the cluster label assigned to node $i$, and $\mathbbm{1}(\cdot)$ is the indicator function.

Treeiso begins by performing a two-stage graph clustering using the cut-pursuit algorithm. In the first stage, a 3D nearest-neighbor graph is constructed from the point cloud and segmented into small clusters using $\ell_0$ cut-pursuit. The second stage applies cut-pursuit again in 2D, using the $xy$-coordinates of the cluster centroids obtained in the first stage.

Although the resulting clusters are relatively large and often approximately correspond to individual trees, the crown and stem clusters are generally separated. To address this, Treeiso merges clusters based on global properties rather than the local connectivity used in cut-pursuit. Stem clusters are first identified by evaluating the elevation-difference-to-length ratio among neighboring clusters. For each non-stem cluster, a composite index $\rho_{\text{score}}$ is then computed against its neighboring stem clusters. The index is based on vertical overlap ratio, horizontal overlap ratio, point gap, and 2D centroid distance. Non-stem clusters are iteratively merged with the stem cluster corresponding to the highest $\rho_{\text{score}}$, until no non-stem clusters remain. The resulting clusters represent the final individual tree segments.

\subsection{Deep learning models}

Four deep-learning-based individual tree segmentation approaches were included in the performance comparison: YOLOv12 \citep{tian2025yolov12}, representing 2D methods, and SegmentAnyTree \citep{wielgosz2024segmentanytree}, TreeLearn \citep{henrich2024treelearn}, and ForestFormer3D \citep{xiang2025forestformer3d}, representing 3D methods. Similarly to the unsupervised ITS algorithms, these methods were primarily selected based on their prevalence in the literature and the availability of source code. We further ensured that all models were applicable to general forest data rather than being designed for a specific forest type. Since the FGI-EMIT dataset contains only instance-level ground truth, we restricted the comparison to methods that do not require semantic annotations for training.

\subsubsection{YOLOv12}

While several works have applied YOLO models for point cloud ITS \citep{chang2022twostage,sun2022individual,straker2023instance,jarahizdaeh2025advancing}, none have introduced substantial modifications to the base architecture specifically aimed at improving LiDAR-based individual tree segmentation. We therefore adopted the most recent YOLO model, YOLOv12 \citep{tian2025yolov12}, as the performance baseline for 2D approaches. To convert the annotated LiDAR point clouds into depth images and 2D ground truth labels, we employed a preprocessing pipeline similar to that of \citet{straker2023instance}, who trained YOLOv5 for instance segmentation on the FOR-Instance dataset. All experiments were conducted using the official PyTorch implementation\footnote{\url{https://github.com/sunsmarterjie/yolov12}} of YOLOv12.

YOLOv12 is the latest YOLO architecture, and the first to replace the CNN-based architecture of previous iterations with an attention-centric framework. The model achieves state-of-the art performance on several image segmentation benchmarks, while maintaining latency comparable to earlier versions. The two primary innovations introduced by YOLOv12 are area attention, which partitions the feature map into a fixed number of regions to provide a large receptive field without complex operations, and residual efficient layer aggregation networks (R-ELANs), which enhance the stability of the efficient layer aggregation networks employed in previous YOLO models \citep{tian2025yolov12}.

In order to train the model, the input point clouds must first be transformed into images. Similarly to the unsupervised watershed algorithm (see \autoref{section:watershed_description}), the $z$-coordinates are normalized by subtracting the estimated ground elevation, and ground points automatically identified with a cloth simulation filter \citep{zhang2016easy} are removed. The point cloud is then rasterized at a resolution of 0.1 meters, with each raster cell assigned the maximum $z$-coordinate of the points within it. Following \citet{straker2023instance}, the rasterized heights $\mathbf{z}$ are standardized by subtracting the mean $\mu_{\mathbf{z}}$ and dividing by the standard deviation $\sigma_{\mathbf{z}}$:
\begin{equation}
    \bar{\mathbf{z}}_i=\frac{\mathbf{z}-\mu_{\mathbf{z}}}{\sigma_{\mathbf{z}}}.
\end{equation}
Subsequently, the standardized values are scaled to the interval $[0,1]$ as follows:
\begin{equation}
    \hat{\mathbf{z}}=\frac{\bar{\mathbf{z}}-\min\{\mathbf{\bar{\mathbf{z}}}\}}{\max\{\mathbf{\bar{\mathbf{z}}}\}-\min\{\mathbf{\bar{\mathbf{z}}}\}}.
\end{equation}
Finally, pseudo-color images are generated from the normalized height maps using the \emph{inferno} color palette, as in \citet{straker2023instance}.

To generate the 2D instance annotations from the 3D ground truth, we adopted a procedure similar to \citet{straker2023instance}. Each pixel whose maximum $z$-coordinate corresponded to a tree point was first assigned the label of that tree. Empty pixels were then filled by majority vote within a $3\times3$ sliding window. Pixels with an assigned instance label were subsequently grouped into connected components. Components were discarded if their size was smaller than $\max\{3,0.001\cdot N_{\text{tp}}\}$, where $N_{\text{tp}}$ denotes the total number of tree pixels in the raster. Remaining instances were converted into polygons by fitting an alpha shape with parameter $\alpha=0.4$, chosen based on visual inspection across different values. Polygons with an area below $10^{-3}$ were discarded. In cases where polygons overlapped, conflicts were resolved by greedily assigning the overlapping regions to the polygon processed first. If this caused an instance to split into multiple polygons, only the largest polygon by area was retained. Finally, all buildings were assigned into a separate class and given an instance label using the same procedure as trees.

\subsubsection{SegmentAnyTree}

SegmentAnyTree \citep{wielgosz2024segmentanytree} is a sensor-agnostic, deep-learning-based, fully three-dimensional tree instance segmentation model. It is built on the panoptic segmentation network of \citet{xiang2023towards}, which was adapted from the PointGroup \citep{jiang2020pointgroup} architecture. We used the official implementation\footnote{\url{https://github.com/SmartForest-no/SegmentAnyTree}} in all experiments.

The model begins by voxelizing the input point cloud, which is then passed through a 3D U-Net feature encoder employing generalized sparse convolutional layers \citep{choy20194d}. The extracted features are split into three distinct branches: a semantic segmentation branch, a center offset branch and a feature embedding branch. The semantic segmentation head classifies inputs into tree and non-tree points, while the center offset branch predicts a 3D offset vector from each point to its corresponding tree center. The feature embedding branch simply maps points into a 5D embedding space used to distinguish individual tree instances. Outputs from the three branches are combined and clustered into preliminary tree instances using a combination of region growing and mean shift clustering. The candidate instances are then further refined by ScoreNet, which filters and merges them based on their IoU with the ground truth. During inference, SegmentAnyTree applies non-maximum suppression (NMS) on the trees predicted by ScoreNet to remove redundant instances. 

Because forest point clouds are typically too large to fit into GPU memory, SegmentAnyTree randomly samples cylindrical neighborhoods from the input during training. At inference time, the input is processed in regularly spaced overlapping cylinders, and instances split across multiple cylinders are greedily merged based on IoU \citep{xiang2023towards}. 

A notable limitation of DL-based segmentation models is their strong dependence on the training data. In particular, the models often fail to generalize across sensor modalities. For example, a model trained on ALS data tends not to transfer well to TLS data, and vice versa. SegmentAnyTree addresses this limitation by aiming for sensor-agnostic performance. The proposed solution is a novel data augmentation strategy, in which training data is downsampled to multiple densities, specifically 1,000; 500; 100; 75; 50; 25; and 10 points/m$^2$.

\subsubsection{TreeLearn}

TreeLearn \citep{henrich2024treelearn} is a 3D DL method for extracting individual trees from ground-based LiDAR point clouds. The method follows the same general principle as Seg\-ment\-Any\-Tree, where features extracted by a multi-branch neural network backbone are clustered to form tree instances. Although the FGI-EMIT benchmark dataset is ALS-based, its high density ensures that tree trunks are visible in most cases. Consequently, TreeLearn remains applicable to our data, despite generally struggling on sparser ALS point clouds, as noted by \citet{xiang2025forestformer3d}. The official PyTorch implementation of TreeLearn\footnote{\url{https://github.com/ecker-lab/TreeLearn}} was used in all experiments.

Like SegmentAnyTree, TreeLearn begins by voxelizing the input point cloud and then extracts features using a sparse-convolution-based 3D U-Net backbone. The outputs are separated into two branches: one for semantic segmentation and one for predicting the point-wise tree center offset. In contrast to SegmentAnyTree, the model predicts 2D offset vectors in the $xy$-plane only. To obtain clearly separated clusters, TreeLearn restricts clustering to points near tree trunks with lower prediction uncertainty. Points are filtered such that their verticality feature, as defined by \citet{hackel2016contour}, exceeds a user-defined threshold $\tau_{\text{vert}}$, and the $z$-component of their offset prediction differs from the tree base $z$-coordinate (set at 3 m) by at most $\tau_{\text{off}}$. Points that meet these criteria are clustered using HDBSCAN \citep{campello2013density}. Finally, unlabeled points are assigned to clusters based on a majority vote among their nearest neighbors.

TreeLearn processes input point clouds in overlapping rectangular tiles. Notably, since offset predictions near tile edges are often not accurate as the corresponding tree base may fall outside the tile, predictions are restricted to the central region of each tile. At inference time, subsequent to predicting semantic scores and offsets for all tiles and prior to clustering, the predictions are concatenated, and overlapping regions are averaged to reduce artifacts introduced by the tiling.

\subsubsection{ForestFormer3D}

ForestFormer3D \citep{xiang2025forestformer3d} is a transformer-based 3D panoptic segmentation model that represents the current state of the art in individual tree segmentation. The model is adapted from OneFormer3D \citep{kolodiazhnyi2024oneformer3d} with several modifications designed to improve segmentation performance on forest point cloud data. We employed the official implementation of ForestFormer3D\footnote{\url{https://github.com/SmartForest-no/ForestFormer3D}} in all experiments.

Similarly to both SegmentAnyTree and TreeLearn, ForestFormer3D voxelizes the input point cloud and employs a sparse-convolution-based 3D U-Net for feature extraction. The resulting 32-dimensional feature vectors are then used in a novel ISA-guided query point selection strategy. Specifically, the features are split into two branches: one learns 5D instance-discriminative feature vectors, and the other performs semantic classification into tree and non-tree voxels. Farthest point sampling (FPS) is then applied in the 5D embedding space of the tree voxels to select a fixed number of instance query points $K_{\text{ins}}\in\integer^+$. Compared to FPS on the original point cloud, ISA-guided selection achieves higher instance coverage on forest data. In addition, it is computationally lighter than the parametric queries used by some 3D segmentation approaches, and produces queries that are easier to interpret visually.

The $K_{\text{ins}}$ instance queries, together with $K_{\text{sem}}\in\integer^+$ randomly initialized, learnable semantic queries, are passed to the query decoder. The 32D U-Net features serve as keys and values. The decoder, consisting of six transformer layers, outputs $K_{\text{ins}}$ instance masks with confidence scores and $K_{\text{sem}}$ semantic masks. In contrast to previous DL-based ITS methods that cluster embeddings to form tree instances, ForestFormer3D directly predicts masks that correspond to individual trees, eliminating the reliance on clustering hyperparameters. 

Since ForestFormer3D explicitly selects instance queries with known spatial locations, the predicted instance masks are inherently aligned with the ground truth instances, which eliminates the need for optimization-based matching strategies commonly used in transformer-based 3D instance segmentation. In contrast to the one-to-one matching setup of OneFormer3D, ForestFormer3D adopts one-to-many association during training, allowing each ground truth instance to match multiple predicted masks. At inference, duplicate predictions are removed based on confidence scores. This strategy was found to yield higher quality instance predictions in forest data.

Much like SegmentAnyTree, ForestFormer3D samples randomly placed cylindrical regions from the input point clouds during training. At test time, predictions are generated for evenly spaced overlapping cylinders. A score-based merging step is then applied: all predicted masks across the scene are ranked by confidence, and lower-scoring overlapping masks are discarded. Following TreeLearn, uncertain predictions near cylinder edges are also removed.

\subsection{Experimental setup}

All unsupervised individual tree segmentation algorithms, as well as the deep-learning-based models, were evaluated on the test split of FGI-EMIT. To ensure an objective comparison, the FGI-EMIT training set was used for both optimizing the hyperparameters of the unsupervised algorithms and training the DL models. The $xyz$-coordinates of the full multispectral point cloud were used in all experiments. Points belonging to trees that reside mostly outside plot boundaries (class 5, see \autoref{section:semantic_annotations}) were excluded from the data. The specifications of the computing hardware used in all experiments are listed in \autoref{table:computational_hardware}.

\begin{table}[!b]
    \centering
    \caption{Specifications of the computing hardware used in the experiments.} \bigskip
    \begin{tabular}{ll}
        \toprule
        \textbf{Device} & \textbf{Specifications} \\ \midrule \midrule
        CPU & Intel\textsuperscript{{\textregistered}} Xeon\textsuperscript{{\textregistered}} w5-3425 \\
        GPU & NVIDIA\textsuperscript{{\textregistered}} RTX\textsuperscript{\texttrademark} A6000 48 GB GDDR6 \\
        Memory & $8\times64$ GB DDR5 4800 MHz \\ 
        \bottomrule
    \end{tabular}
    \label{table:computational_hardware}
\end{table}

In addition to the accuracy metrics described in \autoref{section:accuracy_metrics}, we measured the runtime of each ITS method, reporting the average segmentation time per test plot in seconds. Although this comparison is not fully objective, since the implementations of the methods are not necessarily fully optimized, it nevertheless provides a practical estimate of the runtime that can be expected when using the publicly available source code for each approach.

\subsubsection{Experimental setup of unsupervised ITS algorithms} \label{section:setup_unsupervised_its}

Each unsupervised individual tree segmentation algorithm is associated with a parameter space $\Theta\subset\real^d$,  where $d\in\integer^+$ denotes the number of hyperparameters. Hyperparameter optimization was modeled as an unconstrained optimization problem:
\begin{equation}
    \max_{\theta\in\Theta}\{f(\theta)\},
\end{equation}
where $\theta$ is a parameter combination in $\Theta$ and $f(\cdot)$ is an objective function mapping the hyperparameters to a measure of segmentation performance. Following common conventions in ITS, we used the F1-score over the training set as the objective. The aim was to identify the optimal parameter combination $\theta^\ast$ that maximizes the F1-score.

In practice, the parameter space $\Theta$ was constrained and discretized by defining a range and step size for each hyperparameter. To provide a comprehensive assessment of performance across different parameter values, we employed relatively wide ranges with fine step sizes. The ranges were chosen based on recommended default parameter values while maintaining physical reasonability. Since overfitting is not a major concern for unsupervised ITS algorithms, no separate validation set was used during optimization. Instead, the entire training set was employed when evaluating performance for a given hyperparameter configuration.

Because the parameter space was extremely large and even conventional unsupervised ITS methods can be relatively computationally demanding on high-density point clouds, exhaustive grid search was not feasible. Consequently, inspired by \citet{wielgosz2023point2tree}, who applied Bayesian optimization (BO) to tune the hyperparameters of the TLS2trees instance segmentation algorithm \citep{wilkes2023tls2trees}, we adopted a Bayesian approach for parameter optimization. Bayesian optimization is a subcategory of sequential model-based optimization (SMBO). In SMBO, an expensive-to-evaluate function $f(\cdot)$ is approximated by a computationally cheaper surrogate model $M(\cdot)$, which is iteratively updated by evaluating $f(\cdot)$ at points determined by an acquisition function $\alpha(\cdot)$ \citep{bergstra2011algorithms}. BO employs Bayes' rule to update $M(\cdot)$ based on evaluation data from $f(\cdot)$. It then updates the acquisition function according to the posterior model and maximizes it to determine the next evaluation point \citep{shahriari2016taking}.

The hyperparameter optimization was conducted using the Python library Optuna \citep{akiba2019optuna}. For BO, we used Optuna's AutoSampler, which dynamically selects the most appropriate sampler for each trial, depending on the parameter space, objective function and number of iterations. AutoSampler primarily relies on two established BO methods: Gaussian processes (GPs) and the tree-structured Parzen estimator (TPE).

Gaussian processes, which can be viewed as a generalization of multivariate normal distributions to infinite dimensions, are one of the most widely used surrogate models in BO. As GPs are inherently probabilistic, the surrogate $M(\cdot)$ is associated with both a mean function $\mu(\cdot)$ and a covariance function $k(\cdot)$, often referred to as the kernel. The GP sampler in Optuna employs a Matérn kernel, formally defined as:
\begin{equation}
    k(\mathbf{x}_i,\mathbf{x}_j)=\frac{1}{\Gamma(\nu)2^{\nu-1}}\left(\frac{\sqrt{2\nu}}{\rho}||\mathbf{x}_i-\mathbf{x}_j||\right)^\nu K_\nu\left(\frac{\sqrt{2\nu}}{\rho}||\mathbf{x}_i-\mathbf{x}_j||\right),
\end{equation}  
where $\Gamma(\cdot)$ is the gamma function, $K_\nu(\cdot)$ is the modified Bessel function of the second kind, $\rho$ is the length scale parameter, and $\nu$ is a smoothness parameter. Notably, GPs are closed under sampling, that is, the posterior of a model with a GP prior is itself also a GP.

In contrast to GPs, which model the posterior directly, the tree-structured Parzen estimator approximates $f(\cdot)$ using non-parametric densities $l(\theta)$ and $g(\theta)$ that correspond to regions of the objective space associated with high and low function values, respectively. TPE employs Bayes' rule with:
\begin{equation}
    p(\theta|y)=\begin{cases}
        l(\theta)&\text{if }y<y^\ast\\
        g(\theta)&\text{if }y\geq y^\ast\\
    \end{cases}
\end{equation}
where $y$ denotes the observation and $y^\ast$ is a threshold set at some percentile $\gamma$ of the observed values, such that $p(y<y^\ast)=\gamma$.

GP and TPE may become trapped in a local optima if the surrogate model fails to accurately approximate the true objective function \citep{bergstra2011algorithms}, which can occur when $f(\cdot)$ is discontinuous. Crucially for the parameter optimization process, only the performance of Treeiso has been shown to depend continuously on its hyperparameters \citep{xi20223d}. The same robustness cannot be assumed for the other algorithms. In fact, \citet{cao2023benchmarking} demonstrated highly discontinuous behavior for AMS3D. Consequently, BO may not reliably identify the optimal hyperparameter combination for watershed, AMS3D, and layer stacking. To address this limitation, we also separately optimized the hyperparameters of each algorithm, except Treeiso, using random search. Random search simply performs an user-defined number of trials using randomly sampled hyperparameter configurations. While it offers no guarantee of finding the global optimum, the probability of doing so increases with the number of trials \citep{bergstra2012random}. The primary goal with employing random search was to corroborate the results obtained with BO: if random search achieves similar or worse segmentation accuracy, we can conclude with high confidence that the hyperparameters found by BO are near-optimal.

Since conventional unsupervised ITS algorithms are generally not designed to handle non-tree objects, points classified as buildings, vehicles, and poles (classes 2, 3, and 4, respectively) were removed during both hyperparameter optimization and evaluation. All tested unsupervised ITS algorithms require height-normalized input data without ground points. Consequently, we applied a cloth simulation filter \citep{zhang2016easy} to automatically detect ground points and normalized the point cloud $z$-coordinates by centering the ground points around $z=0$ m. Prior to evaluation, the predicted tree segments from each algorithm were mapped back to the original unnormalized point clouds. Notably, since Treeiso is specifically designed to operate on tree-only input data \citep{xi20223d}, we performed an additional round of hyperparameter optimization for the algorithm using alternative input data that contained only points manually classified as trees (class 1).

Additional details of the hyperparameter optimization, including the identified optimal parameter values and tested ranges, are provided in \appendixref{appendix:hyperparameter_optimization}.

\subsubsection{Experimental setup of deep learning models} \label{section:dl_model_experimental_setup}

All deep-learning-based individual tree segmentation models were trained from scratch on the training split of the FGI-EMIT data set to ensure optimal performance on the test set. Where pretrained model weights were available, we additionally evaluated the models on the FGI-EMIT test set using these weights to verify that training from scratch had not degraded performance, for example, due to the smaller size of the FGI-EMIT training set compared to what the original models had used. In all cases, models trained from scratch performed comparably or better, demonstrating successful training and convergence. The full, unnormalized forest plot point clouds were used as input. Points originally labeled as classes 2, 3 and 4 were reassigned to class 0 to create a straightforward tree/non-tree semantic classification for model training and evaluation. An exception was made for YOLOv12, where buildings (class 2) were retained as a separate class, as this setup yielded a slight improvement in model accuracy.

Since all tested DL methods have previously been evaluated or trained on the original FOR-Instance dataset, which has a point density equivalent to FGI-EMIT, and have demonstrated excellent performance, the model hyperparameters reported in the respective papers are applicable to our data. Consequently, default model hyperparameter values and data augmentation methods were used. However, because the size of FGI-EMIT differs substantially from both FOR-InstanceV2 and the TreeLearn dataset, batch size and number of training epochs were manually optimized for each model. To prevent overfitting and determine the optimal number of training epochs, three plots from the training set were assigned to a separate validation set using stratified random sampling based on forest density, similarly to the procedure used for creating the test set (see \autoref{section:training_and_test}). Specifically, the validation set consisted of plots with IDs 1019, 1022, and 1031. Validation data was excluded from training and used only to compute loss and accuracy metrics at set intervals. However, we note that validation data is included when performance metrics on the training set are reported. For model-specific hyperparameter values used during training, the interested reader is referred to \appendixref{appendix:dl_model_training}.

By default, all tested DL models use only geometric information as input features. To assess the potential benefit of the multispectral reflectance information available in the FGI-EMIT dataset, all 3D deep learning ITS methods were also trained with reflectance values as auxiliary input features. For this purpose, we adopted the outlier-robust normalization scheme of \citet{takhtkeshha20253d}. Given the reflectance values of channel $i$, denoted by $\mathbf{x}_{\text{reflectance}}^i$, we first computed the interquartile range (IQR):
\begin{equation}
    \text{IQR}(\mathbf{x}_{\text{reflectance}}^i)=Q_{75}(\mathbf{x}_{\text{reflectance}}^i)-Q_{25}(\mathbf{x}_{\text{reflectance}}^i)
\end{equation}
where $Q(\cdot)_j$ denotes the $j$th percentile. The normalized reflectance feature vector $\hat{\mathbf{x}}_{\text{reflectance}}^i$ was subsequently computed as:
\begin{align}
    \bar{\mathbf{x}}_{\text{reflectance}}^i&=\frac{\mathbf{x}_{\text{reflectance}}^i-M(\mathbf{x}_{\text{reflectance}}^i)}{\text{IQR}(\mathbf{x}_{\text{reflectance}}^i)}\\
    \hat{\mathbf{x}}_{\text{reflectance}}^i&=\frac{\bar{\mathbf{x}}_{\text{reflectance}}^i-\min\{\bar{\mathbf{x}}_{\text{reflectance}}^i\}}{\max\{\bar{\mathbf{x}}_{\text{reflectance}}^i\}-\min\{\bar{\mathbf{x}}_{\text{reflectance}}^i\}},
\end{align}
where $M(\cdot)$ is the median of the input.

We also explored alternative normalization strategies, including simple scaling of reflectance values to the $[0,1]$ range and the IQR-based normalization of \citet{ruoppa2025unsupervised}, which subtracts the minimum value to ensure the features start from zero without subsequent scaling. However, both approaches yielded slightly worse segmentation performance in initial experiments, which is why the normalization scheme described above was adopted.

\section{Results} \label{section:results}

\subsection{Performance comparison} \label{section:performance_comparison}

Quantitative performance metrics for all benchmarked individual tree segmentation methods are presented in \autoref{table:performance_comparison_test}, while accuracies of each crown category (A--D, see \autoref{section:crown_categories}) are listed in \autoref{table:crown_ctg_comparison_test}. Results for the alternative Treeiso setup, where input data only included points manually classified as trees, are denoted by $\dagger$. These accuracy metrics are not strictly comparable with the others, since ITS becomes an inherently easier task when components such as understory vegetation are removed. We note that the reported average runtimes for all unsupervised algorithms and YOLOv12 exclude the time required for ground filtering using CSF. For completeness, corresponding performance metrics on the FGI-EMIT training set are provided in \appendixref{appendix:performance_comparison_metrics}.

\begin{table*}[t]
    \centering
    \caption{Comparison of unsupervised individual tree segmentation algorithms and deep-learning-based approaches on the \textbf{test split} of the FGI-EMIT dataset. The best performance metrics are shown in \textbf{bold}, and the second-best are \underline{underlined}. Results marked with $\dagger$ indicate that only points classified as trees were used as input.} \bigskip
    \small{
    \begin{tabular}{lc*{5}{S[table-format=2.1]}S[table-format=1]}
        \toprule
        \textbf{Model} & \textbf{DL} & \textbf{Precision (\%)} & \textbf{Recall (\%)} & \textbf{F1-score (\%)} & \textbf{Cov (\%)} & \textbf{AP$_{\mathbf{50}}$ (\%)} & \parbox{1.8cm}{\centering\textbf{Average time (s/plot)}} \\ \midrule \midrule
        Watershed \citep{yu2011predicting} && 70.8 & 36.7 & 48.4 & 34.6 & \text{-} & \underline{5} \\
        AMS3D \citep{ferraz2016lidar} && 64.8 & 30.2 & 41.2 & 31.5 & \text{-} & 206 \\
        Layer stacking \citep{ayrey2017layer} && 61.4 & 24.4 & 34.9 & 24.5 & \text{-} & 65 \\
        Treeiso \citep{xi20223d} && 54.0 & 44.9 & 49.1 & 44.9 & \text{-} & 89 \\
        Treeiso$^\dagger$ \citep{xi20223d} && 62.4 & 45.6 & 52.7 & 46.8 & \text{-} & 141 \\
        YOLOv12 \citep{tian2025yolov12} & $\checkmark$ & \underline{73.8} & 40.2 & 52.0 & 35.6 & 37.1 & \textbf{3} \\
        SegmentAnyTree \citep{wielgosz2024segmentanytree} & $\checkmark$ & 66.5 & \underline{61.3} & \underline{63.8} & \underline{59.6} & \underline{47.0} & 226 \\
        TreeLearn \citep{henrich2024treelearn} & $\checkmark$ & 63.8 & 60.3 & 62.0 & 58.3 & 33.4 & 101 \\
        ForestFormer3D \citep{xiang2025forestformer3d} & $\checkmark$ & \textbf{78.9} & \textbf{68.5} & \textbf{73.3} & \textbf{64.9} & \textbf{64.3} & 178  \\
        \bottomrule
    \end{tabular}}
    \label{table:performance_comparison_test}
\end{table*}

\begin{table*}[t]
    \centering
    \caption{Comparison of crown category-level recall on the \textbf{test split} of the FGI-EMIT dataset. The best performance metrics are shown in \textbf{bold}, and the second-best are \underline{underlined}. Results marked with $\dagger$ indicate that only points classified as trees were used as input.} \bigskip
    \begin{tabular}{lc*{4}{S[table-format=2.1]}}
        \toprule
        \textbf{Model} & \textbf{DL} & \textbf{Recall$_\text{A}$ (\%)} & \textbf{Recall$_\text{B}$ (\%)} & \textbf{Recall$_\text{C}$ (\%)} & \textbf{Recall$_\text{D}$ (\%)} \\ \midrule \midrule
        Watershed \citep{yu2011predicting} && 74.5 & 20.5 & 2.3 & 0.0 \\
        AMS3D \citep{ferraz2016lidar} && 64.2 & 9.6 & 1.6 & 0.0 \\
        Layer stacking \citep{ayrey2017layer} && 52.0 & 9.6 & 0.0 & 0.0 \\
        Treeiso \citep{xi20223d} && 75.0 & 26.0 & 21.9 & 13.8 \\
        Treeiso$^\dagger$ \citep{xi20223d} && 77.9 & 32.9 & 18.8 & 6.9 \\
        YOLOv12 \citep{tian2025yolov12} & $\checkmark$ & 77.5 & 27.4 & 6.3 & 0.0 \\
        SegmentAnyTree \citep{wielgosz2024segmentanytree} & $\checkmark$ & \underline{90.2} & 43.8 & \underline{40.6} & \underline{27.6} \\
        TreeLearn \citep{henrich2024treelearn} & $\checkmark$ & 84.8 & \textbf{65.8} & 36.7 & 19.0 \\
        ForestFormer3D \citep{xiang2025forestformer3d} & $\checkmark$ & \textbf{94.1} & \underline{56.2} & \textbf{47.7} & \textbf{39.7} \\
        \bottomrule
    \end{tabular}
    \label{table:crown_ctg_comparison_test}
\end{table*}

Based on the quantitative accuracy metrics, DL-based 3D ITS models outperform unsupervised approaches by a significant margin. The best-performing DL model, ForestFormer3D, achieved a test set F1-score 20.6 percentage points (pp) higher than Treeiso$^\dagger$, the most accurate unsupervised algorithm. ForestFormer3D also exceeded the F1-score of the other two 3D DL models, SegmentAnyTree and TreeLearn, by 9.5 and 11.3 pp, respectively, improving upon both precision and recall. Seg\-ment\-Any\-Tree and TreeLearn performed comparably overall, with the former achieving slightly higher precision and recall. In addition, both models attained considerably higher F1-scores than any unsupervised approach. Notably, while their precision values were comparable to those of Watershed, AMS3D, and Treeiso, the recall values were generally 15--20 pp higher. The two models differed more significantly in average precision, with SegmentAnyTree yielding 47.0\%, whereas TreeLearn achieved 33.4\% ($-13.6$ pp), indicating substantially lower prediction uncertainty for SegmentAnyTree. ForestFormer3D achieved the highest AP overall at 64.3\%.

Among the unsupervised algorithms, Treeiso$^\dagger$ (using only tree points as input) achieved the highest test set F1-sore at 52.7\%, followed closely by Treeiso with the full input at 49.1\% ($-2.8$ pp). Perhaps surprisingly, watershed was the next best-performing algorithm with an F1-score of 48.4\%. However, this was primarily due to its relatively high precision, as Treeiso achieved a recall approximately 10 pp higher. While AMS3D and layer stacking attained precision values comparable to both Treeiso configurations, watershed and Treeiso clearly outperformed them in terms of recall and overall F1-score.

YOLOv12, the only 2D DL-based approach included in the benchmark, performed comparably to watershed in terms of quantitative metrics, achieving a slight improvement overall. While its performance was characterized by a relatively high precision of 73.8\%, the second highest among all methods after ForestFormer3D, YOLOv12 achieved only a modest 3.5 pp increase in recall over watershed, and was simultaneously outperformed by both Treeiso configurations. Overall, YOLOv12 was by far the weakest DL-based method, with an F1-score of 52.0\%, approximately 10 pp lower than TreeLearn, the next-best DL model. In fact, even the fully unsupervised Treeiso$^\dagger$ attained a slightly higher F1-score at 52.7\%.

Based on the crown category-level recalls presented in \autoref{table:crown_ctg_comparison_test}, understory trees remain challenging even for DL-based approaches. While the 3D DL models attained higher recall values than the unsupervised algorithms across all crown categories, the most significant differences occured in the two understory categories, C and D. In fact, watershed, AMS3D, and layer stacking detected virtually no trees in either category. The same limitation was observed for YOLOv12, which was expected since understory trees are not visible in the 2D CHM. Treeiso, the only unsupervised algorithm to detect a meaningful number of understory trees, achieved recalls of 21.9\% and 13.8\% in categories C and D, respectively, when using the full input configuration. Even ForestFormer3D, the best-performing model overall, yielded recall values below 50\% in both categories at 47.7\% and 39.7\%, further emphasizing the difficulty of accurately segmenting understory trees.

Even trees in the crown category B, which consists of closely grouped trees of similar size, proved difficult for both the unsupervised algorithms and YOLOv12. Among these methods, Treeiso achieved the highest recall at 32.9\%, whereas AMS3D and layer stacking both detected fewer than 10\% of trees in the category. Recall values in Category B were also surprisingly low for the 3D DL models and remained well below those obtained for category A. Interestingly, while ForestFormer3D was the best performing model in all other crown categories, TreeLearn achieved the highest recall in category B by a considerable margin at 65.8\%. We conjecture that this stems from TreeLearn’s reliance on trunk detection during segmentation, as trunk locations tend to correspond closely to tree positions. Given that trunks are largely visible in our high-density ALS dataset, incorporating them likely aids in discerning closely grouped trees of similar height.

In terms of runtime, the two 2D-based approaches, watershed and YOLOv12, were by far the fastest, with average inference times of 5 and 3 seconds per plot, respectively. Notably, both were approximately 10--70 times faster than all 3D-based methods, highlighting the computational advantage of 2D processing. Among the unsupervised algorithms, layer stacking was the second fastest after watershed, followed by Treeiso and AMS3D. For the DL-based approaches, TreeLearn achieved the shortest runtime, followed by ForestFormer3D and SegmentAnyTree. Although the unsupervised 3D methods were generally faster than their DL counterparts, the difference was not several orders of magnitude. For example, Treeiso, the best-performing unsupervised algorithm, required 89--141 seconds per plot depending on the configuration, compared to 178 seconds for ForestFormer3D.

\begin{figure*}[!t]
\centering
\includegraphics[width=0.98\textwidth]{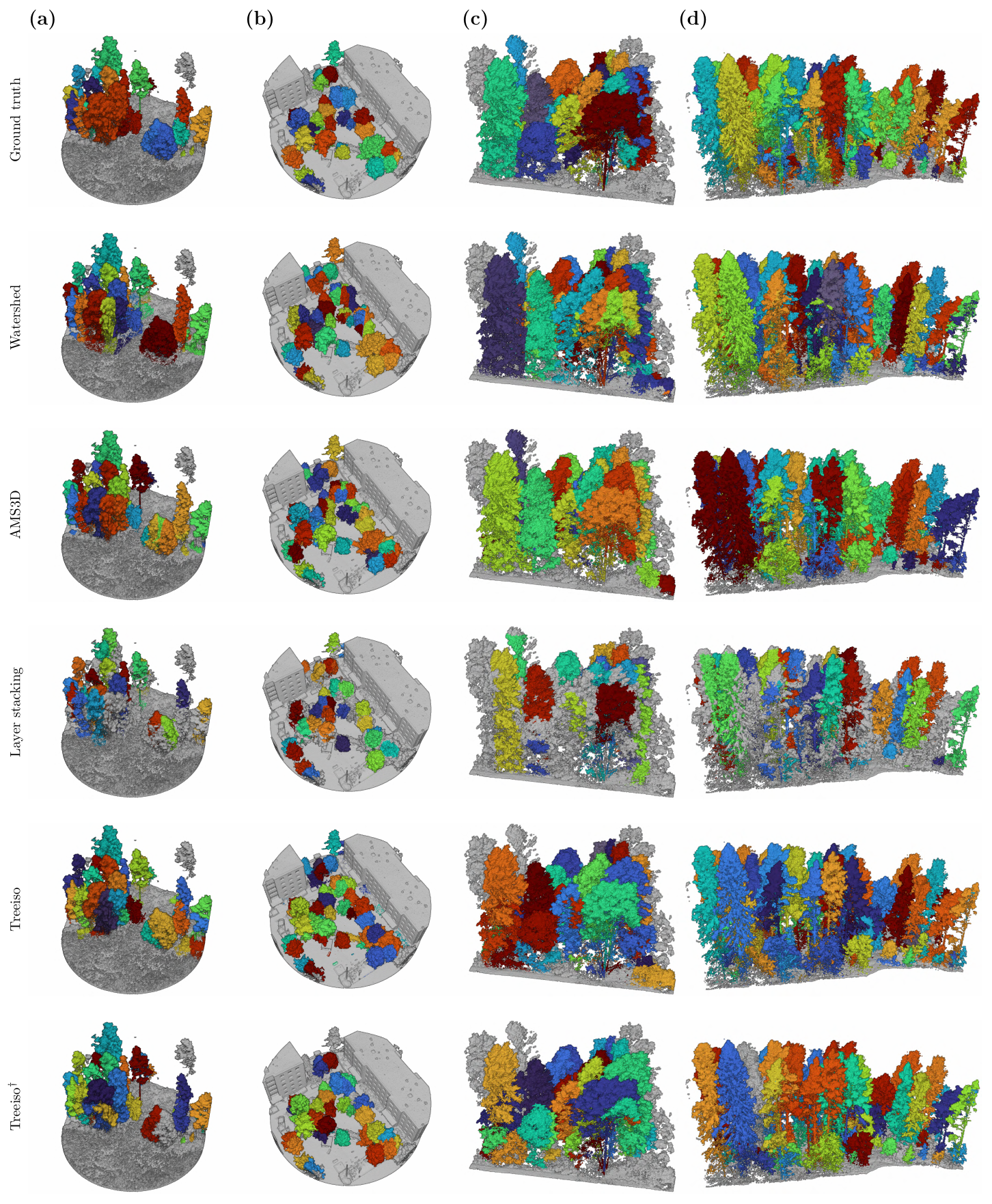}
\caption{Visual comparison of instance predictions from the unsupervised individual tree segmentation algorithms. Predicted tree instances are shown in distinct colors, and non-tree points in gray. (a) Plot 1013. (b) Plot 1028. (c) Subsection of plot 1003. (d) Subsection of plot 1018.}
\label{figure:unsupervised_results_comparison}
\end{figure*}

\begin{figure*}[!t]
\centering
\includegraphics[width=0.98\textwidth]{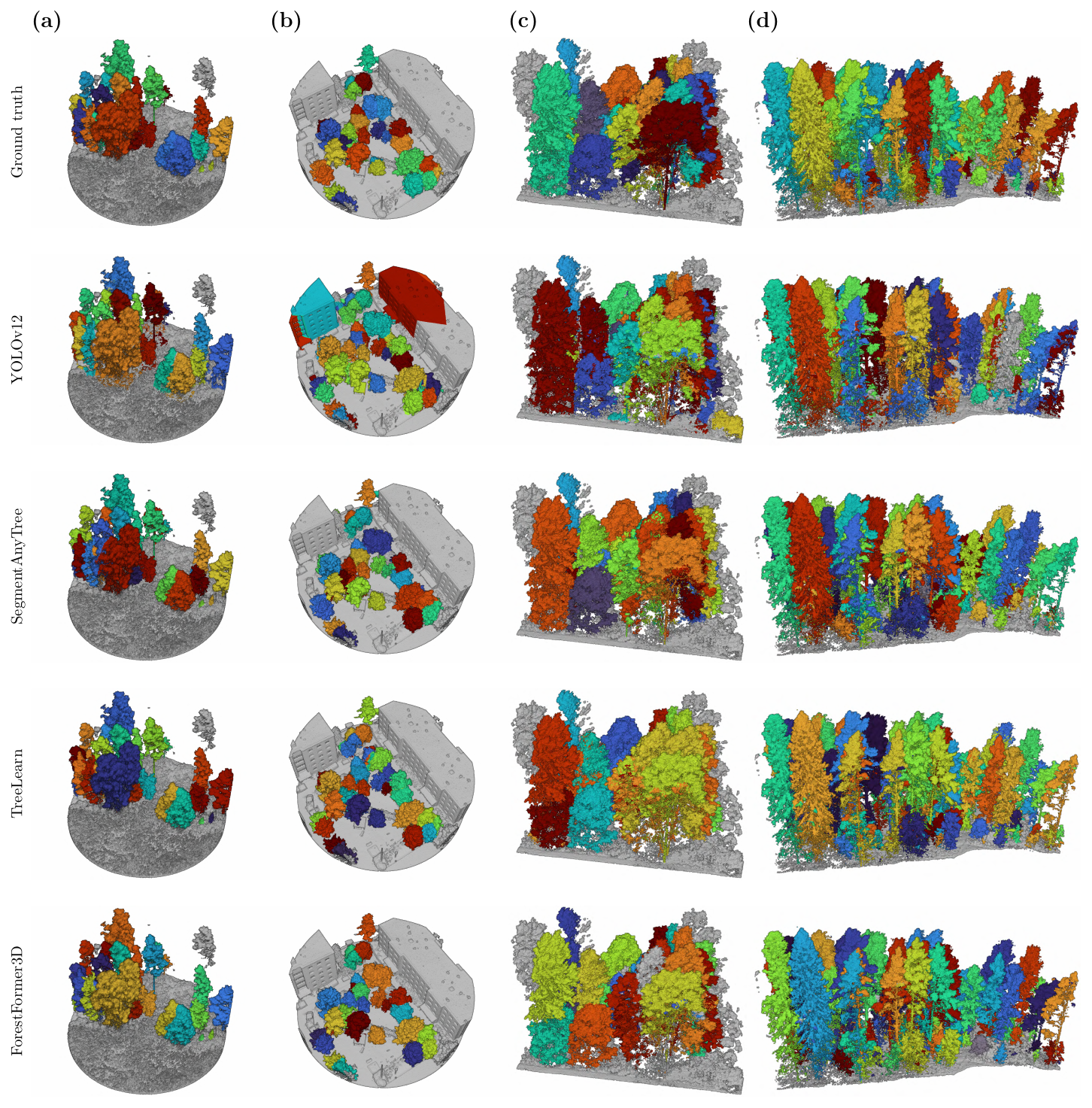}
\caption{Visual comparison of instance predictions from the deep-learning-based individual tree segmentation models. Predicted tree instances are shown in distinct colors, and non-tree points in gray. (a) Plot 1013. (b) Plot 1028. (c) Subsection of plot 1003. (d) Subsection of plot 1018.}
\label{figure:dl_results_comparison}
\end{figure*}

\begin{figure}[!t]
\centering
\includegraphics[width=0.34\textwidth]{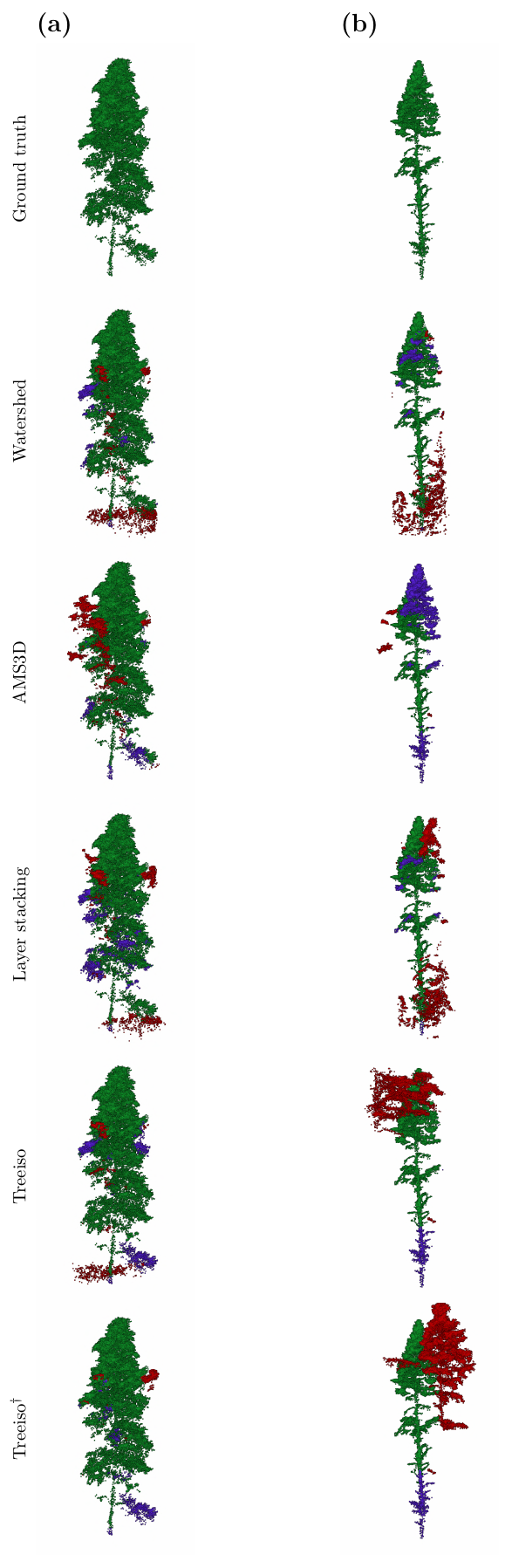}
\caption{Visual comparison of single instance predictions from the unsupervised individual tree segmentation algorithms. True positive points are
shown in green, while false negatives and false positives are shown in purple and red, respectively. (a) Plot 1004, tree number 12. (b) Plot 1018, tree number 142.}
\label{figure:unsupervised_individual_segments_comparison}
\end{figure}

\begin{figure}[!t]
\centering
\includegraphics[width=0.34\textwidth]{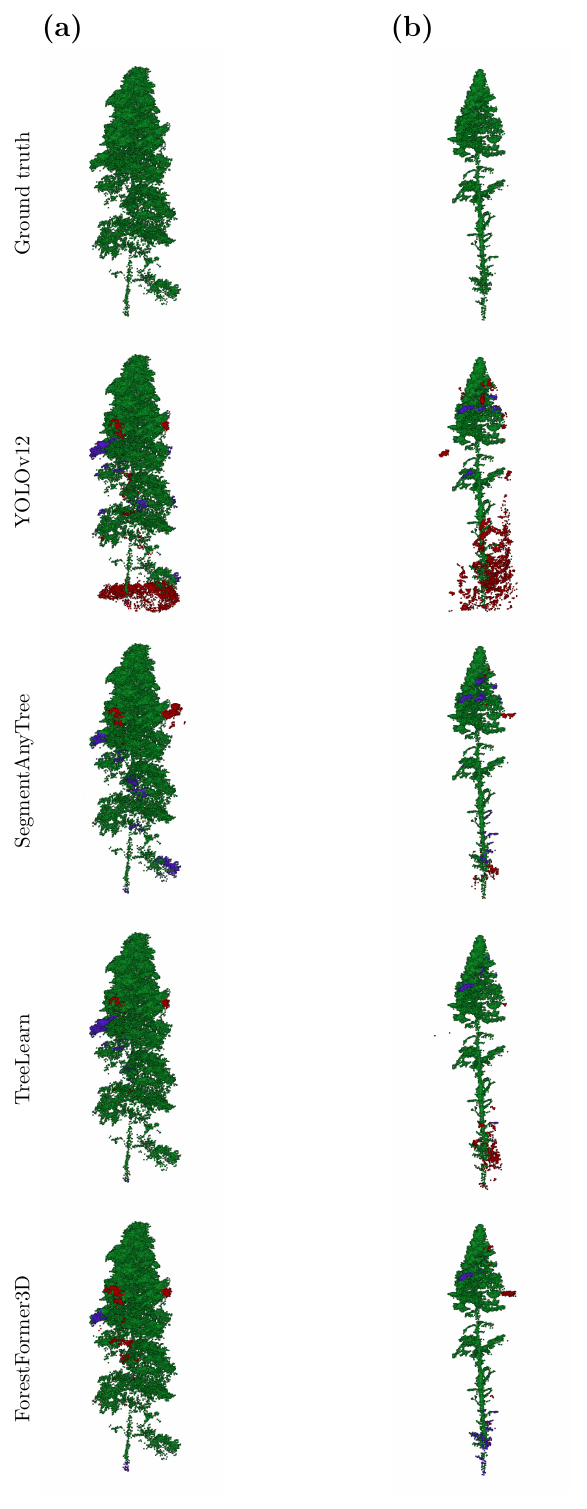}
\caption{Visual comparison of single instance predictions from the deep-learning-based individual tree segmentation models. True positive points are
shown in green, while false negatives and false positives are shown in purple and red, respectively. (a) Plot 1004, tree number 12. (b) Plot 1018, tree number 142.}
\label{figure:dl_individual_segments_comparison}
\end{figure}

\autoref{figure:unsupervised_results_comparison} presents a visual comparison of the unsupervised ITS algorithms, with corresponding visualizations for the DL-based models shown in \autoref{figure:dl_results_comparison}. Since all benchmarked methods generally successfully segment most large trees, visually discerning their shortcomings can be difficult, particularly for the 3D DL-based approaches. Nevertheless, we attempt to highlight the most notable visual differences below.

Based on the visualizations, watershed occasionally produced segments with unnaturally sharp edges, primarily due to the rasterization used when creating the CHM. The segments generated by the other unsupervised algorithms generally appear more natural in shape. Since watershed processes the input point cloud in 2D, the segments of larger trees regularly include understory trees and vegetation. Similarly, even the unsupervised 3D methods frequently merged understory trees with nearby larger trees, although this issue appears less pronounced in the case of Treeiso. Across all unsupervised algorithms, large trees with multiple apexes were often oversegmented, while closely grouped trees of similar height tended to be merged. Owing to its aggressive error filtering, perhaps excessive for high-density ALS data, layer stacking often segmented only the central sections of trees, occasionally erroneously removing valid regions in some layers, partially explaining its low overall accuracy metrics.

Outside of YOLOv12, whose outputs are visually very similar to those of watershed, the DL-based models produced segments that appear largely comparable visually. The main exception is TreeLearn, which clearly struggled more than SegmentAnyTree and ForestFormer3D in dense deciduous plots where tree trunks are not fully visible, as shown in \autoref{figure:dl_results_comparison} (c). Apart from this, all three 3D DL models yielded results that align closely with the ground truth, with ForestFormer3D successfully detecting slightly more understory trees. As is visible in \autoref{figure:dl_results_comparison} (b), while YOLOv12 tended to mistake buildings for trees, the other DL-based approaches have learned to classify them as background, despite the relatively limited amount of man-made structures in FGI-EMIT. It should be noted that behavior similar to YOLOv12 would be expected from the unsupervised algorithms if man-made structures were not excluded from their inputs.

\autoref{figure:unsupervised_individual_segments_comparison} presents examples of individual predicted segments corresponding to the same tree instance from all unsupervised algorithms, with \autoref{figure:dl_individual_segments_comparison} showing the equivalent comparison for the DL-based models. In both figures, (a) corresponds to a relatively isolated tree, and (b) to a tree located within a denser group. As seen in the figures, watershed and YOLOv12 produced strikingly similar results even at the individual-segment level. Among the unsupervised 3D algorithms, the segments generally appear to contain less understory vegetation but exhibit more severe segmentation errors within the crown. In \autoref{figure:unsupervised_individual_segments_comparison} (b), for example, AMS3D has omitted a substantial portion of the crown, while both Treeiso configurations have merged part of a neighboring tree into the segment. However, no consistent pattern in these segmentation errors is apparent. The 3D DL models, SegmentAnyTree, TreeLearn, and ForestFormer3D, clearly produced segments of superior quality compared to the unsupervised algorithms. Overall, the DL-generated segments appear qualitatively excellent, with all three models showing similar minor errors along the crown edges and lower sections of the trees.

\begin{table*}[!t]
    \centering
    \caption{Ablation study on the effects of using mono- and multispectral data. The accuracies are reported for the \textbf{test split} of FGI-EMIT. SAT, TL and FF3D denote SegmentAnyTree \citep{wielgosz2024segmentanytree}, TreeLearn \citep{henrich2024treelearn} and ForestFormer3D \citep{xiang2025forestformer3d}, respectively. For each model, the best performance metrics are shown in \textbf{bold}, and the second-best are \underline{underlined}.} \bigskip
    \scriptsize{
    \begin{tabular}{
        ccc|
        *{3}{S[table-format=2.1]}|
        *{3}{S[table-format=2.1]}|
        *{3}{S[table-format=2.1]}|
        *{3}{S[table-format=2.1]}|
        *{3}{S[table-format=2.1]}
    }
        \toprule
        \multicolumn{3}{c|}{\textbf{Scanners}} &
        \multicolumn{3}{c|}{\textbf{Precision (\%)}} &
        \multicolumn{3}{c|}{\textbf{Recall (\%)}} &
        \multicolumn{3}{c|}{\textbf{F1-score (\%)}} &
        \multicolumn{3}{c|}{\textbf{Cov (\%)}} &
        \multicolumn{3}{c}{\textbf{AP$_{\mathbf{50}}$ (\%)}} \\
        \text{Scanner 1} & \text{Scanner 2} & \text{Scanner 3} & SAT & TL & \text{FF3D} & SAT & TL & \text{FF3D} & SAT & TL & \text{FF3D} & SAT & TL & \text{FF3D}  & SAT & TL & \text{FF3D} \\
        \midrule \midrule
        & & & 66.5 & 63.8 & 78.9 & \underline{61.3} & \underline{60.3} & \textbf{68.5} & \underline{63.8} & \underline{62.0} & \textbf{73.3} & \textbf{59.6} & 58.3 & \textbf{64.9} & 47.0 & \underline{33.4} & \textbf{64.3} \\
        $\checkmark$ & & & 64.7 & \underline{63.9} & 82.0 & 59.4 & 56.2 & \underline{64.1} & 61.9 & 59.8 & \underline{72.0} & 58.4 & 55.5 & \underline{61.6} & 44.2 & 32.7 & \underline{61.4} \\
        & $\checkmark$ & & \textbf{71.1} & 62.9 & 80.1 & \textbf{62.2} & \textbf{61.1} & 63.3 & \textbf{66.4} & \underline{62.0} & 70.7 & 58.6 & 58.5 & 60.9 & \textbf{50.8} & 32.8 & 60.7 \\
        & & $\checkmark$ & 67.0 & \textbf{64.6} & \underline{82.1} & 58.7 & \textbf{61.1} & 61.3 & 62.6 & \textbf{62.8} & 70.2 & 57.3 & \textbf{59.0} & 59.9 & 46.9 & 33.1 & 59.4 \\
        $\checkmark$ & $\checkmark$ & & \underline{67.4} & 62.1 & 80.2 & 59.4 & 57.2 & 60.5 & 63.1 & 59.6 & 69.0 & 57.4 & 56.5 & 59.4 & 47.2 & 31.2 & 58.7 \\
        $\checkmark$ & & $\checkmark$ & 67.2 & 62.6 & 80.6 & 59.0 & 60.0 & 60.9 & 62.8 & 61.3 & 69.4 & 57.7 & 58.1 & 60.1 & 45.4 & 32.2 & 57.8 \\
        & $\checkmark$ & $\checkmark$ & 66.4 & \underline{63.9} & \textbf{88.2} & 55.5 & 60.0 & 55.1 & 60.5 & 61.9 & 67.8 & 55.3 & \underline{58.7} & 52.7 & 44.0 & \textbf{34.3} & 53.8 \\
        $\checkmark$ & $\checkmark$ & $\checkmark$ & 66.0 & \underline{63.9} & 82.0 & 60.7 & 59.0 & 62.0 & 63.2 & 61.3 & 70.6 & \underline{59.0} & 57.5 & 59.9 & \underline{48.0} & 32.3 & 60.2 \\
        \bottomrule
    \end{tabular}
    }
    \label{table:scanner_performance_comparison_test}
\end{table*}

\begin{table*}[!t]
    \centering
    \caption{Ablation study on the effects of using mono- and multispectral data. The crown category-level accuracies are reported for the \textbf{test split} of FGI-EMIT. SAT, TL and FF3D denote SegmentAnyTree \citep{wielgosz2024segmentanytree}, TreeLearn \citep{henrich2024treelearn} and ForestFormer3D \citep{xiang2025forestformer3d}, respectively. For each model, the best performance metrics are shown in \textbf{bold}, and the second-best are \underline{underlined}.} \bigskip
    \small{
    \begin{tabular}{
        ccc|
        *{3}{S[table-format=2.1]}|
        *{3}{S[table-format=2.1]}|
        *{3}{S[table-format=2.1]}|
        *{3}{S[table-format=2.1]}
    }
        \toprule
        \multicolumn{3}{c|}{\textbf{Scanners}} &
        \multicolumn{3}{c|}{\textbf{Recall$_\text{A}$ (\%)}} &
        \multicolumn{3}{c|}{\textbf{Recall$_\text{B}$ (\%)}} &
        \multicolumn{3}{c|}{\textbf{Recall$_\text{C}$ (\%)}} &
        \multicolumn{3}{c}{\textbf{Recall$_\text{D}$ (\%)}} \\
        Scanner 1 & Scanner 2 & Scanner 3 & SAT & TL & \text{FF3D} & SAT & TL & \text{FF3D} & SAT & TL & \text{FF3D} & SAT & TL & \text{FF3D}  \\
        \midrule \midrule
        & & & \underline{90.2} & \textbf{84.8} & \textbf{94.1} & 43.8 & \textbf{65.8} & \underline{56.2} & \textbf{40.6} & 36.7 & \textbf{47.7} & 27.6 & 19.0 & \textbf{39.7} \\
        $\checkmark$ & & & 86.8 & 80.9 & 89.2 & 45.2 & 61.6 & \textbf{57.5} & 36.7 & 32.8 & \underline{43.0} & \underline{31.0} & 13.8 & \underline{31.0} \\
        & $\checkmark$ & & \textbf{91.7} & 81.9 & \underline{91.7} & \underline{46.6} & \underline{64.4} & 50.7 & 37.5 & \underline{43.0} & 42.2 & \textbf{32.8} & \textbf{24.1} & 25.9 \\
        & & $\checkmark$ & 87.3 & 82.8 & \underline{91.7} & \textbf{49.3} & 63.0 & 50.7 & 35.9 & \textbf{43.8} & 38.3 & 20.7 & \underline{20.7} & 20.0 \\
        $\checkmark$ & $\checkmark$ & & \underline{90.2} & 80.9 & 89.2 & 45.2 & 58.9 & 49.3 & 32.8 & 39.1 & 36.0 & 27.6 & 12.1 & 27.6 \\
        $\checkmark$ & & $\checkmark$ & 88.7 & 82.4 & 85.3 & 43.8 & \textbf{65.8} & 53.4 & 36.7 & 39.1 & 39.8 & 22.4 & \underline{20.7} & \underline{31.0} \\
        & $\checkmark$ & $\checkmark$ & 88.7 & \underline{84.3} & 88.2 & 39.7 & 63.0 & 45.2 & 32.0 & 35.9 & 27.3 & 10.3 & \textbf{24.1} & 12.1 \\
        $\checkmark$ & $\checkmark$ & $\checkmark$ & 89.7 & \underline{84.3} & \underline{91.7} & 43.8 & 63.0 & 54.8 & \underline{39.1} & 34.4 & 38.3 & 27.6 & 19.0 & 19.0 \\
        \bottomrule
    \end{tabular}
    }
    \label{table:scanner_crown_ctg_comparison_test}
\end{table*}

\subsection{Effect of multispectral reflectance on the performance of deep learning models} \label{section:multispectral_ablation}

To examine the potential benefits of multispectral data for deep-learning-based individual tree segmentation, we trained all 3D DL models using each reflectance channel individually, as well as all two- and three-channel combinations, as additional input features. To isolate the effects of spectral information from geometric factors, the $xyz$-coordinates of the full multispectral point cloud were used in all experiments. Unsupervised ITS methods were excluded from the comparison, since all benchmarked algorithms were fully geometry-based. Furthermore, previous works have shown that multispectral reflectance can improve the accuracy of conventional ITS algorithms \citep{dai2018new,huo2020individual}, whereas no prior works have explored the use of MS information in the context of DL-based ITS. YOLOv12 was also excluded from the ablation study, since it was substantially outperformed by the other DL models and yielded accuracy metrics comparable to those of watershed. The quantitative accuracy metrics of the ablated models are presented in \autoref{table:scanner_performance_comparison_test}, with the crown category-level results shown in \autoref{table:scanner_crown_ctg_comparison_test}. Corresponding metrics on the FGI-EMIT training set are provided in \appendixref{appendix:multispectral_information}.

Interpreting the results of this ablation study was not straightforward, as segmentation accuracy did not consistently improve with the introduction of additional reflectance features. Moreover, the effect of each reflectance channel combination varied considerably across the tested models, suggesting that while reflectance information can influence the accuracy of DL-based ITS, its impact is highly dependent on the specific model framework.

For both SegmentAnyTree and TreeLearn, the test set F1-scores decreased slightly for most reflectance channel combinations, although the effects were largely negligible. As an exception, a minor improvement in F1-score was observed for SegmentAnyTree when using only reflectance from scanner 2 ($+2.6$ pp). Similarly, TreeLearn achieved a small improvement when using channel 3 ($+0.8$ pp). Notably, multi-channel feature combinations did not improve upon the F1-score of either model when compared to geometry-only inputs.

In contrast to the other two models, the accuracy of ForestFormer3D suffered significantly from the inclusion of reflectance features. Reflectance from scanner 1 had the smallest impact on the test set F1-score ($-3.3$ pp), whereas the worst-performing feature combination, scanners 2 \& 3, reduced recall by 13.4 pp and the F1-score by 5.5 pp compared to the geometry-only case. Although reflectance features consistently improved the precision of ForestFormer3D, the accompanying drop in recall outweighed this benefit, resulting in overall worse F1-scores across all feature combinations.

Interestingly, for TreeLearn, the inclusion of spectral features substantially improved the recall of crown categories C and D in several cases. The effect was particularly pronounced in category C with reflectance from scanner 2 or scanner 3 ($6.3$ pp and $+7.1$ pp, respectively), and in category D with scanner 2 or the combination of scanners 2 \& 3 ($+5.1$ pp in both cases). By contrast, in categories A and B, depending on the feature combination reflectance either had negligible impact or even slightly reduced the recall. These results suggest that, given an appropriate segmentation framework, spectral information may be particularly useful for detecting smaller understory trees, where geometry is less reliable.

\subsection{Effect of point density on segmentation performance}

Effectively all DL-based ITS models included in our benchmark were originally developed for either high-density ALS, MLS or TLS data, whereas many conventional ITS algorithms were designed for much sparser ALS data, typically with densities of 10--100 points/m$^2$. In contrast, the average point density of FGI-EMIT across all plots is $\sim$1,660 points/m$^2$. Consequently, the DL-based models may have disproportionately benefited from the high point density of FGI-EMIT, whereas unsupervised methods could, in principle, perform comparatively better at lower point densities.

To assess the robustness of the benchmarked ITS methods, we generated multiple artificially sparsified versions of FGI-EMIT at varying point densities and used them to evaluate a subset of the benchmarked approaches. Specifically, we selected the two best-performing methods among both the unsupervised algorithms and the DL-based models: watershed, Treeiso, SegmentAnyTree and ForestFormer3D. Following \citet{wielgosz2024segmentanytree}, the sparsified datasets were created by randomly subsampling the original point clouds to target densities of 1,000; 500; 100; 75; 50; 25; and 10 points/m$^2$. Since decreasing the point density can reduce the number of points in smaller understory trees to the extent that detecting them automatically becomes unrealistic, the sparsified datasets were further filtered to remove such instances. For all densities between 1,000 and 10 points/m$^2$, ground truth instances containing fewer than five points were reclassified as background. The number of tree instances left in the training and test sets subsequent to filtering at each density is reported in \autoref{table:trees_after_filtering}.

\begin{table}[!b]
    \centering
    \caption{Number of tree instances remaining in the training and test sets of FGI-EMIT after downsampling the point clouds and discarding very small ground truth instances that are too sparse to be detected automatically. The threshold for removal was 5 points.}
    \bigskip
    \begin{tabular}{l*{2}{S[
        table-format=4.0,
        group-separator={,},
        group-minimum-digits=4,
        group-digits=integer
    ]}}
        \toprule \parbox{2cm}{\centering\textbf{Point density (points/m$^2$)}}
         & \parbox{1.8cm}{\centering\textbf{Training set (n trees)}} & \parbox{1.4cm}{\centering\textbf{Test set (n trees)}} \\ \midrule \midrule
        Original & 1098 & 463 \\
        1,000 & 1098 & 463 \\
        500 & 1098 & 463 \\
        100 & 1094 & 461 \\
        75 & 1088 & 461 \\
        50 & 1080 & 458 \\
        25 & 1045 & 432 \\
        10 & 940 & 385 \\
        \bottomrule
    \end{tabular}
    \label{table:trees_after_filtering}
\end{table}

\begin{figure*}[!t]
\centering
\includegraphics[width=\textwidth]{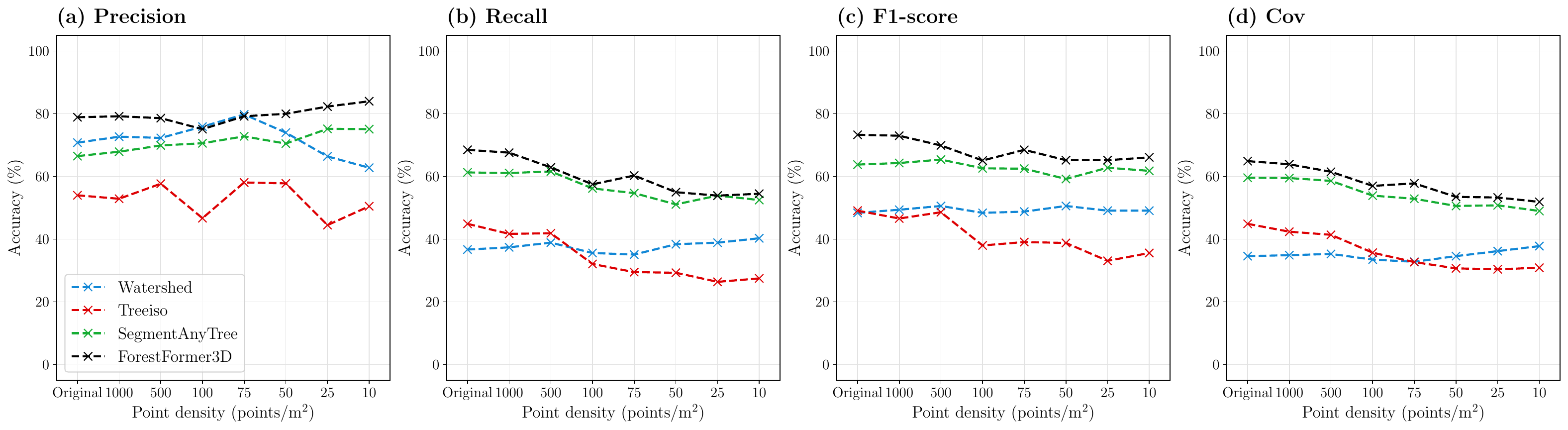}
\caption{Comparison of accuracy metrics on the FGI-EMIT test set for watershed \citep{yu2011predicting}, Treeiso \citep{xi20223d}, SegmentAnyTree \citep{wielgosz2024segmentanytree}, and ForestFormer3D \citep{xiang2025forestformer3d} as a function of approximate point density. Note the non-uniform scale on the $x$-axis.}
\label{figure:density_effects_comparison}
\end{figure*}

The accuracy metrics of the compared ITS methods on the FGI-EMIT test set are shown in \autoref{figure:density_effects_comparison}. To attain the reported accuracies, the hyperparameters of watershed and Treeiso were reoptimized separately for each point density between 1,000 and 10 points/m$^2$, following the procedure described in \autoref{section:setup_unsupervised_its}. Similarly, ForestFormer3D was retrained from scratch at each density. In contrast, for SegmentAnyTree, the original weights were used, as the model is designed to be inherently sensor-agnostic and was trained using data from all point densities considered in the comparison.

The precision of watershed increased steadily until a density of 75 points/m$^2$, after which it declined rapidly, while recall remained largely stable, showing a slight increase at lower point densities. As a result, the F1-score of the algorithm remained approximately constant across all densities, indicating that watershed is highly robust to variations in point density. By contrast, the precision of Treeiso was relatively unstable, with recall, F1-score and coverage all decreasing consistently as point density was reduced. However, this behavior was expected, as Treeiso was originally designed for TLS data, and its graph-based approach depends heavily on consistent point cloud geometry. For both SegmentAnyTree and ForestFormer3D, precision increased, while recall decreased as point density was reduced, leading to an overall stable F1-score. Notably, the decline in recall was more significant for ForestFormer3D, and the two models yielded similar recall values at lower point densities. This suggests that the transformer-based ForestFormer3D may benefit more substantially from high point density and consistent geometry.

While the recall of watershed was in fact higher at 10 points/m$^2$ than at the original density, this increase can be attributed to the removal of a large proportion of small understory trees from the ground truth, which watershed generally fails to segment even at higher densities. Another noteworthy observation is that the recall of watershed and SegmentAnyTree improved slightly at 1,000 and 500 points/m$^2$, even before the number of ground truth trees began to decrease. This improvement primarily occurs because predicted segments with IoU values close to the 50\% matching threshold are more likely to result in successful matches at these slightly reduced point densities, as the criterion effectively becomes less strict when the number of points is reduced slightly.

Crucially, the F1-scores of both DL-based approaches remained significantly higher than those of the unsupervised algorithms across all point densities. These results indicate that, although DL-based ITS methods benefit more from high-density point clouds than some conventional approaches such as watershed, their segmentation accuracy remains superior even at very low point densities.

\subsection{Effect of hyperparameter optimization}

As described in \autoref{section:setup_unsupervised_its}, all algorithms whose performance does not continuously depend on their hyperparameters were optimized using both random search and Bayesian optimization. In all cases, the two methods produced virtually identical results, validating the reliability of the Bayesian approach.

To further assess the impact of hyperparameter optimization on the segmentation accuracy of the unsupervised ITS algorithms, each method was also evaluated using its default parameter values (see \appendixref{appendix:hyperparameter_optimization}). The corresponding comparison of quantitative accuracy metrics is presented in \autoref{table:parameter_optimization_effects}. As can be seen from the table, the adopted hyperparameter optimization strategy improved most accuracy metrics across all benchmarked algorithms. Since the F1-score was used as the objective function for optimization, the precision of watershed and AMS3D decreased slightly. However, these reductions were offset by corresponding gains in recall, resulting in an overall increase in F1-score.

Performance optimization proved most beneficial for algorithms with a larger number of hyperparameters, yielding the greatest improvements in F1-score for layer stacking and Treeiso, both of which have six tunable parameters. In contrast, the gains were more modest for watershed, which has four hyperparameters, and minimal for AMS3D, with only two parameters, where the F1-score increased by just 0.8 percentage points. Notably, watershed outperformed Treeiso when using default parameters, while the opposite was true after optimization, which further highlights the importance of comprehensive hyperparameter optimization for accurate and fair benchmarking of ITS methods.

\begin{table*}[!t]
    \centering
    \caption{Comparison of unsupervised individual tree segmentation methods using default hyperparameter values on the \textbf{test split} of the FGI-EMIT dataset. The relative improvement or decline yielded by parameter optimization is shown in parentheses. Results marked with $\dagger$ indicate that only points classified as trees were used as input.} \bigskip
    \scriptsize{
    \begin{tabular}{l *{4}{S[table-format=2.1]@{}l}| *{4}{S[table-format=2.1]@{}l}}
    \toprule
    \textbf{Model} 
      & \multicolumn{2}{c}{\textbf{Precision (\%)}} 
      & \multicolumn{2}{c}{\textbf{Recall (\%)}} 
      & \multicolumn{2}{c}{\textbf{F1-score (\%)}} 
      & \multicolumn{2}{c|}{\textbf{Cov (\%)}} 
      & \multicolumn{2}{c}{\textbf{Recall$_\text{A}$ (\%)}} 
      & \multicolumn{2}{c}{\textbf{Recall$_\text{B}$ (\%)}} 
      & \multicolumn{2}{c}{\textbf{Recall$_\text{C}$ (\%)}} 
      & \multicolumn{2}{c}{\textbf{Recall$_\text{D}$ (\%)}} \\
    \midrule \midrule
    Watershed \citep{yu2011predicting} 
      & 72.4 & { (\textcolor{Red}{$-1.6$})} 
      & 32.8 & { (\textcolor{Green}{$+3.9$})} 
      & 45.2 & { (\textcolor{Green}{$+3.2$})} 
      & 31.5 & { (\textcolor{Green}{$+3.1$})} 
      & 68.1 & { (\textcolor{Green}{$+6.4$})} 
      & 15.1 & { (\textcolor{Green}{$+5.4$})} 
      & 1.6  & { (\textcolor{Green}{$+0.7$})} 
      & 0.0  & { (\textcolor{Gray}{$\pm0.0$})} \\
    AMS3D \citep{ferraz2016lidar} 
      & 70.4 & { (\textcolor{Red}{$-5.6$})} 
      & 28.3 & { (\textcolor{Green}{$+1.9$})} 
      & 40.4 & { (\textcolor{Green}{$+0.8$})} 
      & 29.4  & { (\textcolor{Green}{$+2.1$})} 
      & 60.3 & { (\textcolor{Green}{$+3.9$})} 
      & 8.2 & { (\textcolor{Green}{$+1.4$})} 
      & 1.6 & { (\textcolor{Gray}{$\pm0.0$})} 
      & 0.0 & { (\textcolor{Gray}{$\pm0.0$})} \\
    Layer stacking \citep{ayrey2017layer} 
      & 23.4 & { (\textcolor{Green}{$+38.0$})}
      & 13.6 & { (\textcolor{Green}{$+10.8$})}
      & 17.2 & { (\textcolor{Green}{$+17.7$})}
      & 19.7  & { (\textcolor{Green}{$+4.8$})}
      & 26.5 & { (\textcolor{Green}{$+25.5$})}
      & 11.0 & { (\textcolor{Red}{$-1.4$})}
      & 0.8 & { (\textcolor{Red}{$-0.8$})}
      & 0.0 & { (\textcolor{Gray}{$\pm0.0$})} \\
    Treeiso \citep{xi20223d} 
      & 52.9 & { (\textcolor{Green}{$+1.1$})} 
      & 35.4 & { (\textcolor{Green}{$+9.5$})} 
      & 42.4 & { (\textcolor{Green}{$+6.7$})} 
      & 38.4 & { (\textcolor{Green}{$+6.5$})} 
      & 63.2 & { (\textcolor{Green}{$+11.8$})} 
      & 23.3 & { (\textcolor{Green}{$+2.7$})} 
      & 13.3 & { (\textcolor{Green}{$+8.6$})} 
      & 1.7  & { (\textcolor{Green}{$+12.1$})} \\
    Treeiso$^\dagger$ \citep{xi20223d} 
      & 61.9 & { (\textcolor{Green}{$+0.5$})} 
      & 39.3 & { (\textcolor{Green}{$+6.3$})} 
      & 48.1 & { (\textcolor{Green}{$+4.6$})} 
      & 42.6 & { (\textcolor{Green}{$+4.2$})} 
      & 66.2 & { (\textcolor{Green}{$+11.7$})} 
      & 26.0 & { (\textcolor{Green}{$+6.9$})} 
      & 19.5 & { (\textcolor{Red}{$-0.7$})} 
      & 5.2 & { (\textcolor{Green}{$+1.7$})} \\
    \bottomrule
    \end{tabular}
    }
    \label{table:parameter_optimization_effects}
\end{table*}

\section{Discussion}

\subsection{Comparison to previous work}

The ITS accuracy metrics observed on FGI-EMIT were generally much lower than those reported in prior literature, particularly for unsupervised algorithms. For example, \citet{nemmaoui2024benchmarking} reported F1-scores between 79.8\% and 81.6\% for the 2D CHM-based ITS algorithm of \citet{dalponte2016tree}, which is comparable to watershed. Similarly, \citet{saeed2024performance} achieved F1-scores of 65.9\% and 66.9\% for watershed and AMS3D, respectively. These discrepancies are primarily due to differences in evaluation methodology: most earlier studies matched predicted and reference trees solely based on position and height, disregarding segment quality completely. Even those employing 2D IoU for matching have typically reported higher accuracies: \citet{cao2023benchmarking} obtained F1-scores of 20--45\% for small trees and 45--70\% for larger trees using AMS3D, while \citet{aubrykientz2019comparative} detected 73.8\% and 55.4\% of trees with AMS3D and a watershed-based algorithm, respectively. In contrast, \citet{cherlet2024benchmarking}, who employed 3D IoU for matching, reported substantially lower accuracies on their TLS dataset, with an F1-score of 33.0\% for Treeiso, 16.1 pp below our result on FGI-EMIT, and 53.5\% for TreeLearn, compared to 62.0\% in our study. To illustrate the impact of the matching criterion, when switching from 3D IoU to the Hausdorff-distance-based positional matching of \citet{yu2006change}, the F1-scores on the FGI-EMIT test set increased to 56.3\% ($+7.9$ pp) for watershed and 64.6\% ($+15.5$ pp) for Treeiso.

When comparing our results to studies focused on DL-based ITS, most of which also employ 3D IoU for matching, we found them largely consistent with prior work. For example, \citet{xiang2025forestformer3d} reported F1-scores of 82.8\% and 72.4\% for ForestFormer3D and ForAINet, respectively, on the FOR-InstanceV2 test set. This mirrors the performance gap observed on FGI-EMIT between ForestFormer3D and SegmentAnyTree, which employs the same base architecture as ForAINet. Similarly, F1-scores of 85.1\% and 68.2\% were reported for ForAINet and Treeiso, respectively, on the original FOR-Instance dataset \citep{xiang2024automated}. Notably, while relative performance differences between models remained consistent across datasets, the absolute accuracy metrics reported on FGI-EMIT were generally lower. This suggests that FGI-EMIT represents a more challenging benchmark, likely due to its emphasis on small understory trees.

At the level of crown categories, several works have reported that accuracies tend to be considerably lower for small trees when using unsupervised approaches \citep[see e.g.][]{cao2023benchmarking,fraser2025quantifying}, which aligns with our findings on FGI-EMIT. While DL-based methods detected substantially more trees in categories C and D, their overall recall for understory trees remained relatively low, consistent with the results of \citet{xiang2024automated} and \citet{wielgosz2024segmentanytree}. Notably, there was no single dominant cause for the missed detections in the understory categories. Instead, we identified at least three recurring factors:
\begin{enumerate}
    \item The understory tree was merged with a nearby dominant tree. Especially common in category D.
    \item The understory tree was erroneously classified as low vegetation and thus not detected. In some cases, the opposite occurred, where low vegetation was mistakenly classified as a tree.
    \item A considerable amount of surrounding low vegetation was merged with the understory tree, resulting in a failed ground truth match.
\end{enumerate}

The effect of the 3D IoU-based matching is also evident in the crown category-level results when compared to earlier studies. For example, \citet{hakula2023individual}, who used the same crown category definitions as in our work, reported accuracies of 86.4\%, 75.2\%, 20.1\% and 2.5\% for categories A, B, C and D, respectively, when using watershed with position-based matching. By contrast, the corresponding metrics attained by watershed on FGI-EMIT were only 74.5\%, 20.5\%, 2.3\% and 0.0\%, respectively, when using 3D IoU for matching. This comparison further highlights how position-based matching can significantly overestimate ITS accuracy, particularly for the more challenging understory categories.

Outside of non-robust approaches for matching predictions to ground truth, two dataset characteristics limit the comparability of ITS benchmark studies: forest type and scanner type. Forest structure and species composition vary substantially across climate zones, which may influence the accuracy of ITS methods. Consequently, while our results should be broadly representative of boreal forests, they may not directly generalize to highly different forest types, such as tropical rainforests, although similar results could be expected in forests with comparable vertical structure. For example, \citet{saeed2024performance} reported that the method of \citet{dalponte2016tree} outperformed AMS3D on ALS data from temperate coniferous-dominated forests, which is consistent with our result of watershed yielding a higher segmentation accuracy than AMS3D. In contrast, \citet{cao2023benchmarking} observed the opposite trend in ALS data from deciduous-dominated temperate and tropical forests.

Scanner type can influence ITS accuracy as strongly as forest type, with certain approaches benefiting substantially from the higher point densities typical of TLS or MLS data. For example, since TreeLearn struggles on sparser ALS data, \citet{xiang2025forestformer3d} reported that ForAINet outperformed it by 21.8 pp in F1-score on the test split of FOR-InstanceV2. In contrast, the corresponding difference between TreeLearn and Seg\-ment\-Any\-Tree was only 1.8 pp on FGI-EMIT, which consists exclusively of high-density ALS data. Moreover, TreeLearn achieves significantly higher accuracy on dense point clouds and even outperformed ForestFormer3D on the PLS-based LAUTx dataset \citep{xiang2025forestformer3d}. These discrepancies driven by dataset characteristics further emphasize the importance of large-scale benchmark datasets spanning diverse forest types and scanner modalities for developing accurate and robust ITS methods.

\subsection{Further analysis on the effects of multispectral reflectance}

\begin{figure}[!t]
\centering
\includegraphics[width=0.45\textwidth]{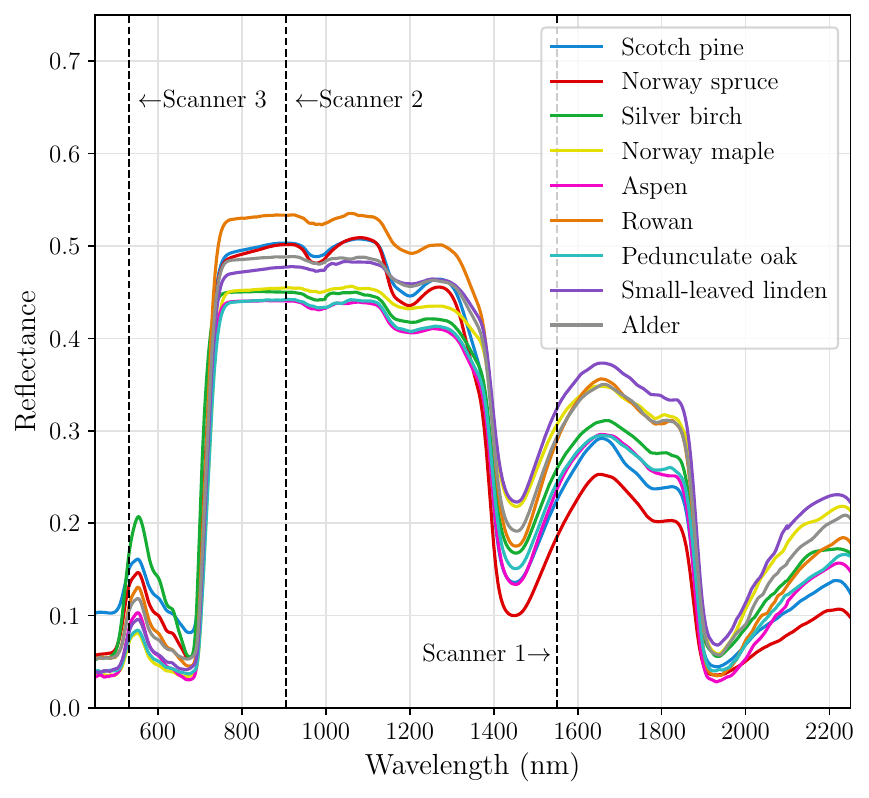}
\caption{Characteristic reflectance spectra of foliage for nine common tree species in the FGI-EMIT data set, with the wavelengths corresponding to scanners 1, 2, and 3 highlighted. Data obtained from \citet{hovi2017spectral}.}
\label{figure:tree_spectra}
\end{figure}

The theoretical rationale for incorporating multispectral reflectance as auxiliary features in individual tree segmentation models is straightforward: different tree species exhibit distinct reflectance characteristics across wavelengths. Furthermore, these interspecies differences vary between wavelengths, thus utilizing multiple channels should facilitate the delineation of individual trees, particularly in mixed forests. To illustrate this, \autoref{figure:tree_spectra} presents the characteristic reflectance spectra of nine species common in the Espoonlahti area \citep{taher2025multispectral}, and thus represented in the FGI-EMIT data set. Specifically, these species are Scotch pine (\emph{Pinus sylvestris} L.), Norway spruce (\emph{Picea abies} (L.) H. Karst.), Silver birch (\emph{Betula pendula} Roth), Norway maple (\emph{Acer platanoides} L.), Aspen (\emph{Populus tremula} L.), Rowan (\emph{Sorbus aucuparia} L.), Pedunculate oak (\emph{Quercus robur} L.), Small-leaved linden (\emph{Tilia cordata} Mill.), and Alder (\emph{Alnus glutinosa} (L.) Gaertn.). As an example, at the wavelength corresponding to scanner 2 (905 nm), a clear difference can be observed between the reflectance of the coniferous (Scotch pine and Norway spruce) and deciduous species.

Despite FGI-EMIT containing a significant amount of mixed forest, the ablation study presented in \autoref{section:multispectral_ablation} showed that incorporating spectral features as additional input for DL-based ITS models yielded effects ranging from modestly positive to clearly detrimental, depending on the model and feature combination. The finding that reflectance channels do not consistently improve segmentation accuracy is in line with previous research concerned with semantic segmentation of multispectral high-density ALS data, where several works have reported that introducing spectral features does not necessarily translate to improved accuracy across all classes \citep{oinonen2024unsupervised,ruoppa2025unsupervised,takhtkeshha20253d}.

In the two cases where accuracy improvements were observed, plausible explanations can be identified. For SegmentAnyTree, the benefit from including reflectance from scanner 2 likely stems from the difference in reflectance between coniferous and deciduous species, which is most pronounced at its corresponding wavelength (905 nm), as shown in \autoref{figure:tree_spectra}. Similarly, while the difference is less significant at the wavelength of scanner 3 (532 nm), it is likely partially compensated for by the higher point density (420 points/m$^2$) compared to scanner 2 (200 points/m$^2$). Although point density was kept constant across all ablated models, the greater number of points carrying true reflectance information in the original multispectral point cloud generally yields more robust spectral features. The difference in the best-performing reflectance combination between SegmentAnyTree and TreeLearn can be primarily attributed to architectural differences between the two models, considering how minor the positive effect of reflectance was in both cases.

Despite previous studies reporting that incorporating multispectral data improved the accuracy of ITS when using conventional approaches \citep{dai2018new,huo2020individual}, we did not observe similar behavior with 3D DL-based methods. We conjecture that there are two likely contributing factors. First, as noted by \citet{ruoppa2025unsupervised}, directly providing reflectance values as input features to 3D DL segmentation networks may not be optimal, since these architectures are typically designed for geometry-only inputs (and occasionally RGB color) and may therefore be unable to fully exploit reflectance information. Second, both \cite{dai2018new} and \cite{huo2020individual} employed multispectral datasets with point densities below 100 points/m$^2$, which is substantially lower than the $\sim1,660$ points/m$^2$ of FGI-EMIT. It is therefore possible that at higher point densities, geometric features dominate over reflectance information, leading to multispectral data having a negligible or even detrimental impact on tree instance segmentation, despite evidence that MS features can improve semantic segmentation and tree species classification performance at comparable densities \citep[see e.g.][]{hakula2023individual,ruoppa2025unsupervised,taher2025multispectral}.

Notably, improvements in segmentation accuracy from the inclusion of reflectance features were only observed for the two clustering-based approaches, SegmentAnyTree and TreeLearn, whereas the accuracy of the transformer-based ForestFormer3D decreased across all setups. Given that the overall effect on accuracy remained largely negligible, the results suggest that clustering-based methods may underutilize reflectance information. By contrast, the transformer-based ForestFormer3D appears to rely too heavily on reflectance features, to the extent that they become detrimental in situations where geometric information would be more reliable.

\subsection{Prospects for future research}

The FGI-EMIT dataset opens several promising avenues for future research, particularly regarding the use of multispectral reflectance in individual tree segmentation. Firstly, since existing DL-based ITS models struggle to effectively utilize reflectance as input features, future work should consider alternative strategies for integrating spectral information. Potential approaches include learnable normalization and scaling methods to mitigate the observed issues with under- or over-utilization of reflectance features, and the dynamic gating strategy proposed by \citet{owen2025pointstowood}, which enables models to selectively utilize reflectance data when beneficial. On the other hand, given that multispectral reflectance was found to improve semantic segmentation accuracy in forest point clouds, extending this to panoptic segmentation presents an interesting research direction. The primary obstacle in this context would be manually generating semantic annotations for FGI-EMIT, since semantically labeling forest data is both extremely challenging and time-consuming, particularly when it comes to separating wood and foliage points. Finally, because prior studies demonstrating the benefits of multispectral reflectance in ITS \citep{dai2018new,huo2020individual} employed significantly sparser data than FGI-EMIT, the geometric information may have dominated spectral features in our case, as previously noted. Consequently, future work should explore the use of multispectral reflectance in sparsified data. If spectral features can offset the loss of geometric detail in low-density point clouds, it may be possible to achieve segmentation performance comparable to high-density ALS data substantially faster and with fewer computational resources.

\section{Conclusions}

In this study, we introduced FGI-EMIT, the first large-scale multispectral airborne laser scanning benchmark dataset for individual tree segmentation. The dataset consists of 1,561 manually annotated trees from various boreal forest types, with a particular emphasis on small understory trees, which remain a major challenge for existing ITS methods. To enhance applicability in urban environments, FGI-EMIT also includes built environment, making it the first ITS benchmark dataset to do so.

Using FGI-EMIT, we compared the performance of four conventional unsupervised algorithms, watershed, AMS3D, layer stacking and Treeiso, and four supervised deep learning approaches, YOLOv12, SegmentAnyTree, TreeLearn and ForestFormer3D. For a fair comparison, optimal hyperparameter values were determined for the unsupervised algorithms using a Bayesian approach, while the DL models were trained from scratch. Among the unsupervised methods, Treeiso achieved the highest test set F1-score of 52.7\%, followed by watershed at 48.4\%. The 3D deep-learning-based approaches proved substantially more accurate, with ForestFormer3D attaining an F1-score of 73.3\% and SegmentAnyTree reaching 63.8\%. The only 2D DL model, YOLOv12, performed comparably to Treeiso with an F1-score of 52.0\%. While 3D DL models outperformed unsupervised algorithms across all crown categories, the most significant difference was observed in understory trees, where ForestFormer3D exceeded Treeiso by 25.9 percentage points in the most challenging category.

An ablation study assessing the effects of incorporating multispectral reflectance as auxiliary input features for 3D DL-based ITS models showed that, while single channel reflectance can marginally improve accuracy in certain cases, particularly for understory trees, existing DL approaches are generally unable to effectively leverage spectral information. The resulting impact ranged from slightly positive to detrimental depending on the model and channel combination. Furthermore, by evaluating the two best-performing unsupervised algorithms and DL models at varying point densities, we demonstrated that although DL approaches benefit more from high-density point clouds, their accuracy remains consistently superior even at 10 points/m$^2$.

This study demonstrated that while deep-learning-based approaches outperform conventional ITS algorithms by a significant margin, small understory trees remain extremely challenging, even for state-of-the-art methods. Moreover, current DL frameworks are unable to replicate improvements previously observed in conventional approaches when incorporating multispectral reflectance. To support future benchmarking efforts and method development for multispectral point clouds, the FGI-EMIT benchmark dataset is made publicly available to the research community.

\section*{CRediT authorship contribution statement}

\textbf{Lassi Ruoppa:} Writing - Original Draft, Conceptualization, Methodology, Software, Investigation, Data Curation, Visualization, Supervision. \textbf{Tarmo Hietala:} Writing - Review \& Editing, Methodology, Software, Investigation, Data Curation. \textbf{Verneri Seppänen:} Writing - Review \& Editing, Methodology, Software, Investigation. \textbf{Josef Taher:} Writing - Review \& Editing, Conceptualization, Data Curation, Resources. \textbf{Xiaowei Yu:} Writing - Review \& Editing, Methodology. \textbf{Teemu Hakala:} Writing - Review \& Editing, Investigation, Data Curation. \textbf{Antero Kukko:} Writing - Review \& Editing, Investigation, Data Curation, Resources,  Funding Acquisition. \textbf{Harri Kaartinen:} Writing - Review \& Editing, Investigation, Data Curation, Resources. \textbf{Juha Hyyppä:} Writing - Review \& Editing, Supervision, Funding Acquisition.

\section*{Acknowledgments}

The study has received funding from the Research Council of Finland (RCF) and the NextGenerationEU instrument through the following grants: ``Collecting accurate individual tree information for harvester operation decision making'' (RCF 359554), ``High-performance computing allowing high-accuracy country-level individual tree carbon sink and biodiversity mapping'' (RCF 359203) and ``Digital technologies, risk management solutions and tools for mitigating forest disturbances'' (RCF 353264, NextGenerationEU). The work was done under RCF Research Flagship ``Forest-Human-Machine Interplay -- Building Resilience, Redefining Value Networks and Enabling Meaningful Experiences'' (RCF 359175) using RCF research Infrastuctures ``Measuring Spatiotemporal Changes in Forest Ecosystem'' (346382) and ``Hydro-RI-platform'' (RCF 346162, NextGenerationEU). The work was supported by the Ministry of Agriculture and Forestry grant ``Future Forest Information System at Individual Tree Level 2.0'' (VA-MMM-2024-25-1).

%% References
\scriptsize{\bibliography{library}}

%% Appendices
\clearpage
\normalsize
\begin{appendices}
\counterwithin{figure}{section}
\counterwithin{table}{section}

\section{Dataset details}

This appendix contains further information on the FGI-EMIT dataset to supplement Sections \ref{section:data_preprocessing} and \ref{section:data_usage}.

\subsection{Forest types} \label{appendix:forest_types}

All 19 plots in the dataset can broadly be categorized as either boreal forest, planted urban vegetation, or a combination of the two. In an effort to provide a more detailed characterization of the different point cloud types in the dataset, the plots were further classified into forest types based on three attributes: \emph{tree species distribution}, \emph{forest density} and \emph{understory}. The first attribute describes the distribution of tree types within the plot, specifically, whether the trees are predominantly coniferous or deciduous. Forest density reflects both the number of trees per hectare and how closely intertwined the canopies are. Finally, the understory attribute quantifies the amount of trees growing below the canopy, corresponding to crown categories C and D.

Each attribute has three categories. For tree species distribution, the categories were defined as \emph{coniferous dominated}, \emph{deciduous dominated} and \emph{mixed}. A plot was classified as either coniferous or deciduous dominated if $\geq70\%$ of the trees were of the respective type. The remaining plots were classified as mixed. Since explicit species information was not available, the distribution was estimated based on visual inspection of the data.

\begin{table}[!b]
    \centering
    \caption{Classification of each plot for each of three forest type attributes. Category combinations with no IDs listed have no corresponding forest plots in the dataset.}
    \bigskip
    \scriptsize{
    \begin{tabular}{lllr}
        \toprule
        \textbf{Species distribution} & \textbf{Forest density} & \textbf{Understory} & \textbf{Plot IDs} \\ \midrule \midrule
        Coniferous dominated & Sparse & Minimal & \\
        & & Moderate & \\
        & & Heavy & 1005 \\ \midrule
        Coniferous dominated & Moderate & Minimal & \\
        & & Moderate & \\
        & & Heavy & 1010 \\ \midrule
        Coniferous dominated & Dense & Minimal & \\
        & & Moderate & 1002 \\
        & & Heavy & 1001 \\ \midrule
        Deciduous dominated & Sparse & Minimal & 1019, 1028 \\
        & & Moderate & \\
        & & Heavy & \\ \midrule
        Deciduous dominated & Moderate & Minimal & 1023, 1031 \\
        & & Moderate & 1004, 1008 \\
        & & Heavy & \\ \midrule
        Deciduous dominated & Dense & Minimal & 1024 \\
        & & Moderate & 1003, 1009 \\
        & & Heavy & \\ \midrule
        Mixed & Sparse & Minimal & 1012, 1013 \\
        & & Moderate & \\
        & & Heavy & \\ \midrule
        Mixed & Moderate & Minimal & \\
        & & Moderate & 1020 \\
        & & Heavy & 1027 \\ \midrule
        Mixed & Dense & Minimal & \\
        & & Moderate & \\
        & & Heavy & 1018, 1022 \\ \midrule
    \end{tabular}
    }
    \label{table:forest_types}
\end{table}

The categories for forest density were chosen as \emph{sparse}, \emph{moderate} and \emph{dense}. Similarly to the species distribution attribute, the classifications were primarily based on visual inspection, supported by the average number of trees per hectare (see \autoref{section:training_and_test}). However, this metric is not always reliable for determining forest density, since a low proportion of understory trees or the presence of clearings within forest plots can skew the value downward, even when trees are closely grouped. For the stratified random sampling of the training and test sets, FGI-EMIT was divided into these three density categories.

Finally, the categories for the understory attribute were defined as:
\begin{enumerate}
    \item \textbf{Minimal:} plots with $<20\%$ of trees in the understory.
    \item \textbf{Moderate:} plots with $\geq20\%$ and $\leq35\%$ of trees in the understory.
    \item \textbf{Heavy:} plots with $>35\%$ of trees in the understory.
\end{enumerate}

The forest type classifications of all 19 plots are presented in \autoref{table:forest_types}.

\subsection{Utilizing FGI-EMIT for benchmarking} \label{appendix:utilizing_for_benchmarking}

The forest plots in the FGI-EMIT dataset are provided as point clouds in the \verb|las| file format, organized into two directories corresponding to the training and test sets. Each ground truth tree instance was assigned a positive integer identifier (ID) that is unique at the plot level. The IDs are stored as an additional attribute in the point clouds (\verb|tree_index|), such that all points belonging to the same trees share the same ID. Points not associated with any tree instance were given the ID 0.

In addition to the manually annotated point clouds, the dataset includes a user-configurable Python script for computing the segmentation accuracy metrics, as described in \autoref{section:accuracy_metrics}, and a file titled \verb|plot_data.yaml|, containing the automatically derived tree locations (see \autoref{section:height_and_location}), and crown categories (see \autoref{section:crown_categories}) for all plots.

To ensure comparability between studies, we encourage adhering to the following standardized benchmarking procedure when developing individual tree segmentation methods using the FGI-EMIT dataset:
\begin{enumerate}
    \item Use the exact training and test set split specified in \autoref{section:training_and_test}. The dataset is provided in a format where the two sets are separated by default. Although a validation set is not strictly necessary, we recommend employing the same validation plots as in this study, specifically plots 1019, 1022 and 1031.
    \item Points belonging to semantic category 5 (out) should be removed prior to inference. For unsupervised approaches, points in semantic categories 2--4 may also be excluded. However, this should \textbf{not} be done for supervised methods.
    \item If any preprocessing or data modification is performed, such as reducing point density, these must be clearly reported.
    \item Predicted segments should be matched to ground truth instances using the 3D intersection over union metric with a threshold of $\text{IoU}\geq50\%$, as described in \autoref{section:accuracy_metrics}. If an alternative matching criterion or threshold is used, it must be clearly specified.
    \item Extremely small predicted segments, which are likely to be erroneous, should be excluded prior to computing accuracy metrics. In our experiments, segments containing fewer than 40 points and less than 1.5 m in height were removed. This postprocessing step is handled automatically by the provided accuracy computation script.
    \item All applicable accuracy metrics defined in \autoref{section:accuracy_metrics} should be reported. Employing the official accuracy computation script provided with the dataset is strongly recommended. Metrics should be computed across the entire data split (training or test), rather than averaged at plot level.
\end{enumerate}

\subsection{Additional point cloud attributes} \label{appendix:additional_point_cloud_attributes}

In addition to the instance annotations, two binary attributes, \verb|edge| and \verb|dead|, were added during the manual labeling process. The former attribute indicates trees located along the plot boundaries that are missing at least some points, while the latter denotes trees that were definitively identified as dead during annotation. Dead trees were primarily recognized based on absence of leaves or needles in the point clouds. Consequently, the dataset may include some recently deceased trees that were not labeled correctly.

\begin{table}[!b]
    \centering
    \caption{Additional attributes available in the FGI-EMIT point clouds. The \emph{missing value} column indicates the placeholder used for attributes when no corresponding record exists for a point.} \bigskip
    \scriptsize{
    \begin{tabular}{lP{1.2cm}L{4.2cm}} \toprule
        \textbf{Attribute} & \textbf{Missing value} & \textbf{Description} \\ \midrule \midrule
        \verb|red| & - & Scaled reflectance from scanner 1. \\
        \verb|green| & - & Scaled reflectance from scanner 2. \\
        \verb|blue| & - & Scaled reflectance from scanner 3. \\
        \verb|intensity_1| & 0 & Intensity from scanner 1. \\
        \verb|intensity_2| & 0 & Intensity from scanner 2. \\
        \verb|intensity_3| & 0 & Intensity from scanner 3. \\
        \verb|amplitude_1| & 0 & Amplitude from scanner 1. \\
        \verb|amplitude_2| & 0 & Amplitude from scanner 2. \\
        \verb|amplitude_3| & 0 & Amplitude from scanner 3. \\
        \verb|reflectance_1| & $-9,999$ & Reflectance from scanner 1. \\
        \verb|reflectance_2| & $-9,999$ & Reflectance from scanner 2. \\
        \verb|reflectance_3| & $-9,999$ & Reflectance from scanner 3. \\
        \verb|deviation_1| & 65,535 & Echo deviation from scanner 1. \\
        \verb|deviation_2| & 65,535 & Echo deviation from scanner 2. \\
        \verb|deviation_3| & 65,535 & Echo deviation from scanner 3. \\
        \verb|user_data| & - & This value indicates which scanner an individual point is originally from. The points have been labeled either 1, 2, or 3 for scanners 1, 2, and 3 respectively. \\
        \verb|tree_index| & - & Manually generated tree instance annotation. All points in a tree instance share the same value, and each tree instance within a plot has a unique index. Points not belonging to any tree instance have \verb|tree_index| set to 0. \\
        \verb|edge| & - & Tree instances on the edge of the plot with some parts missing have this value set to 1 for all points. All other points have this value set to 0. \\
        \verb|dead| & - & Tree instances identified as dead via visual inspection have this value set to 1 for all points. All other points have this value set to 0. \\
        \bottomrule
    \end{tabular}
    }
    \label{table:additional_attributes}
\end{table}

Beyond the manual annotations, the point clouds in FGI-EMIT contain several non-standard attributes originating from each of the three scanners. A complete list of all available attributes is provided in \autoref{table:additional_attributes}. The same information is also included in the \verb|metadata.yaml| file distributed with FGI-EMIT.

\section{Unsupervised ITS algorithm hyperparameter optimization details} \label{appendix:hyperparameter_optimization}

This appendix provides additional details of the hyperparameter optimization of the unsupervised individual tree segmentation algorithms, including the identified optimal parameter values and the explored ranges. For each algorithm, the parameter configuration that achieved the highest F1-score on the training set was selected. The reported default hyperparameter values are based on either the recommendations in the corresponding paper or those specified in the source code. For detailed descriptions of the role of each hyperparameter, we refer the reader to the respective original publications.

\subsection{Watershed}

The hyperparameters optimized for the watershed ITS algorithm included the rasterization resolution (i.e., pixel size in meters), the standard deviation of the Gaussian filter $\sigma$, and the window sizes (in pixels) used by the Gaussian and maximum filters. Both Bayesian optimization and random search yielded similar results, with random search producing a configuration that provided slightly higher accuracy on the training set. The optimal parameter values, together with the tested ranges and step sizes, are presented in \autoref{table:watershed_params}.

\begin{table}[!b]
    \centering
    \caption{Optimal hyperparameter values and tested ranges of the watershed individual tree segmentation algorithm. GF and MF denote Gaussian filter and maximum filter, respectively} \bigskip
    \scriptsize
    \begin{tabular}{lR{1cm}R{1cm}R{1.2cm}r}
        \toprule
        \textbf{Hyperparameter} & \textbf{Default value} & \textbf{Optimal value} & \textbf{Tested range} & \textbf{Step size} \\ \midrule \midrule
        Resolution & 0.5 & 0.15 & $[0.05,1.00]$ & 0.05 \\
        $\sigma$ & 0.7 & 3.1 & $[0.1,6.0]$ & 0.1 \\
        Window size (GF) & 5 & 35 & $[3,41]$ & 2 \\
        Window size (MF) & 5 & 5 & $[3,41]$ & 2 \\
        \bottomrule
    \end{tabular}
    \label{table:watershed_params}
\end{table}

\subsection{3D adaptive mean shift}

AMS3D has two hyperparameters to optimize: the kernel diameter bandwidth slope and the kernel height bandwidth slope, denoted by $s^s$ and $s^z$, respectively. Both Bayesian optimization and random search converged to the same optimal parameter values, which are reported in \autoref{table:ams3d_params}.

\begin{table}[!b]
    \centering
    \caption{Optimal hyperparameter values and tested ranges of the AMS3D individual tree segmentation algorithm.} \bigskip
    \scriptsize
    \begin{tabular}{lR{1cm}R{1cm}R{1.2cm}r}
        \toprule
        \textbf{Hyperparameter} & \textbf{Default value} & \textbf{Optimal value} & \textbf{Tested range} & \textbf{Step size} \\ \midrule \midrule
        $s^s$ & 0.3 & 0.3 & $[0,1]$ & 0.1 \\
        $s^z$ & 0.4 & 0.8 & $[0,1]$ & 0.1 \\
        \bottomrule
    \end{tabular}
    \label{table:ams3d_params}
\end{table}

Since AMS3D was originally designed for ALS point clouds with an average point density of 10 points/m$^2$, its computational complexity renders performing the repeated evaluations required for parameter optimization infeasible on our high-density dataset. To address this, the input point clouds were randomly downsampled to an average density of $\sim100$ points/m$^2$. Because mean shift clustering depends on density distributions, this sparsification should not substantially affect segmentation quality. Moreover, the employed density is still approximately an order of magnitude denser than the data for which AMS3D was developed, ensuring the method remains applicable. At test time, the segmentation results obtained from the sparsified point clouds were propagated back to the original high-density data using $k$-nearest neighbors search with $k=1$.

A key limitation of AMS3D is its tendency to significantly oversegment the lower forest strata, producing a large number of very small segments. This behavior can also be observed in visualizations of the AMS3D segmentation results presented in previous works \citep[see e.g.][]{ferraz2016lidar,aubrykientz2019comparative}. The issue is particularly apparent in our high-density ALS data, which contains significantly more understory points than typical ALS datasets. In some cases, AMS3D identified more than a thousand segments on plots containing fewer than a hundred ground-truth trees. While canopy segments are usually delineated correctly and can be matched to ground truth based on location or 2D IoU, the more robust 3D IoU-based criterion is generally too demanding.

To alleviate this issue, we designed a simple segment merging algorithm to combine small understory segments into larger tree segments. The procedure iterates over all segments with a point count below a user-defined threshold and performs the following steps:
\begin{enumerate}
    \item Select the first segment in the list and compute the weighted distance from its center to all other segments.
    \item If the minimum distance is below a user-defined threshold or smaller than the distance to the ground, merge the nearest segments. Otherwise, remove the current segment from the list.
    \item Recompute the segment list. If it is not empty, return to Step 1.
\end{enumerate}
This simple merging procedure substantially reduced the number of erroneous small segments and improved the training set F1-score by 10--15 percentage points on average across all hyperparameter configurations.

\subsection{Layer stacking}

Layer stacking has a relatively large number of hyperparameters to optimize. Specifically, we considered the coarsest rasterization resolution (in meters), the filtering cutoff for abnormally large clusters, whether to apply DBSCAN to remove understory vegetation in lower forest strata, the width of the buffer fitted around clusters, the width of tree cores, and the window size used by the maximum filter when identifying local maxima from rasters. The optimal values and explored ranges are listed in \autoref{table:layerstacking_params}. Both Bayesian optimization and random search yielded parameter configurations with comparable segmentation accuracy, with the combination identified by BO performing slightly better.

\begin{table}[!t]
    \centering
    \caption{Optimal hyperparameter values and tested ranges of the layer stacking individual tree segmentation algorithm.} \bigskip
    \scriptsize
    \begin{tabular}{lR{1cm}R{1cm}R{1.5cm}r}
        \toprule
        \textbf{Hyperparameter} & \textbf{Default value} & \textbf{Optimal value} & \textbf{Tested range} & \textbf{Step size} \\ \midrule \midrule
        Resolution (coarsest) & 1.0 & 0.8 & $[0.05,1.00]$ & 0.05 \\
        Filtering cutoff & 3.5 & 3.5 & $[2.5,3.5]$ & 0.5 \\
        DBSCAN filtering & \verb|True| & \verb|False| & $\{$\verb|False|,\verb|True|$\}$ & - \\
        Buffer width & 0.6 & 1.5 & $[0.1,1.5]$ & 0.1 \\
        Tree core width & 0.6 & 0.9 & $[0.1,1.0]$ & 0.1 \\
        Window size & 3 & 4 & $[1,15]$ & 1 \\
        \bottomrule
    \end{tabular}
    \label{table:layerstacking_params}
\end{table}

As with AMS3D, layer stacking was originally designed for sparser point clouds, making hyperparameter optimization on the full-density data computationally infeasible. To address this, the point clouds were downsampled to an average density of 500 points/m$^2$. Because this density is still relatively close to the original, the level of geometric detail is preserved, and the sparsification is unlikely to have a significant impact on segmentation accuracy.

\subsection{Treeiso}

The hyperparameter optimization of Treeiso was limited to the set identified as the primary tunable parameters in the original paper \citep{xi20223d}. These include the number of nearest neighbors in the graphs of the first and second stage ($K_1$ and $K_2$), the regularization strength of the first and second cut-pursuit clustering ($\lambda_1$ and $\lambda_2$), the elevation-difference-to-length ratio threshold ($\rho_{\text{zmax}}$), and the horizontal overlap ratio weight ($w^\rho$). The optimal parameter values identified using Bayesian optimization are listed in \autoref{table:treeiso_params} for both the setup that used the same input data as the other unsupervised ITS algorithms and the alternative input containing only tree points.

\begin{table}[!t]
    \centering
    \caption{Optimal hyperparameter values and tested ranges of the Treeiso individual tree segmentation algorithm. Values marked with $\dagger$ correspond to the case where only points classified as trees were used as input.} \bigskip
    \scriptsize
    \begin{tabular}{lR{1cm}R{1cm}R{1cm}R{1cm}R{0.5cm}}
        \toprule
        \textbf{Hyperparameter} & \textbf{Default value} & \textbf{Optimal value} & \textbf{Optimal value$^\dagger$} & \textbf{Tested range} & \textbf{Step size} \\ \midrule \midrule
        $K_1$ & 5 & 3 & 3 & $[3,20]$ & 1 \\
        $K_2$ & 20 & 21 & 37 & $[3,40]$ & 1 \\
        $\lambda_1$ & 1 & 40 & 0.1 & $[0.1,40]$ & 0.1 \\
        $\lambda_2$ & 20 & 14.5 & 5.0 & $[5,40]$ & 0.1 \\
        $\rho_{\text{zmax}}$ & 0.5 & 0.95 & 0.7 & $[0.1,1]$ & 0.05 \\
        $w^\rho$ & 0.5 & 0.1 & 0.1 & $[0.1,2]$ & 0.1 \\
        \bottomrule
    \end{tabular}
    \label{table:treeiso_params}
\end{table}

Although most of the optimal hyperparameter values were relatively similar between the two setups, the regularization strengths differed substantially. In particular, the setup that included non-tree points benefited from significantly higher values, especially for $\lambda_1$, which was 400 times larger than that used for the tree-only input. Since the regularization strength directly controls the number of clusters, with higher values yielding fewer clusters, this difference is intuitive: when non-tree points, usually dominated by understory vegetation, are present, segmentation accuracy improves when the number of clusters is kept low. We conjecture that this is primarily due to preserving large, well-separated understory clusters that remain distinct from tree bases, thereby reducing the risk of merging them with the actual tree segments in later stages of the algorithm.

\section{Deep learning model training details} \label{appendix:dl_model_training}

This appendix provides additional details of the training process of the deep-learning-based individual tree segmentation models. As described in \autoref{section:dl_model_experimental_setup}, most model hyperparameters were adopted directly from the original publications, since they had been optimized for datasets with point densities comparable to FGI-EMIT, making them applicable in our case as well. For consistency, the same hyperparameter configurations were used across all input setups in the reflectance feature ablation study performed in \autoref{section:multispectral_ablation}. Where hyperparameter values are not explicitly listed, default values were used.

\subsection{YOLOv12}

YOLOv12 was trained using the default input image size of $640\times640$ pixels and a batch size of 4. The training was conducted using an AdamW optimizer \citep{loshchilov2017decoupled} with an initial learning rate of $0.001667$, weight decay of $0.0005$, and a cosine learning rate scheduler \citep{loshchilov2016sgdr}. The number of training epochs was set to 1,000, with early stopping triggered if no improvement in validation set accuracy metrics was observed over 100 consecutive epochs. The model weights were initialized from a network pretrained on the MSCOCO 2017 dataset \citep{lin2014microsoft}, which provided a slight improvement in segmentation accuracy compared to training from scratch. Due to the relatively small dataset size, the training process was fast, requiring only approximately 15 minutes.

\subsection{SegmentAnyTree}

SegmentAnyTree was trained with the default input cylinder radius of 8 m and a batch size 4. An Adam optimizer \citep{kingma2014adam} with a learning rate of 0.001, $\beta=(0.9,0.999)$, no weight decay, and an exponential learning rate scheduler was used. The number of training epochs was set to 160, based on training and validation loss curves as well as accuracy trends. The first 30 epochs were a warm-up period during which clustering was not performed. Following the original paper, we applied the proposed data augmentation strategy in which point clouds downsampled to approximate densities of 1,000; 500; 100; 75; 50; 25; and 10 points/m$^2$ were included in the training data. We additionally experimented with training the model using only the original full-density point clouds, but this resulted in slightly lower segmentation accuracy. Consequently, the augmented setup with sparsified inputs was adopted. Training Seg\-ment\-Any\-Tree required approximately 70 hours.

\subsection{TreeLearn}

Following \citet{xiang2025forestformer3d}, the number of training epochs was reduced from the default 1,400 to 1,200, the initial learning rate from 0.002 to 0.001, and the total number of training samples from 25,000 to 2,500. The batch size was increased from 2 to 4, which was the maximum permitted by the available GPU memory, while the size of the rectangular input tiles was kept at the default value of 35 meters. As in the original paper, we used an AdamW optimizer with a weight decay of $0.001$, $\beta=(0.9,0.999)$ and a cosine learning rate scheduler. The total training time was approximately 30 hours.

\subsection{ForestFormer3D}

Following the original paper, ForestFormer3D was trained using an AdamW optimizer with an initial learning rate of 0.0001, $\beta=(0.9,0.999)$, a weight decay of 0.05, and a polynomial learning rate scheduler. The number of training epochs was set to 6,500 based on training and validation losses, as well as accuracy metrics. Due to GPU memory limitations, the input cylinder radius was reduced from 16 to 12 meters. The batch size was kept at the default value of 2. The total training time on our system was approximately 15 hours.

As recommended in the source code documentation, inference was performed twice on extremely dense forest plots: once on the original point clouds and a second time on the remaining points. On FGI-EMIT, this was only necessary for plot 1018. On all other plots, performing multiple inferences drastically reduced model precision and, by extension, the overall F1-score.

Although ForestFormer3D is a panoptic segmentation model capable of simultaneously performing semantic and instance predictions, we restricted the number of semantic classes to two (tree and non-tree), since the FGI-EMIT dataset does not contain semantic annotations for wood, foliage, or ground. We also experimented with training the model using semantic predictions from a ForestFormer3D model trained on FOR-InstanceV2 as the ground truth. However, this setup yielded results comparable to training with only the tree/non-tree semantic ground truth, with the only notable difference being slightly faster convergence at 5,500 epochs.

\section{Training set accuracy metrics}

For completeness, this appendix provides the FGI-EMIT training set accuracy metrics, corresponding to the test set metrics reported in \autoref{section:results}.

\subsection{Performance comparison metrics} \label{appendix:performance_comparison_metrics}

This section reports the quantitative performance metrics on the FGI-EMIT training set, corresponding to the test set metrics presented in the performance comparison in \autoref{section:performance_comparison}. Metrics for all benchmarked individual tree segmentation methods are provided in \autoref{table:performance_comparison_test}, while crown category-level accuracies are listed in \autoref{table:crown_ctg_comparison_test}

Overall, the training set metrics for all deep learning models are higher than the corresponding test set values. However, the absolute difference remains approximately constant across all methods, which suggests the higher metrics are not due to model overfitting, but rather reflect the training set’s slightly lower segmentation difficulty.  Particularly, this appears to stem from trees in the more challenging crown categories (B, C and D) being easier to segment correctly on the training plots. This also explains why the metrics reported for the unsupervised algorithms are approximately equivalent between the training and test sets, since the methods detect relatively few trees in the more difficult crown categories.

\begin{table*}[!t]
    \centering
    \caption{Comparison of unsupervised individual tree segmentation algorithms and deep-learning-based approaches on the \textbf{training split} of the FGI-EMIT dataset. The best performance metrics are shown in \textbf{bold}, and the second-best are \underline{underlined}. Results marked with $\dagger$ indicate that only points classified as trees were used as input.} \bigskip
    \small{
    \begin{tabular}{lc*{5}{S[table-format=2.1]}S[table-format=1]}
        \toprule
        \textbf{Model} & \textbf{DL} & \textbf{Precision (\%)} & \textbf{Recall (\%)} & \textbf{F1-score (\%)} & \textbf{Cov (\%)} & \textbf{AP$_{\mathbf{50}}$ (\%)} & \parbox{1.8cm}{\centering\textbf{Average time (s/plot)}} \\ \midrule \midrule
        Watershed \citep{yu2011predicting} && 64.0 & 35.4 & 45.6 & 34.7 & \text{-} & \underline{6} \\
        AMS3D \citep{ferraz2016lidar} && 57.2 & 26.4 & 36.1 & 29.5 & \text{-} & 196 \\
        Layer stacking \citep{ayrey2017layer} && 63.5 & 24.2 & 35.1 & 24.1 & \text{-} & 68 \\
        Treeiso \citep{xi20223d} && 49.9 & 42.5 & 45.9 & 44.4 & \text{-} & 103 \\
        Treeiso$^\dagger$ \citep{xi20223d} && 59.8 & 44.7 & 51.2 & 46.8 & \text{-} & 162 \\
        YOLOv12 \citep{tian2025yolov12} & $\checkmark$ & \underline{86.6} & 53.7 & 66.3 & 46.3 & 49.7 & \textbf{3} \\
        SegmentAnyTree \citep{wielgosz2024segmentanytree} & $\checkmark$ & 85.0 & \underline{75.4} & \underline{79.9} & \underline{69.6} & \underline{68.0} & 263 \\
        TreeLearn \citep{henrich2024treelearn} & $\checkmark$ & 83.1 & 71.7 & 77.0 & \underline{69.6} & 54.1 & 132 \\
        ForestFormer3D \citep{xiang2025forestformer3d} & $\checkmark$ & \textbf{96.5} & \textbf{78.2} & \textbf{86.4} & \textbf{73.3} & \textbf{77.5} & 204 \\
        \bottomrule
    \end{tabular}}
    \label{table:performance_comparison_train}
\end{table*}

\begin{table*}[!t]
    \centering
    \caption{Comparison of crown category-level recall on the \textbf{training split} of the FGI-EMIT dataset. The best performance metrics are shown in \textbf{bold}, and the second-best are \underline{underlined}. Results marked with $\dagger$ indicate that only points classified as trees were used as input.} \bigskip
    \begin{tabular}{lc*{4}{S[table-format=2.1]}}
        \toprule
        \textbf{Model} & \textbf{DL} & \textbf{Recall$_\text{A}$ (\%)} & \textbf{Recall$_\text{B}$ (\%)} & \textbf{Recall$_\text{C}$ (\%)} & \textbf{Recall$_\text{D}$ (\%)} \\ \midrule \midrule
        Watershed \citep{yu2011predicting} && 70.8 & 34.9 & 5.9 & 0.0 \\
        AMS3D \citep{ferraz2016lidar} && 58.5 & 18.7 & 2.2 & 0.8 \\
        Layer stacking \citep{ayrey2017layer} && 57.2 & 13.6 & 0.3 & 0.0 \\
        Treeiso \citep{xi20223d} && 77.6 & 31.1 & 21.4 & 6.8 \\
        Treeiso$^\dagger$ \citep{xi20223d} && 76.4 & 37.0 & 25.7 & 7.5 \\
        YOLOv12 \citep{tian2025yolov12} & $\checkmark$ & 90.7 & 60.4 & 23.8 & 1.5 \\
        SegmentAnyTree \citep{wielgosz2024segmentanytree} & $\checkmark$ & \underline{96.6} & 71.1 & \underline{68.1} & \textbf{36.1} \\
        TreeLearn \citep{henrich2024treelearn} & $\checkmark$ & 89.4 & \underline{74.0} & 65.9 & 27.1 \\
        ForestFormer3D \citep{xiang2025forestformer3d} & $\checkmark$ & \textbf{96.8} & \textbf{77.9} & \textbf{72.8} & \underline{35.3} \\
        \bottomrule
    \end{tabular}
    \label{table:crown_ctg_comparison_train}
\end{table*}

\subsection{Multispectral reflectance ablation study metrics} \label{appendix:multispectral_information}

This section provides the training set performance metrics corresponding to the test set results reported in the multispectral reflectance ablation study described in \autoref{section:multispectral_ablation}. The overall performance metrics of the ablated models are presented in \autoref{table:scanner_performance_comparison_test}, with the crown category-level results shown in \autoref{table:scanner_crown_ctg_comparison_test}.

\begin{table*}[!t]
    \centering
    \caption{Ablation study on the effects of using mono- and multispectral data. The accuracies are reported for the \textbf{training split} of FGI-EMIT. SAT, TL and FF3D denote SegmentAnyTree \citep{wielgosz2024segmentanytree}, TreeLearn \citep{henrich2024treelearn} and ForestFormer3D \citep{xiang2025forestformer3d}, respectively. For each model, the best performance metrics are shown in \textbf{bold}, and the second-best are \underline{underlined}.} \bigskip
    \scriptsize{
    \begin{tabular}{
        ccc|
        *{3}{S[table-format=2.1]}|
        *{3}{S[table-format=2.1]}|
        *{3}{S[table-format=2.1]}|
        *{3}{S[table-format=2.1]}|
        *{3}{S[table-format=2.1]}
    }
        \toprule
        \multicolumn{3}{c|}{\textbf{Scanners}} &
        \multicolumn{3}{c|}{\textbf{Precision (\%)}} &
        \multicolumn{3}{c|}{\textbf{Recall (\%)}} &
        \multicolumn{3}{c|}{\textbf{F1-score (\%)}} &
        \multicolumn{3}{c|}{\textbf{Cov (\%)}} &
        \multicolumn{3}{c}{\textbf{AP$_{\mathbf{50}}$ (\%)}} \\
        \text{Scanner 1} & \text{Scanner 2} & \text{Scanner 3} & SAT & TL & \text{FF3D} & SAT & TL & \text{FF3D} & SAT & TL & \text{FF3D} & SAT & TL & \text{FF3D}  & SAT & TL & \text{FF3D} \\
        \midrule \midrule
        & & & \textbf{85.0} & 83.1 & 96.5 & \underline{75.4} & 71.7 & \textbf{78.2} & \textbf{79.9} & 77.0 & \textbf{86.4} & \underline{69.6} & 69.6 & \textbf{73.3} & \underline{68.0} & 54.1 & \textbf{77.5} \\
        $\checkmark$ & & & 81.5 & 84.3 & 96.0 & 73.2 & 72.3 & \underline{77.4} & 77.2 & 77.8 & \underline{85.7} & 68.7 & 69.4 & \underline{72.3} & 65.1 & 55.9 & \underline{77.0} \\
        & $\checkmark$ & & 84.4 & 82.8 & 95.2 & 75.1 & 72.3 & 76.4 & 79.5 & 77.2 & 84.8 & 69.3 & 69.6 & 72.2 & 67.9 & 54.3 & 76.0 \\
        & & $\checkmark$ & 84.2 & 84.6 & \textbf{97.1} & 74.5 & \underline{72.4} & 76.2 & 79.1 & 78.0 & 85.4 & 69.2 & \underline{69.7} & 71.7 & 66.4 & 56.4 & 75.8 \\
        $\checkmark$ & $\checkmark$ & & \underline{84.7} & \underline{85.2} & 96.1 & 73.6 & 72.1 & 74.9 & 78.7 & \underline{78.1} & 84.2 & 68.3 & 69.6 & 70.8 & 65.5 & \underline{58.3} & 74.3 \\
        $\checkmark$ & & $\checkmark$ & 84.1 & 83.4 & 96.4 & 73.3 & \textbf{73.0} & 75.4 & 78.3 & 77.8 & 84.6 & 68.6 & \textbf{70.7} & 71.1 & 65.3 & 56.1 & 74.6 \\
        & $\checkmark$ & $\checkmark$ & 84.0 & \textbf{85.4} & 94.3 & 73.0 & 72.0 & 61.7 & 78.1 & \textbf{78.2} & 74.6 & 68.2 & \underline{69.7} & 59.0 & 65.4 & \textbf{58.5} & 60.9 \\
        $\checkmark$ & $\checkmark$ & $\checkmark$ & 84.3 & 84.0 & \underline{96.6} & \textbf{75.6} & 71.9 & 75.3 & \underline{79.7} & 77.5 & 84.6 & \textbf{70.1} & 69.5 & 70.7 & \textbf{68.4} & 56.0 & 74.8 \\
        \bottomrule
    \end{tabular}
    }
    \label{table:scanner_performance_comparison_train}
\end{table*}

\begin{table*}[!t]
    \centering
    \caption{Ablation study on the effects of using mono- and multispectral data. The crown category-level accuracies are reported for the \textbf{training split} of FGI-EMIT. SAT, TL and FF3D denote SegmentAnyTree \citep{wielgosz2024segmentanytree}, TreeLearn \citep{henrich2024treelearn} and ForestFormer3D \citep{xiang2025forestformer3d}, respectively. For each model, the best performance metrics are shown in \textbf{bold}, and the second-best are \underline{underlined}.} \bigskip
    \small{
    \begin{tabular}{
        ccc|
        *{3}{S[table-format=2.1]}|
        *{3}{S[table-format=2.1]}|
        *{3}{S[table-format=2.1]}|
        *{3}{S[table-format=2.1]}
    }
        \toprule
        \multicolumn{3}{c|}{\textbf{Scanners}} &
        \multicolumn{3}{c|}{\textbf{Recall$_\text{A}$ (\%)}} &
        \multicolumn{3}{c|}{\textbf{Recall$_\text{B}$ (\%)}} &
        \multicolumn{3}{c|}{\textbf{Recall$_\text{C}$ (\%)}} &
        \multicolumn{3}{c}{\textbf{Recall$_\text{D}$ (\%)}} \\
        Scanner 1 & Scanner 2 & Scanner 3 & SAT & TL & \text{FF3D} & SAT & TL & \text{FF3D} & SAT & TL & \text{FF3D} & SAT & TL & \text{FF3D}  \\
        \midrule \midrule
        & & & \underline{96.3} & 89.4 & 96.8 & 71.5 & 74.0 & \textbf{77.9} & \textbf{68.1} & \underline{65.9} & \textbf{72.8} & \underline{35.9} & 27.1 & 35.3 \\
        $\checkmark$ & & & 95.1 & \underline{91.4} & \underline{97.3} & 71.1 & \textbf{75.7} & 75.3 & 63.5 & 63.8 & \underline{70.3} & 33.8 & \underline{28.6} & \textbf{37.6} \\
        & $\checkmark$ & & \textbf{96.8} & 90.4 & \textbf{97.8} & \underline{72.3} & \underline{75.3} & \underline{75.7} & 66.2 & 65.6 & 67.5 & 35.3 & 27.8 & 33.8 \\
        & & $\checkmark$ & 95.6 & 90.9 & \underline{97.3} & 70.2 & \textbf{75.7} & 74.0 & \underline{66.6} & 65.3 & 67.8 & \textbf{36.8} & 28.1 & \underline{36.1} \\
        $\checkmark$ & $\checkmark$ & & \underline{96.3} & 90.2 & 96.3 & 69.4 & 73.6 & 73.2 & 65.3 & \underline{65.9} & 65.9 & 31.6 & \textbf{29.3} & 33.8 \\
        $\checkmark$ & & $\checkmark$ & 94.1 & \textbf{91.9} & 96.8 & 71.9 & 73.6 & 74.0 & 65.0 & \textbf{66.6} & 66.3 & 32.3 & \textbf{29.3} & 34.6 \\
        & $\checkmark$ & $\checkmark$ & 95.6 & 90.9 & 92.6 & 68.9 & \underline{75.3} & 60.9 & 64.8 & 64.4 & 43.3 & 32.3 & 27.1 & 12.8 \\
        $\checkmark$ & $\checkmark$ & $\checkmark$ & 95.8 & \underline{91.4} & 97.1 & \textbf{76.6} & 73.2 & 74.9 & \underline{66.6} & 64.1 & 66.3 & 33.8 & \textbf{29.3} & 31.6 \\
        \bottomrule
    \end{tabular}
    }
    \label{table:scanner_crown_ctg_comparison_train}
\end{table*}

\end{appendices}

\end{document}